\documentclass[11pt]{article}

\usepackage[final]{acl}
\usepackage{times}
\usepackage{float}
\usepackage{placeins}
\usepackage{latexsym}
\usepackage{tcolorbox}
\usepackage{multirow}

\usepackage{hyperref}
\usepackage[T1]{fontenc}

\usepackage[utf8]{inputenc}

\usepackage{microtype}

\usepackage{inconsolata}

\usepackage{graphicx}
\usepackage{amsmath}
\usepackage{booktabs,tabularx}
\usepackage{subcaption}
\usepackage{soul,xcolor}
\usepackage{CJKutf8}
\usepackage{wasysym}
\usepackage{capt-of}

\title{Bias in the Ear of the Listener: Assessing Sensitivity in Audio Language Models Across Linguistic, Demographic, and Positional Variations}

\newcommand{\samethanks}[1][\value{footnote}]{\footnotemark[#1]}

\author{
    Sheng-Lun Wei\thanks{Equal contribution.}\textsuperscript{$\alpha$} \ 
    Yu-Ling Liao\samethanks\textsuperscript{$\alpha$}\ 
    Yen-Hua Chang\textsuperscript{$\alpha$}\ 
    Hen-Hsen Huang\textsuperscript{$\beta$}\ 
    Hsin-Hsi Chen\textsuperscript{$\alpha$$\gamma$}
    \\
    \textsuperscript{$\alpha$}Department of Computer Science and Information Engineering, \\
    National Taiwan University, Taiwan
    \\
    \textsuperscript{$\beta$}Institute of Information Science, Academia Sinica, Taiwan
    \\
    \textsuperscript{$\gamma$}AI Research Center (AINTU), National Taiwan University, Taiwan
    \\
    \texttt{\{weisl,ylliao,yhchang\}@nlg.csie.ntu.edu.tw},
    \\
    \texttt{hhhuang@iis.sinica.edu.tw,\quad hhchen@ntu.edu.tw}
}

\begin{document}
\maketitle
\begin{abstract}
This work presents the first systematic investigation of speech bias in multilingual MLLMs.
We construct and release the \textbf{\textsc{BiasInEar}} dataset, a speech-augmented benchmark based on Global MMLU Lite, spanning English, Chinese, and Korean, balanced by \textit{gender} and \textit{accent}, and totaling 70.8 hours ($\approx$4,249 minutes) of speech with 11,200 questions.
Using four complementary metrics (accuracy, entropy, APES, and Fleiss’~$\kappa$), we evaluate nine representative models under linguistic (\textit{language} and \textit{accent}), demographic (\textit{gender}), and structural (\textit{option order}) perturbations.
Our findings reveal that MLLMs are relatively robust to demographic factors but highly sensitive to \textit{language} and \textit{option order}, suggesting that speech can amplify existing structural biases.
Moreover, architectural design and reasoning strategy substantially affect robustness across languages.
Overall, this study establishes a unified framework for assessing fairness and robustness in speech-integrated LLMs, bridging the gap between text- and speech-based evaluation. The resources can be found at \href{https://github.com/ntunlplab/BiasInEar}{https://github.com/ntunlplab/BiasInEar}

\end{abstract}

\newcommand{\hlc}[2]{{\sethlcolor{#1}\hl{#2}}}
\begin{table*}[t]
\centering
\scriptsize
\renewcommand{\arraystretch}{1.2}
\begin{tabular}{@{} p{0.28\textwidth} p{0.32\textwidth} p{0.34\textwidth} @{}}
\toprule
\textbf{Raw Text} & \textbf{Direct TTS (Naïve Output)} & \textbf{Converted Text (Ours)}\\
\midrule

\verb|(1, 2, 5, 4)(2, 3)| 
& The \hlc{red!30}{product} of the cycles one, two, five, four and the cycle two, three.
&  The \hlc{green!30}{permutation}
 consisting of the cycle one–two–five–four, and the cycle two–three \\
\addlinespace

$\Sigma_{v \in V}\; \mathrm{degree}(v)$ 
& the sum of the degrees of \hlc{red!30}{all vertices in V is v}.
& the sum over \hlc{green!30}{lowercase v in uppercase V of degree of v}.  \\
\addlinespace

When traveling north from the United States into Canada you’ll see the North Star (Polaris) getting \_\_\_\_\_
& When traveling north from the United States into Canada you’ll see the North Star (Polaris) getting. 
& When traveling north from the United States into Canada you’ll see the North Star Polaris getting \hlc{green!30}{blank}.\\
\addlinespace

H2PO4-, HPO42-.
&  \hlc{red!30}{H two P O four minus, H P O four two minus}.
& \hlc{green!30}{Dihydrogen phosphate, hydrogen phosphate}.\\
\addlinespace

NH4+(aq) + NO2-(aq) → N2(g) + 2H2O(l).
& \hlc{red!30}{N H four plus aqueous} plus \hlc{red!30}{N O two minus} aqueous yields \hlc{red!30}{N two} gas plus two \hlc{red!30}{H two O} liquid.
& The \hlc{green!30}{reaction} between \hlc{green!30}{ammonium ion} in \hlc{green!30}{aqueous solution} and \hlc{green!30}{nitrite ion} in aqueous solution yields \hlc{green!30}{nitrogen} gas and two water molecules in liquid form.\\
I. GATT ; II. IMF.
& \hlc{red!30}{One GATT two I M F}.
& \hlc{green!30}{Roman numeral one}, \hlc{green!30}{General Agreement on Tariffs and Trad}; \hlc{green!30}{Roman numeral two}, \hlc{green!30}{International Monetary Fund}\\

\bottomrule
\end{tabular}
\caption{Illustration of the difference between naïve TTS and spoken-readable conversions.}
\label{tab:rewrite-examples}
\end{table*}

\section{Introduction}

The rapid progress of large language models (LLMs) has fundamentally reshaped natural language processing~\cite{OpenAIChatGPT,Anil2023GeminiAF,anthropic2025claude37}.
Recent advances extend LLMs beyond text-only inputs to multimodal settings, incorporating modalities such as vision~\cite{openai_gpt-4o_2024,agrawal2024pixtral12b,meta2025llama4} and speech~\cite{comanici2025gemini25pushingfrontier,liu2025voxtral,microsoft2025phi4minitechnicalreportcompact}, and achieving remarkable performance on a wide range of downstream tasks.
In particular, systems that accept spoken inputs are capable of directly handle spoken queries, enabling applications in spoken question answering~\cite{nachmani2024spoken,shih24b_interspeech}, conversational assistants~\cite{tang2024salmonn,zhang-etal-2023-speechgpt,rubenstein2023audiopalmlargelanguagemodel}, and educational technologies~\cite{10.1145/3706599.3720162,ma2025assessmentl2oralproficiency}.

\begin{figure}
    \centering
    \includegraphics[width=1\linewidth]{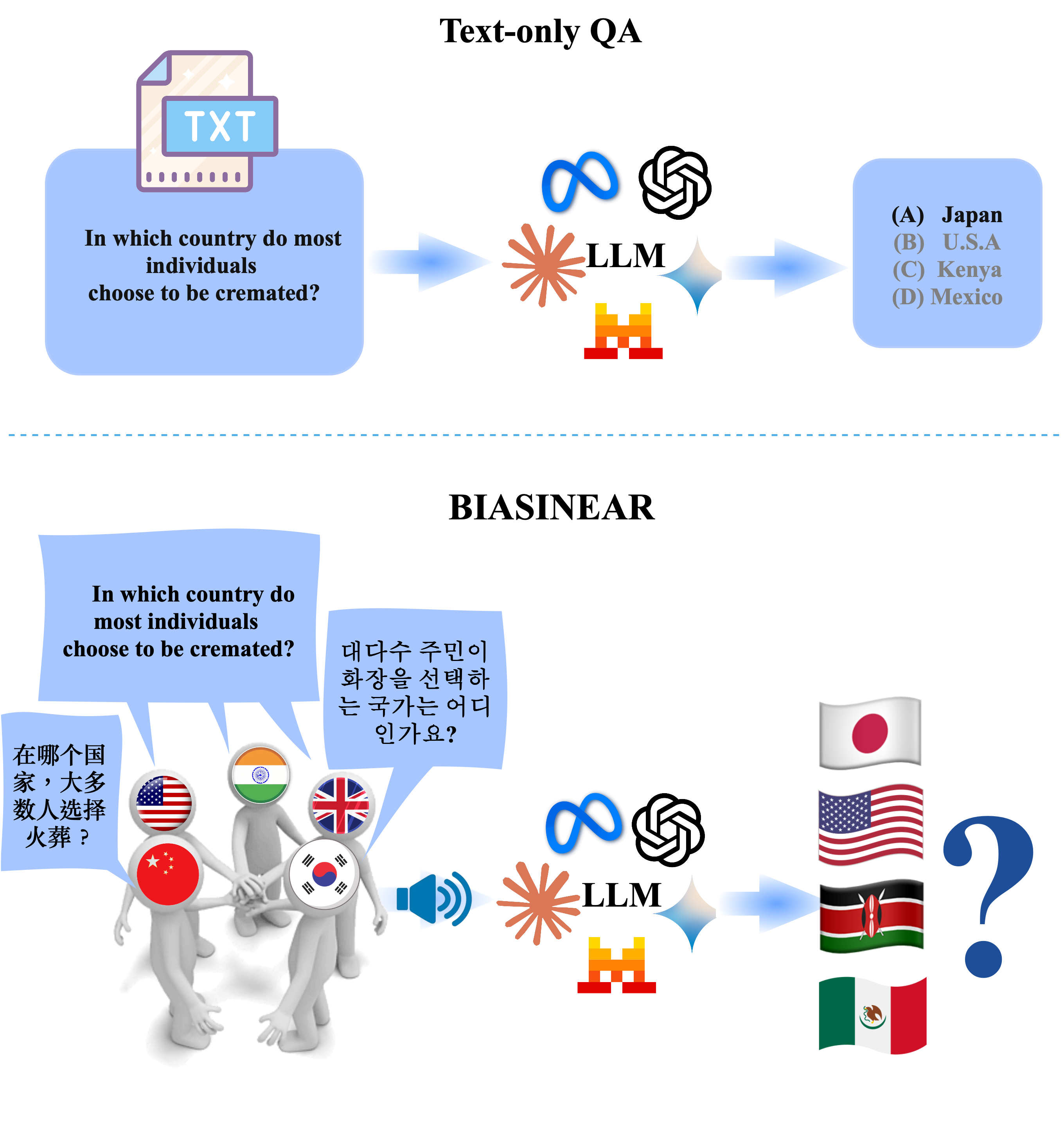}
    \caption{Overview of this work, which extends question answering from text inputs to multilingual spoken contents across languages, accents, and speakers.}
\end{figure}

However, recent studies have made clear that LLMs are not free from systematic biases across both demographic and structural dimensions. For instance, they have demonstrated that large language models encode various forms of social and cultural bias, including gender~\cite{ICLR2024_37771cc0, vo2025bscore}, race, dialect~\cite{hofmann2024dialectprejudicepredictsai}, nationality, and religion~\cite{shrawgi-etal-2024-uncovering, li2024culturepark, naous-etal-2024-beer}, as well as imbalances arising from Western-centric training data. 
Beyond these demographic dimensions, \citet{wei-etal-2024-unveiling} demonstrate that LLMs also suffer from selection bias when the order of answer options is altered in multiple-choice question answering tasks.
Taken together, these findings show that LLM predictions are influenced not only by semantic content but by superficial inputs and latent social factors, raising serious concerns about fairness and robustness in decision-making contexts.
At the same time, speech technologies introduce additional sources of bias. Prior research has shown that automatic speech recognition (ASR) systems often exhibit systematic performance disparities across demographic and linguistic factors, including gender~\cite{harris-etal-2024-modeling, doi10.1073/pnas.1915768117, Kulkarni_2024, attanasio-etal-2024-twists}, accent~\cite{10.1121/10.0024876, ijcai2023p578, tadimeti-etal-2022-evaluation, chan22b_interspeech}, and language resource availability~\cite{babu22_interspeech}. These findings suggest that transitioning QA tasks from text to speech may not only inherit existing LLM biases but amplify them through additional layers of demographic and linguistic variability.

Our main contributions are threefold:
\textbf{a)} We construct and release the \textbf{\textsc{BiasInEar}} dataset, 
a multilingual spoken QA benchmark covering English (with American, British, and Indian accents), 
Chinese (with Beijing and Northeastern accents), and Korean (with Seoul and Jeolla accents), with balanced male and female speakers. 
The dataset comprises 70.8 hours ($\approx$4,249 minutes) of speech and 11,200 questions, enabling large-scale and balanced evaluation across languages and demographic factors.
\textbf{b)} Leveraging this dataset, we perform comprehensive analyses across linguistic (\textit{language} and \textit{accent}), demographic (\textit{gender}), and structural (\textit{option order}) dimensions, extending the selection bias framework proposed in prior text-based studies~\cite{wei-etal-2024-unveiling} to the speech modality. 
\textbf{c)} Our study thus bridges the gap between LLM bias research and speech applications, 
offering new insights into fairness and robustness in multilingual speech technologies.

\section{\textbf{\textsc{BiasInEar}} Dataset}
\label{sec:dataset}
To investigate audio sensitivity in multilingual settings, we build upon the Global MMLU Lite~\cite{singh-etal-2025-global} by extending its text-based questions into spoken inputs, enabling a systematic analysis of model behavior under diverse audio conditions. Global MMLU Lite is a curated subset of Global MMLU, a high-quality multilingual extension of MMLU~\cite{hendrycks2021measuring}, and includes both culturally sensitive (CS) and culturally agnostic (CA) labels annotated by human experts. In this work, we focus on English, Chinese, and Korean as representative languages, each varying along factors such as \textit{gender}, \textit{accent}, and \textit{option order}, to comprehensively address our research questions. Specifically, we construct a multilingual speech-based version of MMLU that incorporates diverse \textit{gender} and \textit{accent} features, allowing us to probe the robustness of LLMs in spoken question answering. The final dataset comprises 70.8 hours  ($\approx$ 4,249 minutes) of speech across English, Chinese, and Korean, covering
 11,200 questions in total.

\subsection{Dataset Construction}
\paragraph{Question Rewriting} 
A direct conversion of text-based questions into speech can yield undesirable outcomes, particularly when the questions contain mathematical expressions, domain-specific symbols, or placeholders. Prior work~\cite{chen2024voicebenchbenchmarkingllmbasedvoice,tan-etal-2025-ssr} has addressed this issue by filtering out math-intensive subjects, thereby avoiding recognition errors. However, this approach reduces the diversity of the dataset by excluding STEM-related questions. To overcome this limitation, and inspired by~\citet{roychowdhury25_interspeech}, we introduce a rewriting step in which each question and its options are reformulated into a format that can be naturally and unambiguously read aloud.
Representative rewritten examples are presented in Table~\ref{tab:rewrite-examples}. Specifically, we employ the \texttt{GPT OSS 120B} to perform the rewriting, guided by the instruction prompt shown in Figure~\ref{fig:spoken_rendering_prompt} of Appendix~\ref{sec:rewriting-details}. Additional implementation details are provided therein due to space constraints.

\paragraph{Voice Generation} 
We generate audio for each question and option using the spoken-readable text produced during the rewriting stage. 
All text-to-speech (TTS) synthesis is conducted with the \texttt{Gemini 2.5 Flash Preview TTS} model, which supports multilingual generation. 
To ensure that the synthesized audio accurately reflects the target \textit{language} and \textit{accent}, we use the structured prompt shown in Figure~\ref{fig:tts-prompt} of Appendix~\ref{sec:tts_generation_prompt}.
This prompt explicitly specifies linguistic attributes, enabling consistent generation across English, Chinese, and Korean. 
This design allows controlled variation in \textit{language} and \textit{accent}, supporting robust multilingual evaluation.



\subsection{Quality Assessment}
\label{sec:quality-check-main}

\paragraph{Question Rewriting}
We normalize the rewritten outputs and the original Global MMLU Lite inputs by removing whitespace and converting all characters to lowercase. We then conduct a diff-based comparison to identify discrepancies. Flagged cases are manually reviewed, and any detected errors in the rewritten outputs are corrected to maintain consistency and accuracy. Table~\ref{tab:rewrite-qc} in Appendix ~\ref{sec:rewriting_qc}  summarizes the proportion of automatically flagged instances and the true error rate confirmed through manual inspection.



\paragraph{Voice Generation} 
We assess TTS quality using a two stage pipeline that combines automatic screening with manual verification.
In the automatic stage, each audio sample is transcribed using two widely adopted ASR systems, Whisper Large v3~\cite{radford2022whisper}and Omnilingual ASR~\cite{omnilingualasrteam2025omnilingualasropensourcemultilingual}.
Because a single ASR model may introduce recognition errors, using two independent systems improves the reliability of WER based quality checks.
For each sample, we compute WER against the rewritten text for both transcripts and take the minimum value as the final score.
Samples are then grouped into four WER ranges ($0$, $(0,0.2]$, $(0.2,0.6]$, and $>0.6$), and per language distributions are reported in Table~\ref{tab:wer-distribution}. This automatic screening step serves as a quality control mechanism to identify potential synthesis errors before human inspection.


\begin{table}[htbp]
  \centering
  \scriptsize
  \setlength{\tabcolsep}{4pt}
  \renewcommand{\arraystretch}{1.15}
  \resizebox{\columnwidth}{!}{%
  \begin{tabular}{@{}c|ccc@{}}
    \toprule
    \textbf{Interval} & \textbf{English} & \textbf{Chinese} & \textbf{Korean} \\
    \midrule
    $0$              & 10882 (90.68\%) & 5406 (67.58\%) & 5503 (68.79\%) \\
    $(0,\,0.2]$         & 937 (7.81\%)    & 1758 (21.98\%)  &  1714 (21.43\%)  \\
    $(0.2,\,0.6]$    & 143 (1.19\%)   & 621 (7.76\%)  & 663 (8.29\%)  \\
    $>0.6$             & 38 (0.32\%) & 215 (2.69\%)   & 120  (1.50\%)   \\
    \bottomrule
  \end{tabular}}
  \caption{Distribution across WER intervals by language.}
  \label{tab:wer-distribution}
\end{table}
\paragraph{Human Evaluation of TTS Quality}
To mitigate the risk of transcription errors underestimating dataset quality, we complement automatic evaluation with manual annotation. 
From each nonzero WER bin, 40 clips per language are randomly sampled using a stratified strategy to ensure representativeness. 
Annotation details are in Appendix~\ref{sec:quality_assessment}. 
Each clip is rated on a three-level scale: 
\textbf{Correct} (accurate and intelligible), 
\textbf{Acceptable} (minor mispronunciations but understandable), and 
\textbf{Incorrect} (severe errors causing misunderstanding). 
Table~\ref{tab:tts-annotation} shows the distribution of ratings across WER bins and languages. 
Most clips are rated as "Correct", indicating that TTS outputs are fluent, faithful, and well-aligned with the rewritten text. 
Many clips with nonzero WER also receive high manual ratings, implying that discrepancies mainly stem from ASR transcription or homophone errors rather than genuine TTS degradation. 
Together with automatic filtering, human evaluation forms a two-stage process ensuring dataset quality and consistency.
\begin{table}[h]
\centering
\small
\begin{tabular}{lccc}
    \toprule
    \textbf{Language} & \textbf{Incorrect} & \textbf{Acceptable} & \textbf{Correct} \\
    \midrule
    English & 6 (6.12\%) & 11 (11.22\%) &  81(82.65\%) \\
    Chinese & 7 (7\%) & 7 (7\%) &  86(86\%) \\
    Korean  & 8 (8\%) & 17 (17\%) & 75 (75\%) \\
    \bottomrule
\end{tabular}
\caption{Manual annotation results by \textit{language} with ratings of \textbf{Correct}, \textbf{Acceptable}, and \textbf{Incorrect}.}
\label{tab:tts-annotation}
\end{table}

\begin{table}[t]
\centering
\small
\begin{tabular}{p{1.8cm} p{4.5cm}}
\toprule
\textbf{Variable} & \textbf{Levels} \\
\midrule
Language & English, Chinese (high-resource) \newline Korean (medium-resource) \\
\hline
Accent   & English: American, British, Indian \newline Chinese: Beijing Mandarin, Northeastern Mandarin
\newline Korean: Seoul, Jeolla \\ 
\hline
Gender   & Male (Orus), Female (Zephyr) \\
\hline
Option Order & Original, Reversed \\
\bottomrule
\end{tabular}
\caption{Controlled variables used to generate speech-based MCQ inputs. Combining these factors yields up to 28 configurations per question.}
\label{tab:variables}
\end{table}
\section{Experimental Setup}
\subsection{Task and Variables}
\paragraph{Task Definition}
Our objective is to investigate the robustness of multimodal large language models (MLLMs) in spoken multiple-choice question (MCQ) tasks. Unlike conventional text-only evaluations, this setting requires models to process an audio input consisting of a question followed by answer options, and then select the correct choice. This formulation introduces a central challenge: models must not only comprehend the linguistic content but also maintain consistency when the same question is presented under varying speech conditions in realistic settings.

\paragraph{Experimental Variables}
To systematically examine robustness, we conduct our experiments on the \textbf{\textsc{BiasInEar}} benchmark introduced in Section~\ref{sec:dataset}.
Each question is instantiated under controlled perturbations spanning linguistic (\textit{language}, \textit{accent}), demographic (\textit{gender}), and structural (\textit{option order}) dimensions, as summarized in Table~\ref{tab:variables}. For \textit{gender} variation, we adopt the \textit{Orus} and \textit{Zephyr} voices from Gemini\footnote{\url{https://ai.google.dev/gemini-api/docs/speech-generation\#voices}}. For \textit{option order}, the \textit{original} setting represents the canonical sequence 
\texttt{A: \{Option A\}, B: \{Option B\}, C: \{Option C\}, D: \{Option D\}}, 
while the \textit{reversed} setting presents the sequence in reverse, 
\texttt{A: \{Option D\}, B: \{Option C\}, C: \{Option B\}, D: \{Option A\}}. 
By combining these factors, a single question can yield up to 28 distinct configurations, enabling evaluation not only of absolute accuracy but also of stability across diverse speech conditions.

\subsection{Models and Implementation}
\paragraph{Models}
\label{sec:model-main}
We evaluate nine MLLMs to assess their robustness under diverse experimental settings, including closed-weight models such as the Gemini family and open-source models such as the Gemma 3n, Voxtral, and Phi 4 families. Model details are provided in  Appendix~\ref{sec:appendix-model} due to space constraints. To ensure the stability and scalability of the experiments, we access the models through APIs provided by Google, NVIDIA, and Mistral.

\paragraph{Implementation Details}
The audio samples generated in Section~\ref{sec:dataset} consist of a question followed by its separate answer options. Before inputting them into the MLLM, we concatenate the respective audio segments according to the experimental condition (\textit{original} or \textit{reversed}). Details of the audio concatenation pipeline are provided in Appendix~\ref{appendix-audio-concat} for brevity. For model inference, we set the temperature to 0 to ensure reproducibility. The prompts used for standard and chain-of-thought (CoT) prompting are shown in Figures~\ref{fig:standard-prompt} and~\ref{fig:cot-prompt} in Appendix~\ref{sec:model-inference-and-post-processing}. Additionally, we apply post-processing to the model outputs to correct formatting errors, ensuring that our robustness analysis reflects genuine model behavior rather than artifacts from output format inconsistencies.

\subsection{Evaluation Metrics}
To evaluate robustness under input perturbations, we employ three complementary metrics: entropy, APES, and Fleiss’ Kappa. These measures go beyond accuracy by assessing not only correctness but also the stability and consistency of model behavior. Detailed definitions are provided below. At a high level, they address the following questions:
\begin{itemize}
    \item \textbf{Entropy}: \textit{Does the model’s answer distribution remain concentrated or become scattered across conditions?}
    \item \textbf{APES}: \textit{Does its confidence vary when input conditions change?}
    \item \textbf{Fleiss’ Kappa}: \textit{Does the final prediction stay consistent under perturbations?}
\end{itemize}

\begin{figure*}[t!]
\centering
\includegraphics[width=\linewidth]{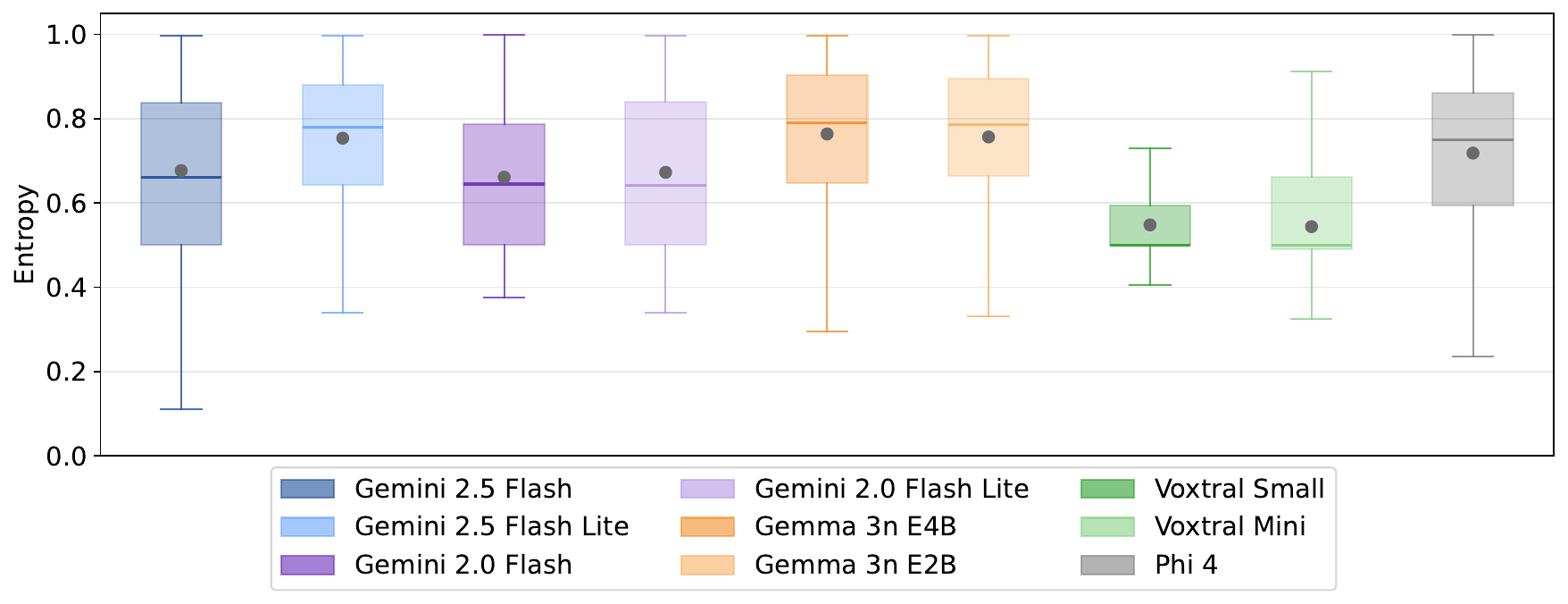}
\caption{Mean question entropy across models. Higher entropy indicates greater uncertainty in model predictions.}
\label{fig:mean_entropy}
\end{figure*}
\begin{figure*}[t!]
    \centering
    \includegraphics[width=\linewidth]{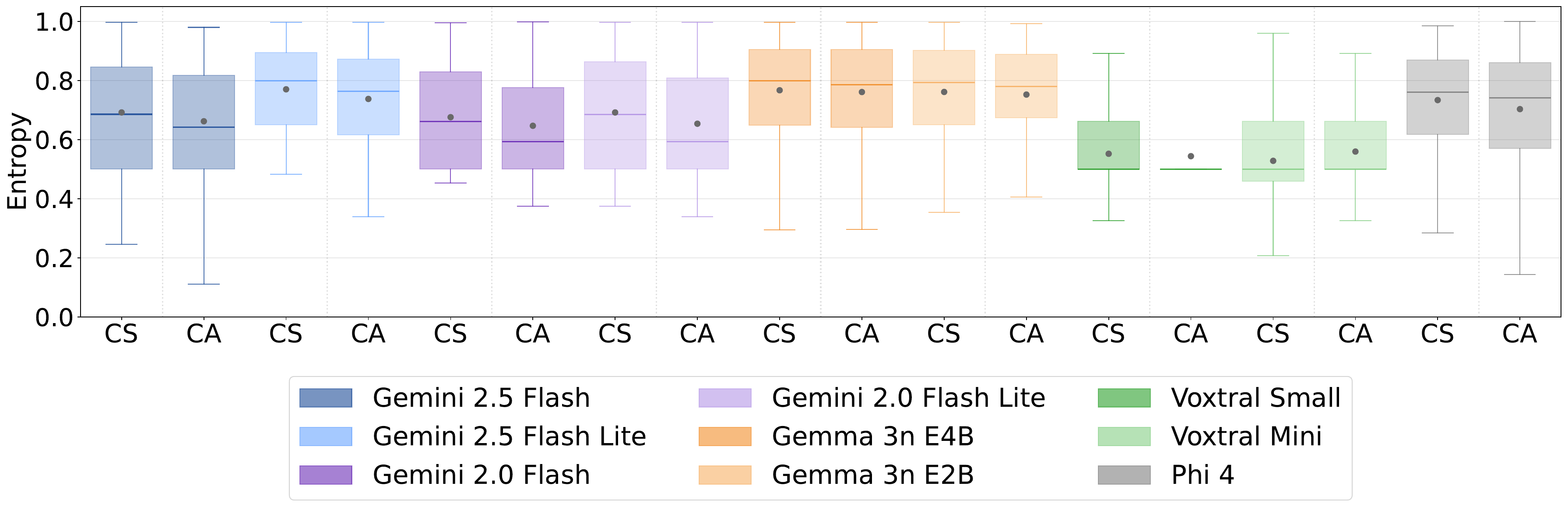}
    \caption{Entropy comparison between Culturally Sensitive (CS) and Culturally Agnostic (CA) questions.}
    \label{fig:CSCA}
\end{figure*}
\paragraph{Question Entropy.} 
For each question $q$, we compute the Shannon entropy~\cite{Shannon-entropy} of the model’s answer distribution:
\begin{equation}
H_q = - \sum_{o \in \{A, B, C, D\}} p_{q}(o) \log_4 p_{q}(o),
\end{equation}
where $p_{q}(o)$ is the probability assigned to option $o$. Normalization with base 4 ensures $H_q \in [0,1]$.  

\paragraph{Level Entropy and APES.} 
Given a variable $v$ with levels $\mathbf{L}_v$ (e.g., $\{\text{female}, \text{male}\}$), we compute entropy at each level $l$ as
\begin{equation}
H_q^l = - \sum_{o \in \{A,B,C,D\}} p_q(o|l)\log_4 p_q(o|l),
\end{equation}
where $p_q(o|l)$ denotes the probability assigned to option $o$ under level $l$.  

The Average Pairwise Entropy Shift (APES) quantifies entropy variation across levels:
\begin{equation}
\text{APES}_q^v = \frac{2}{L(L-1)} 
\sum_{\substack{l_i, l_j \in \mathbf{L}_v \\ i < j}} 
\big| H_q^{l_i} - H_q^{l_j} \big|,
\end{equation}
where $L = |\mathbf{L}_v|$, and $H_q^{l_i}$ be the entropy of $l_i \in \mathbf{L}_v$.

\paragraph{Fleiss' Kappa.} For each question $q$, we compute Fleiss’ $\kappa$ \cite{fleiss1971} to measure categorical agreement across variable perturbations while correcting for chance, defined as 
\begin{equation}
  \kappa = \frac{\bar{P} - P_e}{1 - P_e},  
\end{equation}
 where $\bar{P}$ is the average observed agreement and $P_e$ is the expected agreement. The detailed formulation is provided in Appendix~\ref{app:fleiss_kappa_derivation}. $\kappa \approx 1$ indicates strong consistency across conditions, $\kappa \approx 0$ suggests agreement no better than chance, and a negative $\kappa$ reflects systematic disagreement worse than random expectation.

 \begin{table*}[t]
\centering
\setlength{\tabcolsep}{3pt}
\small
\begin{tabular}{lcccccccccccc}
\toprule
\multirow{2}{*}{\textbf{Model}} 
& \multicolumn{3}{c}{\textbf{Accent}} 
& \multicolumn{3}{c}{\textbf{Language}}
& \multicolumn{3}{c}{\textbf{Option Order}} 
& \multicolumn{3}{c}{\textbf{Gender}} \\
\cmidrule(lr){2-4}
\cmidrule(lr){5-7}
\cmidrule(lr){8-10}
\cmidrule(lr){11-13}
& CS & CA & $\Delta$
& CS & CA & $\Delta$
& CS & CA & $\Delta$
& CS & CA & $\Delta$ \\
\midrule
Gemini 2.5 Flash
& \textbf{0.614} & 0.593 & \colorbox{green!30}{0.021}
& \textbf{0.648} & 0.623 & \colorbox{green!30}{0.025}
& \textbf{0.333} & 0.279 & \colorbox{green!30}{0.054}
& \textbf{0.678} & 0.650 & \colorbox{green!30}{0.028} \\
\midrule
Gemini 2.5 Flash Lite
& \textbf{0.680} & 0.647 & \colorbox{green!30}{0.033}
& \textbf{0.731} & 0.692 & \colorbox{green!30}{0.039}
& \textbf{0.542} & 0.459 & \colorbox{green!30}{0.083}
& \textbf{0.779} & 0.740 & \colorbox{green!30}{0.039} \\
\midrule
Gemma 3n E2B
& \textbf{0.468} & 0.461 & \colorbox{green!30}{0.007}
& \textbf{0.579} & 0.559 & \colorbox{green!30}{0.020}
& \textbf{0.604} & 0.578 & \colorbox{green!30}{0.026}
& \textbf{0.731} & 0.723 & \colorbox{green!30}{0.008} \\
\midrule
Gemma 3n E4B
& 0.490 & \textbf{0.498} & \colorbox{blue!30}{-0.008}
& 0.582 & \textbf{0.587} & \colorbox{blue!30}{-0.005}
& \textbf{0.571} & 0.540 & \colorbox{green!30}{0.031}
& \textbf{0.746} & 0.738 & \colorbox{green!30}{0.008} \\
\midrule
Phi 4
& 0.461 & \textbf{0.467} & \colorbox{blue!30}{-0.006}
& \textbf{0.566} & 0.549 & \colorbox{green!30}{0.017}
& \textbf{0.499} & 0.446 & \colorbox{green!30}{0.053}
& \textbf{0.697} & 0.669 & \colorbox{green!30}{0.028} \\
\bottomrule
\end{tabular}
\caption{Mean entropy of culturally sensitive (CS) vs. culturally agnostic (CA) questions across variables.}
\label{tab:6}
\end{table*}
 \begin{table*}[t]
\footnotesize
\centering
\setlength{\tabcolsep}{5pt}
\renewcommand{\arraystretch}{1.12}
\begin{tabular}{
>{\centering\arraybackslash}m{0.7cm} 
lccc  
lccc  
lccc  
}
\toprule
\multicolumn{1}{c}{} &
\multicolumn{4}{c}{Chinese} &
\multicolumn{4}{c}{English} &
\multicolumn{4}{c}{Korean} \\
\cmidrule(lr){2-5} \cmidrule(lr){6-9} \cmidrule(lr){10-13}
\multicolumn{1}{c}{Gender} &
Accent & Orig. & Rev. & $\Delta$ &
Accent & Orig. & Rev. & $\Delta$ &
Accent & Orig. & Rev. & $\Delta$ \\
\midrule
\multirow{3}{*}{\textcolor{blue}{\male}}
& Beijing       & \textbf{62.75} & 61.50 & \colorbox{green!30}{1.25}
& American      & \textbf{82.00} & 79.75 & \colorbox{green!30}{2.25}
& Jeolla        & \textbf{61.50} & 61.00 & \colorbox{green!30}{0.50} \\
& Northeastern  & \textbf{64.75} & 61.25 & \colorbox{green!30}{3.50}
& British       & \textbf{81.00} & 78.50 & \colorbox{green!30}{2.50}
& Seoul         & \textbf{63.00} & 62.00 & \colorbox{green!30}{1.00} \\
&               &                 &       &
& Indian        & \textbf{80.00} & 79.25 & \colorbox{green!30}{0.75}
&               &                 &       & \\[2pt]
\midrule
\multirow{3}{*}{\textcolor{magenta}{\female}}
& Beijing       & \textbf{66.75} & 64.00 & \colorbox{green!30}{2.75}
& American      & \textbf{80.00} & 78.50 & \colorbox{green!30}{1.50}
& Jeolla        & \textbf{63.50} & 58.50 & \colorbox{green!30}{5.00} \\
& Northeastern  & \textbf{64.50} & 57.75 & \colorbox{green!30}{6.75}
& British       & \textbf{81.25} & 76.75 & \colorbox{green!30}{4.50}
& Seoul         & \textbf{63.75} & 62.00 & \colorbox{green!30}{1.75} \\
&               &                 &       &
& Indian        & \textbf{80.50} & 78.25 & \colorbox{green!30}{2.25}
&               &                 &       & \\[-2pt]
\bottomrule
\end{tabular}
\caption{Accuracy comparison across \textit{option order} conditions for \textbf{Gemini 2.5 Flash}. 
Each cell reports the mean accuracy (\%) for the \textit{original} and \textit{reversed} option orders, 
with $\Delta$ denoting the difference between them. 
Results are grouped by \textit{language} (Chinese, English, Korean), \textit{accent}, and \textit{gender}.} 
\label{tab:accuracy-gemini-2.5-flash}
\end{table*}

\section{Investigation on Speech Bias}
\subsection{Overall Observation}
Figure~\ref{fig:mean_entropy} illustrates the overall entropy trends across the nine evaluated models. For each model, we compute per-question entropy across all configurations and then average the results over 400 questions per setting. The results reveal that the Gemini and Gemma families exhibit consistently higher entropy, indicating greater uncertainty in their answer distributions under diverse conditions. In contrast, the Voxtral family show lower entropy with narrower dispersion, reflecting more concentrated and confident predictions, whereas the Phi~4 model displays a larger interquartile range, suggesting greater variability in prediction confidence across questions. 
Within each model family, the lighter variants (e.g., \texttt{Voxtral Mini} vs. \texttt{Voxtral Small}) exhibit slightly higher mean entropy than their larger counterparts, suggesting that smaller parameter scales tend to produce less stable behavior and greater prediction uncertainty overall.

\subsection{Comparison Between CS and CA}
The Global MMLU Lite benchmark categorizes questions as Culturally Sensitive (CS), which require contextual or culture-specific knowledge, and Culturally Agnostic (CA), which rely primarily on domain knowledge.
Figure~\ref{fig:CSCA} presents entropy comparisons across nine models under CS and CA settings.
Overall, CA questions consistently exhibit lower entropy, indicating more concentrated and stable answer distributions aligned with factual reasoning, whereas CS questions display broader entropy ranges.
As shown in Table~\ref{tab:6}, this CS--CA entropy gap persists across variables: perturbations in \textit{accent} and \textit{gender} introduce only minor differences, while \textit{option order} produces the largest gap, suggesting higher positional sensitivity for culturally grounded items.
Complementary CS/CA robustness results under cross-variable perturbations are reported in Appendix~\ref{cscavariableperturbations}.
\subsection{Accuracy across Variable Levels}
We next perform a level-wise analysis to examine how different variable levels influence model behavior, beginning with accuracy. Table~\ref{tab:accuracy-gemini-2.5-flash} reports the performance of \texttt{Gemini~2.5~Flash} across Chinese, English, and Korean. The accuracy gap between \textit{option order} configurations ranges from 0.5\% to 6.75\%, with the \textit{original} order consistently outperforming the \textit{reversed} order across all language-accent settings. Results for other models, presented in Tables~\ref{tab:accuracy-gemini-2.5-flash-lite}-\ref{tab:accuracy-voxtral-mini-2507} in Appendix~\ref{sec:appendix-acc}, reveal a similar pattern: most models achieve higher accuracy under the \textit{original} configuration. These findings indicate that \textit{option order} introduces a systematic bias in model predictions. The effects of other variables
are also summarized in same Appendix.

\begin{figure*}[t!]
    \centering
    \includegraphics[width=\linewidth]{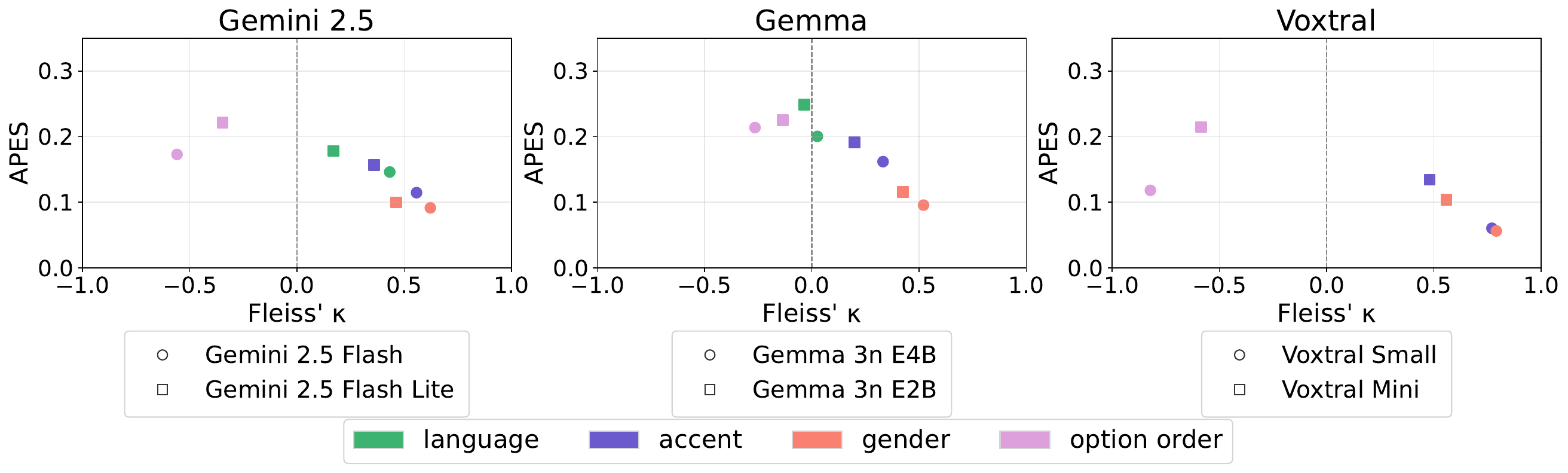}
    \caption{Fleiss' $\kappa$ versus APES across model families. Variables such as \textit{language}, \textit{accent}, and \textit{gender} show higher agreement and stability, while \textit{option order} yields higher APES and lower $\kappa$, indicating strong sensitivity.}
    \label{fig:kappashift}
\end{figure*}

\subsection{Robustness across Variable Levels}
\paragraph{Level-wise Analysis}
Beyond correctness, Figure~\ref{fig:kappashift} visualizes robustness patterns across model families. The \textit{gender} and \textit{accent} lie in the lower-right quadrant (high $\kappa$, low APES), indicating robust predictions characterized by strong within-level agreement and stable uncertainty across levels. In contrast, \textit{language} occupies intermediate regions ($\kappa \approx 0$, higher APES) with substantial variation across model families, suggesting that cross-lingual generalization remains a key robustness challenge. Note that Voxtral family do not support Chinese or Korean, and thus language-level analysis is omitted for this family. Finally, \textit{option order} consistently emerges as the weakest factor across all models, typically appearing in the left quadrants with negative $\kappa$ and relatively high APES, reflecting pronounced sensitivity to input order. 
Overall, while models exhibit relatively greater robustness to speaker-related factors (\textit{gender} and \textit{accent}), agreement under these perturbations only reaches the "Moderate" to "Substantial" range ($\kappa \approx 0.4$-$0.8$), rather than "Almost perfect" ($\kappa > 0.8$) that robust system ideally require. This gap indicates substantial room for improving speech robustness.

\begin{table}[t!]
\centering
\setlength{\tabcolsep}{3pt}
\footnotesize
\begin{tabular}{l c c c c}
\toprule
\textbf{Model} &
\multicolumn{2}{c}{\textbf{Accent}} &
\multicolumn{2}{c}{\textbf{Option Order}} \\
\cmidrule(lr){2-3}\cmidrule(lr){4-5}
& \textbf{Direct} & \textbf{Cloning} & \textbf{Direct} & \textbf{Cloning} \\
\midrule
Gemini 2.5 Flash Lite    & 0.100 & \textbf{0.096} & \textbf{0.155} & 0.194 \\
Gemini 2.5 Flash         & 0.060 & \textbf{0.053} & \textbf{0.101} & 0.132 \\
Gemma 3n E2B   & 0.125 & \textbf{0.027} & 0.228 & \textbf{0.215} \\
Gemma 3n E4B   & \textbf{0.096} & 0.098 & 0.179 & \textbf{0.164} \\
Phi 4 & \textbf{0.103} & 0.112 &0.208 &\textbf{0.173}\\
\bottomrule
\end{tabular}
\caption{{APES comparison across \textit{accent} and \textit{option order}. Lower values indicate higher robustness.}}
\label{tab:10}
\end{table}
\begin{figure}[t]
    \centering
    \includegraphics[width=\linewidth]{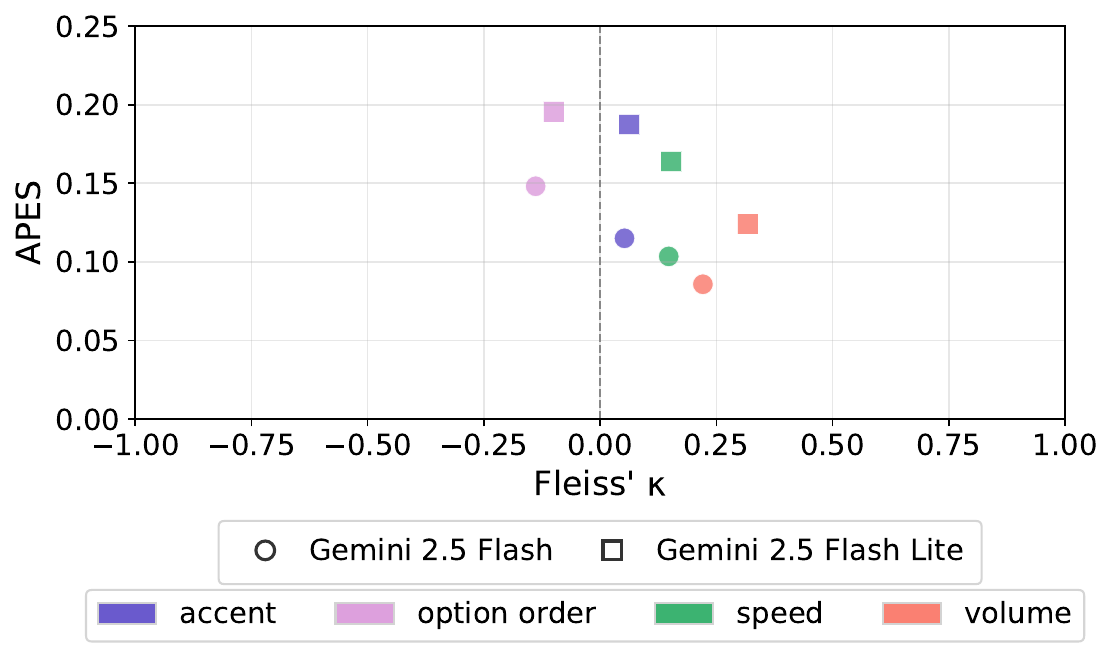}
    \caption{Fleiss' $\kappa$ versus APES across \textit{accent}, \textit{option order}, \textit{speed}, and volume variables.}
    \label{fig:speedvolume}
\end{figure}

\paragraph{Impact of Model Scale}
Figure~\ref{fig:kappashift} also compares model robustness across scales within three representative families. Results for Gemini 2.0 family are provided in Appendix ~\ref{appendix:model_scaling} for completeness. Larger models consistently demonstrate higher $\kappa$ and lower APES for the \textit{gender}, \textit{accent}, and \textit{language} variables, indicating more stable and consistent behavior under input perturbations. For \textit{option order}, larger models also achieve lower APES, although $\kappa$ remains negative, making direct comparison less meaningful. These results suggest that parameter reduction amplifies vulnerability to input perturbations, rendering smaller or lite variants less robust than their full-scale counterparts.

\paragraph{Option Order Variants}
Following ~\citet{wei-etal-2024-unveiling}, we evaluate whether \textit{option order} bias generalizes beyond a single reversal by applying multiple option permutations, including \textit{original}, \textit{fully reversed}, \textit{token-backward}, and \textit{order-backward}.
Results in Appendix~\ref{appendix:Option_Reordering} (Tables~\ref{tab:7}–\ref{tab:8}) show that option reordering has limited impact on APES across other variables, preserves the factor ranking (\textit{language} > \textit{accent} > \textit{gender}), and that fully \textit{reversed} orders induce the highest uncertainty, consistent with our main findings.

\begin{figure*}[ht]
    \centering
    \includegraphics[width=\linewidth]{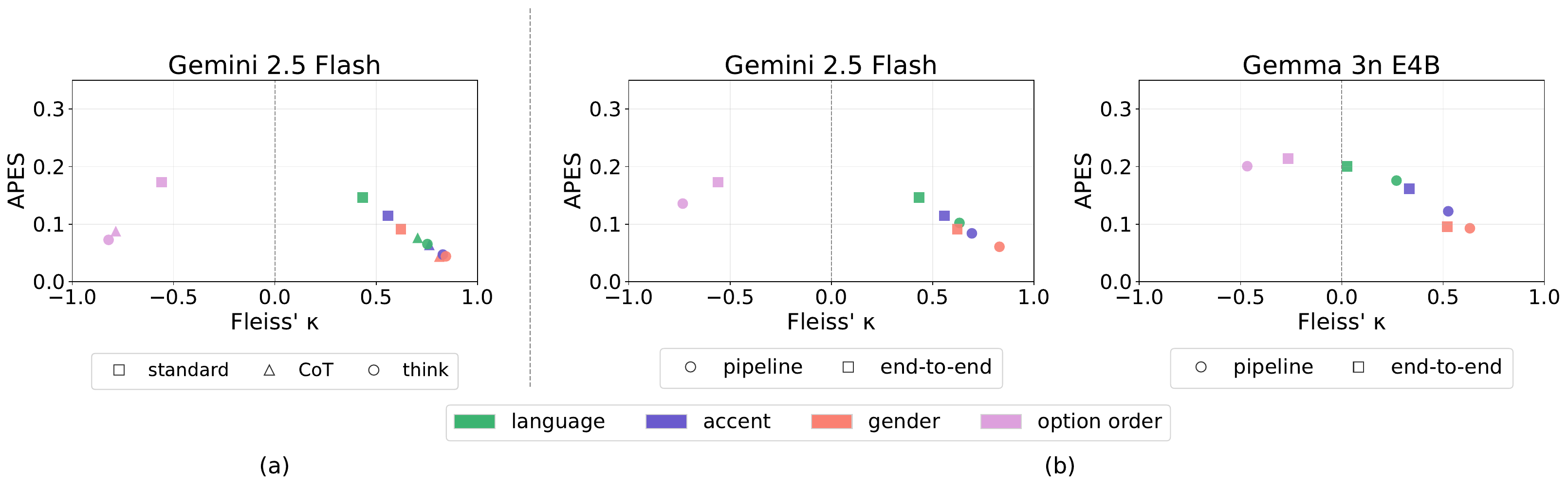}
\caption{ Effect of (a) reasoning complexity and (b) architectural paradigm on model robustness. Higher reasoning complexity and pipeline designs yield higher agreement (Fleiss’~$\kappa$) and lower uncertainty (APES).}
    \label{fig:reasoning-and-pipeline}
\end{figure*}

\section{Discussion}
\subsection{Real World Speaker Variability}
Most experiments in this work rely on TTS generated speech, which may raise concerns about whether the observed speech biases generalize to real world settings.
To address this, we conduct two complementary analyses to better approximate real world speaker variability and acoustic conditions.

\paragraph{Speaker Identity Realism}
To move beyond purely synthetic voices, we collect short recordings from three real speakers representing American, British, and Indian English accents, and use Chatterbox~\cite{chatterboxtts2025}, a neural voice cloning TTS model, to generate the full 400 question English Global MMLU Lite dataset for each accent.
As shown in Table~\ref{tab:10}, the core trends and relative model rankings remain consistent across these cloned voices.
Importantly, the bias patterns closely match those observed under direct TTS generation, suggesting that our findings are not artifacts of a specific synthetic voice, but persist under speaker characteristics closer to real world application conditions.

\paragraph{Acoustic Variability}
To further account for variability in recording conditions, we perturb the cloned English data with different speech rates (0.75$\times$, 1.0$\times$, and 1.25$\times$) and loudness levels (0.5$\times$, 1.0$\times$, and 1.5$\times$).
Results in Figure~\ref{fig:speedvolume} indicate that variations in speech rate introduce noticeably larger bias than changes in volume, as reflected by higher APES values across all models.
Despite these perturbations, the main conclusions of our study remain stable, reinforcing the robustness of the observed speech bias patterns under more realistic acoustic variability.


\subsection{Impact of Reasoning Complexity}
We examine how increasing reasoning complexity affects model robustness by comparing standard prompting, chain-of-thought (CoT) prompting~\cite{cot-jason-wei}, and explicit reasoning (thinking) modes.
Figure~\ref{fig:reasoning-and-pipeline}(a) reports results for \texttt{Gemini 2.5 Flash}.
Overall, CoT prompting substantially improves agreement, with Fleiss’~$\kappa$ increasing by an average of $19.01\%$, $20.50\%$, and $27.20\%$ for \textit{gender}, \textit{accent}, and \textit{language}, respectively.
It also yields greater robustness, reflected in mean APES reductions of $4.79\%$, $5.07\%$, $6.98\%$, and $8.50\%$ for \textit{gender}, \textit{accent}, \textit{language}, and \textit{option order}.
Further enabling explicit reasoning leads to additional gains in both agreement and robustness beyond CoT prompting alone.
These results suggest that increased reasoning complexity mitigates input induced variability and stabilizes model predictions under diverse perturbations.


\subsection{Impact of Architectural Paradigm}
We next examine the role of architectural paradigm, contrasting end-to-end multimodal LLMs with a pipeline design. While end-to-end models process audio directly, they may rely on ASR-like layers that filter out paralinguistic cues (e.g., \textit{accent}, \textit{gender}). To test this, we construct a pipeline setup where the model first transcribes the audio into text before answering. This comparison assesses whether explicit transcription removes speaker-dependent cues affecting robustness across conditions.
We apply this comparison to two representative models, \texttt{Gemini 2.5 Flash} and \texttt{Gemma 3n E4B}, selected for their strong multimodal input capabilities.
As shown in Figure~\ref{fig:reasoning-and-pipeline} (b), under the pipeline setting, both models exhibit higher Fleiss' $\kappa$ and lower APES across \textit{language}, \textit{accent}, and \textit{gender}, relative to their end-to-end counterparts. This pattern indicates that explicit transcription suppresses speaker-dependent variability, thereby mitigating accent- and gender-induced biases in the final predictions. Taken together, these results highlight architectural paradigm as a key lever for robustness, with the pipeline procedure reducing paralinguistic sensitivity and promoting more consistent behavior across conditions.


\begin{table}[th]
\centering
\setlength{\tabcolsep}{3pt}
\small
\begin{tabular}{lcccc}
\toprule
 & \multicolumn{2}{c}{\textbf{Language}} 
 & \multicolumn{2}{c}{\textbf{Option order}} \\
\textbf{model} 
& \textbf{Text} 
& \textbf{Audio} 
& \textbf{Text} 
& \textbf{Audio}\\
\midrule
Gemini 2.5 Flash Lite
& \textbf{0.081} & 0.178 &\textbf{0.096} & 0.221 \\
\midrule
Gemini 2.5 Flash
& \textbf{0.080} & 0.146 & \textbf{0.085} & 0.173 \\
\midrule
Gemma 3n E2B
& \textbf{0.163} & 0.249 & \textbf{0.194} & 0.225 \\
\midrule
Gemma 3n E4B
& \textbf{0.148} & 0.200 & \textbf{0.167} & 0.214 \\
\midrule
Phi 4 Multimodal
& \textbf{0.208}& 0.244 & \textbf{0.192} & 0.242 \\
\bottomrule
\end{tabular}
\caption{Comparison of APES between text and audio inputs across \textit{language} and \textit{option order}.
Audio inputs consistently yield higher APES, indicating amplification of existing robustness sensitivities.}
\label{tab:textvsaudio}
\end{table}

\subsection{Speech as a Bias Amplifier}
Before attributing the observed robustness differences to properties unique to speech, we examine whether these patterns reflect amplification of biases already present in text-based question answering.
We therefore compare text and audio inputs along two variables known to induce sensitivity, namely \textit{language} and \textit{option order}.
As shown in Table~\ref{tab:textvsaudio}, all models exhibit consistently higher APES values under the audio condition than under text.
This indicates that the robustness patterns observed in speech are not unique to the audio modality, but correspond to systematic amplification of existing biases when questions are presented in spoken form.
This cross-modal comparison serves as a sanity check that grounds our analysis, supporting the interpretation that speech primarily magnifies sensitivities already present in text-based models rather than introducing qualitatively new bias patterns.


\section{Related Work}
\paragraph{Speech Bias in ASR.}
Prior work on speech bias has primarily focused on automatic speech recognition (ASR) systems, which consistently exhibit performance disparities across gender, accent, and language.
Gender related biases have been widely reported, with higher word error rates (WER) observed for female speakers in YouTube auto captions~\cite{tatman-2017-gender}, male speech in Whisper Small~\cite{elghazaly2025exploringgenderdisparitiesautomatic}, and model dependent reversals across systems~\cite{10.1121/10.0024876}.
Accent bias is similarly pervasive, with American and Canadian English yielding lower WERs than non native accents~\cite{10.1121/10.0024876}, and substantial variation across regional accents in datasets such as SQuAD SRC~\cite{ijcai2023p578}.
In multilingual settings, high resource languages consistently outperform low resource or tonal languages, as shown by higher WERs for Chinese and Korean in Meta’s XLS R model~\cite{babu22_interspeech}.
Overall, these findings attribute ASR bias to the combined effects of gender, accent, and data imbalance.
Our work extends this line of research beyond transcription accuracy to examine \textbf{speech bias in multilingual MLLMs}, enabling a unified evaluation across linguistic and demographic dimensions.

\paragraph{LLM Robustness.}
Recent studies document systematic biases in LLMs across demographic and structural dimensions.
Gender bias persists even without explicit markers~\cite{ICLR2024_37771cc0}, with models exhibiting systematic disadvantages against women~\cite{vo2025bscore}, while racial and dialectal biases include covert negative stereotypes toward African American English~\cite{hofmann2024dialectprejudicepredictsai} and broader racial, national, and religious stereotypes under complex reasoning settings~\cite{shrawgi-etal-2024-uncovering}.
Cultural bias further arises from the dominance of Western centric training data, leading to disparities in multilingual and multicultural contexts~\cite{li2024culturepark}.
Beyond demographic factors, \citet{wei-etal-2024-unveiling} identify selection bias, a structural sensitivity to non semantic cues such as option order or symbolic formatting in multiple choice questions.
Motivated by these findings, our work extends the study of selection bias from text based evaluations to the speech modality, examining positional sensitivities under spoken inputs.

\section{Conclusion}
This work presents the first systematic study of \textbf{speech bias and robustness} in multilingual MLLMs.
We introduce \textbf{\textsc{BiasInEar}}, a speech-augmented benchmark built on Global MMLU Lite, covering English, Chinese, and Korean, balanced across \textit{gender} and \textit{accent}, and comprising 11,200 questions with 70.8 hours ($\approx$4,249 minutes) of speech.
Using four complementary metrics (accuracy, entropy, APES, and Fleiss'~$\kappa$), we evaluate nine representative models from four model families and analyze robustness across linguistic, demographic, and structural factors.
Our results show that \textit{option order} induces the most pronounced robustness degradation, while \textit{accent} and \textit{gender} lead to smaller but consistent confidence shifts.
We further demonstrate that increased reasoning complexity and pipeline-based architectural designs improve robustness, and that speech systematically amplifies biases already present in text-based settings.
Together, these findings reveal underexplored vulnerabilities in current MLLMs and offer practical insights for designing fairer and more stable speech-integrated AI systems.

\section*{Limitations}
\paragraph{Voice Generation}
Although our dataset systematically controls for \textit{language}, \textit{accent}, and \textit{gender}, the use of text-to-speech (TTS) for automated audio generation introduces inherent challenges in defining and standardizing “accent.” Even within a single \textit{language}, \textit{accent} variation often exists on a continuous spectrum rather than as discrete categories. The boundaries between regional or social varieties are fuzzy and, in some cases, linguistically indeterminate, making it difficult to ensure that our current setup fully captures the natural and continuous variation present in human speech.
Furthermore, due to computational constraints, we were unable to synthesize a larger number of voice variants for each condition. Nevertheless, because our dataset is derived from Global-MMLU-Lite, which covers a broad range of topics and languages, we believe that our results remain representative and robust in capturing overall cross-linguistic and paralinguistic trends.
\paragraph{Evaluated Models}
Due to computational and API interface constraints, this study evaluated only nine representative multimodal large language models (MLLMs) spanning both commercial and open-source categories. While these models capture diversity in architecture and reasoning pipelines, they do not fully cover the spectrum of existing systems. Some open-source models were excluded due to limited stability, insufficient scalability for large-scale inference, or the lack of a publicly available, stable, and efficient multimodal API. Future work could broaden the evaluation as standardized interfaces and reproducible deployment pipelines mature, enabling a more comprehensive assessment of cross-model consistency and generalization.

\section*{Use of AI Assistants}
We used ChatGPT as an assistant to refine the manuscript, improve clarity, and enhance the structure and readability. While the final content remains entirely our own, this assistance helped improve the overall presentation of our work.

\section*{Acknowledgements}
This work was supported by National Science and Technology Council, Taiwan, under grant NSTC 114-2221-E-002 -070 -MY3, NSTC 113-2634-F-002-003 -, and Ministry of Education (MOE) in Taiwan under grants NTU-114L900901.

\bibliography{custom}

@misc{chatterboxtts2025,
  author       = {{Resemble AI}},
  title        = {{Chatterbox-TTS}},
  year         = {2025},
  howpublished = {\url{https://github.com/resemble-ai/chatterbox}},
  note         = {GitHub repository}
}

@misc{omnilingualasrteam2025omnilingualasropensourcemultilingual,
      title={Omnilingual ASR: Open-Source Multilingual Speech Recognition for 1600+ Languages}, 
      author={Omnilingual ASR team and Gil Keren and Artyom Kozhevnikov and Yen Meng and Christophe Ropers and Matthew Setzler and Skyler Wang and Ife Adebara and Michael Auli and Can Balioglu and Kevin Chan and Chierh Cheng and Joe Chuang and Caley Droof and Mark Duppenthaler and Paul-Ambroise Duquenne and Alexander Erben and Cynthia Gao and Gabriel Mejia Gonzalez and Kehan Lyu and Sagar Miglani and Vineel Pratap and Kaushik Ram Sadagopan and Safiyyah Saleem and Arina Turkatenko and Albert Ventayol-Boada and Zheng-Xin Yong and Yu-An Chung and Jean Maillard and Rashel Moritz and Alexandre Mourachko and Mary Williamson and Shireen Yates},
      year={2025},
      eprint={2511.09690},
      archivePrefix={arXiv},
      primaryClass={cs.CL},
      url={https://arxiv.org/abs/2511.09690}, 
}

@misc{radford2022whisper,
  doi = {10.48550/ARXIV.2212.04356},
  url = {https://arxiv.org/abs/2212.04356},
  author = {Radford, Alec and Kim, Jong Wook and Xu, Tao and Brockman, Greg and McLeavey, Christine and Sutskever, Ilya},
  title = {Robust Speech Recognition via Large-Scale Weak Supervision},
  publisher = {arXiv},
  year = {2022},
  copyright = {arXiv.org perpetual, non-exclusive license}
}

@misc{OpenAIChatGPT,
  author = {{OpenAI}},
  title = {{Introducing ChatGPT}},
  year = {2022},
  howpublished = {\url{https://openai.com/blog/chatgpt}}
}

@article{Anil2023GeminiAF,
  title={Gemini: A Family of Highly Capable Multimodal Models},
  author={Gemini Team, Google},
  journal={ArXiv},
  year={2023},
  volume={abs/2312.11805},
  url={https://arxiv.org/pdf/2312.11805}
}

@misc{anthropic2025claude37,
  author       = {Anthropic},
  title        = {Claude 3.7 Sonnet and Claude Code},
  year         = {2025},
  month        = feb,
  day          = {24},
  url          = {https://www.anthropic.com/news/claude-3-7-sonnet},
  note         = {Accessed: 2025-09-29}
}

@misc{openai_gpt-4o_2024,
    title = {{GPT}-4o {System} {Card}},
    url = {http://arxiv.org/abs/2410.21276},
    doi = {10.48550/arXiv.2410.21276},
    publisher = {arXiv},
    author = {OpenAI},
    year = {2024},
    note = {arXiv:2410.21276},
}

@misc{agrawal2024pixtral12b,
      title={Pixtral 12B}, 
      author={Pravesh Agrawal and Szymon Antoniak and Emma Bou Hanna and Baptiste Bout and Devendra Chaplot and Jessica Chudnovsky and Diogo Costa and Baudouin De Monicault and Saurabh Garg and Theophile Gervet and Soham Ghosh and Amélie Héliou and Paul Jacob and Albert Q. Jiang and Kartik Khandelwal and Timothée Lacroix and Guillaume Lample and Diego Las Casas and Thibaut Lavril and Teven Le Scao and Andy Lo and William Marshall and Louis Martin and Arthur Mensch and Pavankumar Muddireddy and Valera Nemychnikova and Marie Pellat and Patrick Von Platen and Nikhil Raghuraman and Baptiste Rozière and Alexandre Sablayrolles and Lucile Saulnier and Romain Sauvestre and Wendy Shang and Roman Soletskyi and Lawrence Stewart and Pierre Stock and Joachim Studnia and Sandeep Subramanian and Sagar Vaze and Thomas Wang and Sophia Yang},
      year={2024},
      eprint={2410.07073},
      archivePrefix={arXiv},
      primaryClass={cs.CV},
      url={https://arxiv.org/abs/2410.07073}, 
}

@misc{meta2025llama4,
  title        = {The Llama 4 herd: The beginning of a new era of natively multimodal intelligence},
  author       = {{Meta AI}},
  year         = {2025},
  url          = {https://ai.meta.com/blog/llama-4-multimodal-intelligence/}
}

@misc{comanici2025gemini25pushingfrontier,
      title={Gemini 2.5: Pushing the Frontier with Advanced Reasoning, Multimodality, Long Context, and Next Generation Agentic Capabilities}, 
      author={Gheorghe Comanici and Eric Bieber and Mike Schaekermann and Ice Pasupat and Noveen Sachdeva and Inderjit Dhillon and Marcel Blistein and Ori Ram and Dan Zhang and Evan Rosen and Luke Marris and Sam Petulla and Colin Gaffney and Asaf Aharoni and Nathan Lintz and Tiago Cardal Pais and Henrik Jacobsson and Idan Szpektor and Nan-Jiang Jiang and Krishna Haridasan and Ahmed Omran and Nikunj Saunshi and Dara Bahri and Gaurav Mishra and Eric Chu and Toby Boyd and Brad Hekman and Aaron Parisi and Chaoyi Zhang and Kornraphop Kawintiranon and Tania Bedrax-Weiss and Oliver Wang and Ya Xu and Ollie Purkiss and Uri Mendlovic and Ilaï Deutel and Nam Nguyen and Adam Langley and Flip Korn and Lucia Rossazza and Alexandre Ramé and Sagar Waghmare and Helen Miller and Nathan Byrd and Ashrith Sheshan and Raia Hadsell and Sangnie Bhardwaj and Pawel Janus and Tero Rissa and Dan Horgan and Alvin Abdagic and Lior Belenki and James Allingham and Anima Singh and Theo Guidroz and Srivatsan Srinivasan and Herman Schmit and Kristen Chiafullo and Andre Elisseeff and Nilpa Jha and Prateek Kolhar and Leonard Berrada and Frank Ding and Xiance Si and Shrestha Basu Mallick and Franz Och and Sofia Erell and Eric Ni and Tejasi Latkar and Sherry Yang and Petar Sirkovic and Ziqiang Feng and Robert Leland and Rachel Hornung and Gang Wu and Charles Blundell and Hamidreza Alvari and Po-Sen Huang and Cathy Yip and Sanja Deur and Li Liu and Gabriela Surita and Pablo Duque and Dima Damen and Johnson Jia and Arthur Guez and Markus Mircea and Animesh Sinha and Alberto Magni and Paweł Stradomski and Tal Marian and Vlado Galić and Wenhu Chen and Hisham Husain and Achintya Singhal and Dominik Grewe and François-Xavier Aubet and Shuang Song and Lorenzo Blanco and Leland Rechis and Lewis Ho and Rich Munoz and Kelvin Zheng and Jessica Hamrick and Kevin Mather and Hagai Taitelbaum and Eliza Rutherford and Yun Lei and Kuangyuan Chen and Anand Shukla and Erica Moreira and Eric Doi and Berivan Isik and Nir Shabat and Dominika Rogozińska and Kashyap Kolipaka and Jason Chang and Eugen Vušak and Srinivasan Venkatachary and Shadi Noghabi and Tarun Bharti and Younghoon Jun and Aleksandr Zaks and Simon Green and Jeshwanth Challagundla and William Wong and Muqthar Mohammad and Dean Hirsch and Yong Cheng and Iftekhar Naim and Lev Proleev and Damien Vincent and Aayush Singh and Maxim Krikun and Dilip Krishnan and Zoubin Ghahramani and Aviel Atias and Rajeev Aggarwal and Christo Kirov and Dimitrios Vytiniotis and Christy Koh and Alexandra Chronopoulou and Pawan Dogra and Vlad-Doru Ion and Gladys Tyen and Jason Lee and Felix Weissenberger and Trevor Strohman and Ashwin Balakrishna and Jack Rae and Marko Velic and Raoul de Liedekerke and Oded Elyada and Wentao Yuan and Canoee Liu and Lior Shani and Sergey Kishchenko and Bea Alessio and Yandong Li and Richard Song and Sam Kwei and Orion Jankowski and Aneesh Pappu and Youhei Namiki and Yenai Ma and Nilesh Tripuraneni and Colin Cherry and Marissa Ikonomidis and Yu-Cheng Ling and Colin Ji and Beka Westberg and Auriel Wright and Da Yu and David Parkinson and Swaroop Ramaswamy and Jerome Connor and Soheil Hassas Yeganeh and Snchit Grover and George Kenwright and Lubo Litchev and Chris Apps and Alex Tomala and Felix Halim and Alex Castro-Ros and Zefei Li and Anudhyan Boral and Pauline Sho and Michal Yarom and Eric Malmi and David Klinghoffer and Rebecca Lin and Alan Ansell and Pradeep Kumar S and Shubin Zhao and Siqi Zuo and Adam Santoro and Heng-Tze Cheng and Solomon Demmessie and Yuchi Liu and Nicole Brichtova and Allie Culp and Nathaniel Braun and Dan Graur and Will Ng and Nikhil Mehta and Aaron Phillips and Patrik Sundberg and Varun Godbole and Fangyu Liu and Yash Katariya and David Rim and Mojtaba Seyedhosseini and Sean Ammirati and Jonas Valfridsson and Mahan Malihi and Timothy Knight and Andeep Toor and Thomas Lampe and Abe Ittycheriah and Lewis Chiang and Chak Yeung and Alexandre Fréchette and Jinmeng Rao and Huisheng Wang and Himanshu Srivastava and Richard Zhang and Rocky Rhodes and Ariel Brand and Dean Weesner and Ilya Figotin and Felix Gimeno and Rachana Fellinger and Pierre Marcenac and José Leal and Eyal Marcus and Victor Cotruta and Rodrigo Cabrera and Sheryl Luo and Dan Garrette and Vera Axelrod and Sorin Baltateanu and David Barker and Dongkai Chen and Horia Toma and Ben Ingram and Jason Riesa and Chinmay Kulkarni and Yujing Zhang and Hongbin Liu and Chao Wang and Martin Polacek and Will Wu and Kai Hui and Adrian N Reyes and Yi Su and Megan Barnes and Ishaan Malhi and Anfal Siddiqui and Qixuan Feng and Mihai Damaschin and Daniele Pighin and Andreas Steiner and Samuel Yang and Ramya Sree Boppana and Simeon Ivanov and Arun Kandoor and Aditya Shah and Asier Mujika and Da Huang and Christopher A. Choquette-Choo and Mohak Patel and Tianhe Yu and Toni Creswell and Jerry and Liu and Catarina Barros and Yasaman Razeghi and Aurko Roy and Phil Culliton and Binbin Xiong and Jiaqi Pan and Thomas Strohmann and Tolly Powell and Babi Seal and Doug DeCarlo and Pranav Shyam and Kaan Katircioglu and Xuezhi Wang and Cassidy Hardin and Immanuel Odisho and Josef Broder and Oscar Chang and Arun Nair and Artem Shtefan and Maura O'Brien and Manu Agarwal and Sahitya Potluri and Siddharth Goyal and Amit Jhindal and Saksham Thakur and Yury Stuken and James Lyon and Kristina Toutanova and Fangxiaoyu Feng and Austin Wu and Ben Horn and Alek Wang and Alex Cullum and Gabe Taubman and Disha Shrivastava and Chongyang Shi and Hamish Tomlinson and Roma Patel and Tao Tu and Ada Maksutaj Oflazer and Francesco Pongetti and Mingyao Yang and Adrien Ali Taïga and Vincent Perot and Nuo Wang Pierse and Feng Han and Yoel Drori and Iñaki Iturrate and Ayan Chakrabarti and Legg Yeung and Dave Dopson and Yi-ting Chen and Apoorv Kulshreshtha and Tongfei Guo and Philip Pham and Tal Schuster and Junquan Chen and Alex Polozov and Jinwei Xing and Huanjie Zhou and Praneeth Kacham and Doron Kukliansky and Antoine Miech and Sergey Yaroshenko and Ed Chi and Sholto Douglas and Hongliang Fei and Mathieu Blondel and Preethi Myla and Lior Madmoni and Xing Wu and Daniel Keysers and Kristian Kjems and Isabela Albuquerque and Lijun Yu and Joel D'sa and Michelle Plantan and Vlad Ionescu and Jaume Sanchez Elias and Abhirut Gupta and Manish Reddy Vuyyuru and Fred Alcober and Tong Zhou and Kaiyang Ji and Florian Hartmann and Subha Puttagunta and Hugo Song and Ehsan Amid and Anca Stefanoiu and Andrew Lee and Paul Pucciarelli and Emma Wang and Amit Raul and Slav Petrov and Isaac Tian and Valentin Anklin and Nana Nti and Victor Gomes and Max Schumacher and Grace Vesom and Alex Panagopoulos and Konstantinos Bousmalis and Daniel Andor and Josh Jacob and Yuan Zhang and Bill Rosgen and Matija Kecman and Matthew Tung and Alexandra Belias and Noah Goodman and Paul Covington and Brian Wieder and Nikita Saxena and Elnaz Davoodi and Muhuan Huang and Sharath Maddineni and Vincent Roulet and Folawiyo Campbell-Ajala and Pier Giuseppe Sessa and Xintian and Wu and Guangda Lai and Paul Collins and Alex Haig and Vytenis Sakenas and Xiaowei Xu and Marissa Giustina and Laurent El Shafey and Pichi Charoenpanit and Shefali Garg and Joshua Ainslie and Boone Severson and Montse Gonzalez Arenas and Shreya Pathak and Sujee Rajayogam and Jie Feng and Michiel Bakker and Sheng Li and Nevan Wichers and Jamie Rogers and Xinyang Geng and Yeqing Li and Rolf Jagerman and Chao Jia and Nadav Olmert and David Sharon and Matthew Mauger and Sandeep Mariserla and Hongxu Ma and Megha Mohabey and Kyuyeun Kim and Alek Andreev and Scott Pollom and Juliette Love and Vihan Jain and Priyanka Agrawal and Yannick Schroecker and Alisa Fortin and Manfred Warmuth and Ji Liu and Andrew Leach and Irina Blok and Ganesh Poomal Girirajan and Roee Aharoni and Benigno Uria and Andrei Sozanschi and Dan Goldberg and Lucian Ionita and Marco Tulio Ribeiro and Martin Zlocha and Vighnesh Birodkar and Sami Lachgar and Liangzhe Yuan and Himadri Choudhury and Matt Ginsberg and Fei Zheng and Gregory Dibb and Emily Graves and Swachhand Lokhande and Gabriel Rasskin and George-Cristian Muraru and Corbin Quick and Sandeep Tata and Pierre Sermanet and Aditya Chawla and Itay Karo and Yan Wang and Susan Zhang and Orgad Keller and Anca Dragan and Guolong Su and Ian Chou and Xi Liu and Yiqing Tao and Shruthi Prabhakara and Marc Wilson and Ruibo Liu and Shibo Wang and Georgie Evans and David Du and Alfonso Castaño and Gautam Prasad and Mona El Mahdy and Sebastian Gerlach and Machel Reid and Jarrod Kahn and Amir Zait and Thanumalayan Sankaranarayana Pillai and Thatcher Ulrich and Guanyu Wang and Jan Wassenberg and Efrat Farkash and Kiran Yalasangi and Congchao Wang and Maria Bauza and Simon Bucher and Ting Liu and Jun Yan and Gary Leung and Vikas Sindhwani and Parker Barnes and Avi Singh and Ivan Jurin and Jichuan Chang and Niket Kumar Bhumihar and Sivan Eiger and Gui Citovsky and Ben Withbroe and Zhang Li and Siyang Xue and Niccolò Dal Santo and Georgi Stoyanov and Yves Raimond and Steven Zheng and Yilin Gao and Vít Listík and Sławek Kwasiborski and Rachel Saputro and Adnan Ozturel and Ganesh Mallya and Kushal Majmundar and Ross West and Paul Caron and Jinliang Wei and Lluis Castrejon and Sharad Vikram and Deepak Ramachandran and Nikhil Dhawan and Jiho Park and Sara Smoot and George van den Driessche and Yochai Blau and Chase Malik and Wei Liang and Roy Hirsch and Cicero Nogueira dos Santos and Eugene Weinstein and Aäron van den Oord and Sid Lall and Nicholas FitzGerald and Zixuan Jiang and Xuan Yang and Dale Webster and Ali Elqursh and Aedan Pope and Georges Rotival and David Raposo and Wanzheng Zhu and Jeff Dean and Sami Alabed and Dustin Tran and Arushi Gupta and Zach Gleicher and Jessica Austin and Edouard Rosseel and Megh Umekar and Dipanjan Das and Yinghao Sun and Kai Chen and Karolis Misiunas and Xiang Zhou and Yixian Di and Alyssa Loo and Josh Newlan and Bo Li and Vinay Ramasesh and Ying Xu and Alex Chen and Sudeep Gandhe and Radu Soricut and Nikita Gupta and Shuguang Hu and Seliem El-Sayed and Xavier Garcia and Idan Brusilovsky and Pu-Chin Chen and Andrew Bolt and Lu Huang and Alex Gurney and Zhiying Zhang and Alexander Pritzel and Jarek Wilkiewicz and Bryan Seybold and Bhargav Kanagal Shamanna and Felix Fischer and Josef Dean and Karan Gill and Ross Mcilroy and Abhishek Bhowmick and Jeremy Selier and Antoine Yang and Derek Cheng and Vladimir Magay and Jie Tan and Dhriti Varma and Christian Walder and Tomas Kocisky and Ryo Nakashima and Paul Natsev and Mike Kwong and Ionel Gog and Chiyuan Zhang and Sander Dieleman and Thomas Jimma and Andrey Ryabtsev and Siddhartha Brahma and David Steiner and Dayou Du and Ante Žužul and Mislav Žanić and Mukund Raghavachari and Willi Gierke and Zeyu Zheng and Dessie Petrova and Yann Dauphin and Yuchuan Liu and Ido Kessler and Steven Hand and Chris Duvarney and Seokhwan Kim and Hyo Lee and Léonard Hussenot and Jeffrey Hui and Josh Smith and Deepali Jain and Jiawei Xia and Gaurav Singh Tomar and Keyvan Amiri and Du Phan and Fabian Fuchs and Tobias Weyand and Nenad Tomasev and Alexandra Cordell and Xin Liu and Jonathan Mallinson and Pankaj Joshi and Andy Crawford and Arun Suggala and Steve Chien and Nick Fernando and Mariella Sanchez-Vargas and Duncan Williams and Phil Crone and Xiyang Luo and Igor Karpov and Jyn Shan and Terry Thurk and Robin Strudel and Paul Voigtlaender and Piyush Patil and Tim Dozat and Ali Khodaei and Sahil Singla and Piotr Ambroszczyk and Qiyin Wu and Yifan Chang and Brian Roark and Chaitra Hegde and Tianli Ding and Angelos Filos and Zhongru Wu and André Susano Pinto and Shuang Liu and Saarthak Khanna and Aditya Pandey and Siobhan Mcloughlin and Qiujia Li and Sam Haves and Allan Zhou and Elena Buchatskaya and Isabel Leal and Peter de Boursac and Nami Akazawa and Nina Anderson and Terry Chen and Krishna Somandepalli and Chen Liang and Sheela Goenka and Stephanie Winkler and Alexander Grushetsky and Yifan Ding and Jamie Smith and Fan Ye and Jordi Pont-Tuset and Eric Li and Ruichao Li and Tomer Golany and Dawid Wegner and Tao Jiang and Omer Barak and Yuan Shangguan and Eszter Vértes and Renee Wong and Jörg Bornschein and Alex Tudor and Michele Bevilacqua and Tom Schaul and Ankit Singh Rawat and Yang Zhao and Kyriakos Axiotis and Lei Meng and Cory McLean and Jonathan Lai and Jennifer Beattie and Nate Kushman and Yaxin Liu and Blair Kutzman and Fiona Lang and Jingchen Ye and Praneeth Netrapalli and Pushkar Mishra and Myriam Khan and Megha Goel and Rob Willoughby and David Tian and Honglei Zhuang and JD Chen and Zak Tsai and Tasos Kementsietsidis and Arjun Khare and James Keeling and Keyang Xu and Nathan Waters and Florent Altché and Ashok Popat and Bhavishya Mittal and David Saxton and Dalia El Badawy and Michael Mathieu and Zheng Zheng and Hao Zhou and Nishant Ranka and Richard Shin and Qingnan Duan and Tim Salimans and Ioana Mihailescu and Uri Shaham and Ming-Wei Chang and Yannis Assael and Nishanth Dikkala and Martin Izzard and Vincent Cohen-Addad and Cat Graves and Vlad Feinberg and Grace Chung and DJ Strouse and Danny Karmon and Sahand Sharifzadeh and Zoe Ashwood and Khiem Pham and Jon Blanton and Alex Vasiloff and Jarred Barber and Mark Geller and Aurick Zhou and Fedir Zubach and Tzu-Kuo Huang and Lei Zhang and Himanshu Gupta and Matt Young and Julia Proskurnia and Ronny Votel and Valentin Gabeur and Gabriel Barcik and Aditya Tripathi and Hongkun Yu and Geng Yan and Beer Changpinyo and Filip Pavetić and Amy Coyle and Yasuhisa Fujii and Jorge Gonzalez Mendez and Tianhao Zhou and Harish Rajamani and Blake Hechtman and Eddie Cao and Da-Cheng Juan and Yi-Xuan Tan and Valentin Dalibard and Yilun Du and Natalie Clay and Kaisheng Yao and Wenhao Jia and Dimple Vijaykumar and Yuxiang Zhou and Xinyi Bai and Wei-Chih Hung and Steven Pecht and Georgi Todorov and Nikhil Khadke and Pramod Gupta and Preethi Lahoti and Arnaud Autef and Karthik Duddu and James Lee-Thorp and Alexander Bykovsky and Tautvydas Misiunas and Sebastian Flennerhag and Santhosh Thangaraj and Jed McGiffin and Zack Nado and Markus Kunesch and Andreas Noever and Amir Hertz and Marco Liang and Victor Stone and Evan Palmer and Samira Daruki and Arijit Pramanik and Siim Põder and Austin Kyker and Mina Khan and Evgeny Sluzhaev and Marvin Ritter and Avraham Ruderman and Wenlei Zhou and Chirag Nagpal and Kiran Vodrahalli and George Necula and Paul Barham and Ellie Pavlick and Jay Hartford and Izhak Shafran and Long Zhao and Maciej Mikuła and Tom Eccles and Hidetoshi Shimokawa and Kanav Garg and Luke Vilnis and Hanwen Chen and Ilia Shumailov and Kuang-Huei Lee and Abdelrahman Abdelhamed and Meiyan Xie and Vered Cohen and Ester Hlavnova and Dan Malkin and Chawin Sitawarin and James Lottes and Pauline Coquinot and Tianli Yu and Sandeep Kumar and Jingwei Zhang and Aroma Mahendru and Zafarali Ahmed and James Martens and Tao Chen and Aviel Boag and Daiyi Peng and Coline Devin and Arseniy Klimovskiy and Mary Phuong and Danny Vainstein and Jin Xie and Bhuvana Ramabhadran and Nathan Howard and Xinxin Yu and Gitartha Goswami and Jingyu Cui and Sam Shleifer and Mario Pinto and Chih-Kuan Yeh and Ming-Hsuan Yang and Sara Javanmardi and Dan Ethier and Chace Lee and Jordi Orbay and Suyog Kotecha and Carla Bromberg and Pete Shaw and James Thornton and Adi Gerzi Rosenthal and Shane Gu and Matt Thomas and Ian Gemp and Aditya Ayyar and Asahi Ushio and Aarush Selvan and Joel Wee and Chenxi Liu and Maryam Majzoubi and Weiren Yu and Jake Abernethy and Tyler Liechty and Renke Pan and Hoang Nguyen and Qiong and Hu and Sarah Perrin and Abhinav Arora and Emily Pitler and Weiyi Wang and Kaushik Shivakumar and Flavien Prost and Ben Limonchik and Jing Wang and Yi Gao and Timothee Cour and Shyamal Buch and Huan Gui and Maria Ivanova and Philipp Neubeck and Kelvin Chan and Lucy Kim and Huizhong Chen and Naman Goyal and Da-Woon Chung and Lu Liu and Yao Su and Anastasia Petrushkina and Jiajun Shen and Armand Joulin and Yuanzhong Xu and Stein Xudong Lin and Yana Kulizhskaya and Ciprian Chelba and Shobha Vasudevan and Eli Collins and Vasilisa Bashlovkina and Tony Lu and Doug Fritz and Jongbin Park and Yanqi Zhou and Chen Su and Richard Tanburn and Mikhail Sushkov and Mitchelle Rasquinha and Jinning Li and Jennifer Prendki and Yiming Li and Pallavi LV and Shriya Sharma and Hen Fitoussi and Hui Huang and Andrew Dai and Phuong Dao and Mike Burrows and Henry Prior and Danfeng Qin and Golan Pundak and Lars Lowe Sjoesund and Art Khurshudov and Zhenkai Zhu and Albert Webson and Elizabeth Kemp and Tat Tan and Saurabh Agrawal and Susie Sargsyan and Liqun Cheng and Jim Stephan and Tom Kwiatkowski and David Reid and Arunkumar Byravan and Assaf Hurwitz Michaely and Nicolas Heess and Luowei Zhou and Sonam Goenka and Viral Carpenter and Anselm Levskaya and Bo Wang and Reed Roberts and Rémi Leblond and Sharat Chikkerur and Stav Ginzburg and Max Chang and Robert Riachi and Chuqiao and Xu and Zalán Borsos and Michael Pliskin and Julia Pawar and Morgane Lustman and Hannah Kirkwood and Ankit Anand and Aditi Chaudhary and Norbert Kalb and Kieran Milan and Sean Augenstein and Anna Goldie and Laurel Prince and Karthik Raman and Yanhua Sun and Vivian Xia and Aaron Cohen and Zhouyuan Huo and Josh Camp and Seher Ellis and Lukas Zilka and David Vilar Torres and Lisa Patel and Sho Arora and Betty Chan and Jonas Adler and Kareem Ayoub and Jacky Liang and Fayaz Jamil and Jiepu Jiang and Simon Baumgartner and Haitian Sun and Yael Karov and Yaroslav Akulov and Hui Zheng and Irene Cai and Claudio Fantacci and James Rubin and Alex Rav Acha and Mengchao Wang and Nina D'Souza and Rohit Sathyanarayana and Shengyang Dai and Simon Rowe and Andrey Simanovsky and Omer Goldman and Yuheng Kuang and Xiaoyue Pan and Andrew Rosenberg and Tania Rojas-Esponda and Praneet Dutta and Amy Zeng and Irina Jurenka and Greg Farquhar and Yamini Bansal and Shariq Iqbal and Becca Roelofs and Ga-Young Joung and Parker Beak and Changwan Ryu and Ryan Poplin and Yan Wu and Jean-Baptiste Alayrac and Senaka Buthpitiya and Olaf Ronneberger and Caleb Habtegebriel and Wei Li and Paul Cavallaro and Aurora Wei and Guy Bensky and Timo Denk and Harish Ganapathy and Jeff Stanway and Pratik Joshi and Francesco Bertolini and Jessica Lo and Olivia Ma and Zachary Charles and Geta Sampemane and Himanshu Sahni and Xu Chen and Harry Askham and David Gaddy and Peter Young and Jiewen Tan and Matan Eyal and Arthur Bražinskas and Li Zhong and Zhichun Wu and Mark Epstein and Kai Bailey and Andrew Hard and Kamyu Lee and Sasha Goldshtein and Alex Ruiz and Mohammed Badawi and Matthias Lochbrunner and JK Kearns and Ashley Brown and Fabio Pardo and Theophane Weber and Haichuan Yang and Pan-Pan Jiang and Berkin Akin and Zhao Fu and Marcus Wainwright and Chi Zou and Meenu Gaba and Pierre-Antoine Manzagol and Wendy Kan and Yang Song and Karina Zainullina and Rui Lin and Jeongwoo Ko and Salil Deshmukh and Apoorv Jindal and James Svensson and Divya Tyam and Heri Zhao and Christine Kaeser-Chen and Scott Baird and Pooya Moradi and Jamie Hall and Qiuchen Guo and Vincent Tsang and Bowen Liang and Fernando Pereira and Suhas Ganesh and Ivan Korotkov and Jakub Adamek and Sridhar Thiagarajan and Vinh Tran and Charles Chen and Chris Tar and Sanil Jain and Ishita Dasgupta and Taylan Bilal and David Reitter and Kai Zhao and Giulia Vezzani and Yasmin Gehman and Pulkit Mehta and Lauren Beltrone and Xerxes Dotiwalla and Sergio Guadarrama and Zaheer Abbas and Stefani Karp and Petko Georgiev and Chun-Sung Ferng and Marc Brockschmidt and Liqian Peng and Christoph Hirnschall and Vikas Verma and Yingying Bi and Ying Xiao and Avigail Dabush and Kelvin Xu and Phil Wallis and Randall Parker and Qifei Wang and Yang Xu and Ilkin Safarli and Dinesh Tewari and Yin Zhang and Seungyeon Kim and Andrea Gesmundo and Mackenzie Thomas and Sergey Levi and Ahmed Chowdhury and Kanishka Rao and Peter Garst and Sam Conway-Rahman and Helen Ran and Kay McKinney and Zhisheng Xiao and Wenhao Yu and Rohan Agrawal and Axel Stjerngren and Catalin Ionescu and Jingjing Chen and Vivek Sharma and Justin Chiu and Fei Liu and Ken Franko and Clayton Sanford and Xingyu Cai and Paul Michel and Sanjay Ganapathy and Jane Labanowski and Zachary Garrett and Ben Vargas and Sean Sun and Bryan Gale and Thomas Buschmann and Guillaume Desjardins and Nimesh Ghelani and Palak Jain and Mudit Verma and Chulayuth Asawaroengchai and Julian Eisenschlos and Jitendra Harlalka and Hideto Kazawa and Don Metzler and Joshua Howland and Ying Jian and Jake Ades and Viral Shah and Tynan Gangwani and Seungji Lee and Roman Ring and Steven M. Hernandez and Dean Reich and Amer Sinha and Ashutosh Sathe and Joe Kovac and Ashleah Gill and Ajay Kannan and Andrea D'olimpio and Martin Sevenich and Jay Whang and Been Kim and Khe Chai Sim and Jilin Chen and Jiageng Zhang and Shuba Lall and Yossi Matias and Bill Jia and Abe Friesen and Sara Nasso and Ashish Thapliyal and Bryan Perozzi and Ting Yu and Anna Shekhawat and Safeen Huda and Peter Grabowski and Eric Wang and Ashwin Sreevatsa and Hilal Dib and Mehadi Hassen and Parker Schuh and Vedrana Milutinovic and Chris Welty and Michael Quinn and Ali Shah and Bangju Wang and Gabe Barth-Maron and Justin Frye and Natalie Axelsson and Tao Zhu and Yukun Ma and Irene Giannoumis and Hanie Sedghi and Chang Ye and Yi Luan and Kevin Aydin and Bilva Chandra and Vivek Sampathkumar and Ronny Huang and Victor Lavrenko and Ahmed Eleryan and Zhi Hong and Steven Hansen and Sara Mc Carthy and Bidisha Samanta and Domagoj Ćevid and Xin Wang and Fangtao Li and Michael Voznesensky and Matt Hoffman and Andreas Terzis and Vikash Sehwag and Gil Fidel and Luheng He and Mu Cai and Yanzhang He and Alex Feng and Martin Nikoltchev and Samrat Phatale and Jason Chase and Rory Lawton and Ming Zhang and Tom Ouyang and Manuel Tragut and Mehdi Hafezi Manshadi and Arjun Narayanan and Jiaming Shen and Xu Gao and Tolga Bolukbasi and Nick Roy and Xin Li and Daniel Golovin and Liviu Panait and Zhen Qin and Guangxing Han and Thomas Anthony and Sneha Kudugunta and Viorica Patraucean and Aniket Ray and Xinyun Chen and Xiaochen Yang and Tanuj Bhatia and Pranav Talluri and Alex Morris and Andrija Ražnatović and Bethanie Brownfield and James An and Sheng Peng and Patrick Kane and Ce Zheng and Nico Duduta and Joshua Kessinger and James Noraky and Siqi Liu and Keran Rong and Petar Veličković and Keith Rush and Alex Goldin and Fanny Wei and Shiva Mohan Reddy Garlapati and Caroline Pantofaru and Okwan Kwon and Jianmo Ni and Eric Noland and Julia Di Trapani and Françoise Beaufays and Abhijit Guha Roy and Yinlam Chow and Aybuke Turker and Geoffrey Cideron and Lantao Mei and Jon Clark and Qingyun Dou and Matko Bošnjak and Ralph Leith and Yuqing Du and Amir Yazdanbakhsh and Milad Nasr and Chester Kwak and Suraj Satishkumar Sheth and Alex Kaskasoli and Ankesh Anand and Balaji Lakshminarayanan and Sammy Jerome and David Bieber and Chun-Te Chu and Alexandre Senges and Tianxiao Shen and Mukund Sridhar and Ndaba Ndebele and Benjamin Beyret and Shakir Mohamed and Mia Chen and Markus Freitag and Jiaxian Guo and Luyang Liu and Paul Roit and Heng Chen and Shen Yan and Tom Stone and JD Co-Reyes and Jeremy Cole and Salvatore Scellato and Shekoofeh Azizi and Hadi Hashemi and Alicia Jin and Anand Iyer and Marcella Valentine and András György and Arun Ahuja and Daniel Hernandez Diaz and Chen-Yu Lee and Nathan Clement and Weize Kong and Drew Garmon and Ishaan Watts and Kush Bhatia and Khyatti Gupta and Matt Miecnikowski and Hugo Vallet and Ankur Taly and Edward Loper and Saket Joshi and James Atwood and Jo Chick and Mark Collier and Fotis Iliopoulos and Ryan Trostle and Beliz Gunel and Ramiro Leal-Cavazos and Arnar Mar Hrafnkelsson and Michael Guzman and Xiaoen Ju and Andy Forbes and Jesse Emond and Kushal Chauhan and Ben Caine and Li Xiao and Wenjun Zeng and Alexandre Moufarek and Daniel Murphy and Maya Meng and Nitish Gupta and Felix Riedel and Anil Das and Elijah Lawal and Shashi Narayan and Tiberiu Sosea and James Swirhun and Linda Friso and Behnam Neyshabur and Jing Lu and Sertan Girgin and Michael Wunder and Edouard Yvinec and Aroonalok Pyne and Victor Carbune and Shruti Rijhwani and Yang Guo and Tulsee Doshi and Anton Briukhov and Max Bain and Ayal Hitron and Xuanhui Wang and Ashish Gupta and Ke Chen and Cosmo Du and Weiyang Zhang and Dhruv Shah and Arjun Akula and Max Dylla and Ashyana Kachra and Weicheng Kuo and Tingting Zou and Lily Wang and Luyao Xu and Jifan Zhu and Justin Snyder and Sachit Menon and Orhan Firat and Igor Mordatch and Yuan Yuan and Natalia Ponomareva and Rory Blevins and Lawrence Moore and Weijun Wang and Phil Chen and Martin Scholz and Artur Dwornik and Jason Lin and Sicheng Li and Diego Antognini and Te I and Xiaodan Song and Matt Miller and Uday Kalra and Adam Raveret and Oscar Akerlund and Felix Wu and Andrew Nystrom and Namrata Godbole and Tianqi Liu and Hannah DeBalsi and Jewel Zhao and Buhuang Liu and Avi Caciularu and Lauren Lax and Urvashi Khandelwal and Victoria Langston and Eric Bailey and Silvio Lattanzi and Yufei Wang and Neel Kovelamudi and Sneha Mondal and Guru Guruganesh and Nan Hua and Ofir Roval and Paweł Wesołowski and Rishikesh Ingale and Jonathan Halcrow and Tim Sohn and Christof Angermueller and Bahram Raad and Eli Stickgold and Eva Lu and Alec Kosik and Jing Xie and Timothy Lillicrap and Austin Huang and Lydia Lihui Zhang and Dominik Paulus and Clement Farabet and Alex Wertheim and Bing Wang and Rishabh Joshi and Chu-ling Ko and Yonghui Wu and Shubham Agrawal and Lily Lin and XiangHai Sheng and Peter Sung and Tyler Breland-King and Christina Butterfield and Swapnil Gawde and Sumeet Singh and Qiao Zhang and Raj Apte and Shilpa Shetty and Adrian Hutter and Tao Li and Elizabeth Salesky and Federico Lebron and Jonni Kanerva and Michela Paganini and Arthur Nguyen and Rohith Vallu and Jan-Thorsten Peter and Sarmishta Velury and David Kao and Jay Hoover and Anna Bortsova and Colton Bishop and Shoshana Jakobovits and Alessandro Agostini and Alekh Agarwal and Chang Liu and Charles Kwong and Sasan Tavakkol and Ioana Bica and Alex Greve and Anirudh GP and Jake Marcus and Le Hou and Tom Duerig and Rivka Moroshko and Dave Lacey and Andy Davis and Julien Amelot and Guohui Wang and Frank Kim and Theofilos Strinopoulos and Hui Wan and Charline Le Lan and Shankar Krishnan and Haotian Tang and Peter Humphreys and Junwen Bai and Idan Heimlich Shtacher and Diego Machado and Chenxi Pang and Ken Burke and Dangyi Liu and Renga Aravamudhan and Yue Song and Ed Hirst and Abhimanyu Singh and Brendan Jou and Liang Bai and Francesco Piccinno and Chuyuan Kelly Fu and Robin Alazard and Barak Meiri and Daniel Winter and Charlie Chen and Mingda Zhang and Jens Heitkaemper and John Lambert and Jinhyuk Lee and Alexander Frömmgen and Sergey Rogulenko and Pranav Nair and Paul Niemczyk and Anton Bulyenov and Bibo Xu and Hadar Shemtov and Morteza Zadimoghaddam and Serge Toropov and Mateo Wirth and Hanjun Dai and Sreenivas Gollapudi and Daniel Zheng and Alex Kurakin and Chansoo Lee and Kalesha Bullard and Nicolas Serrano and Ivana Balazevic and Yang Li and Johan Schalkwyk and Mark Murphy and Mingyang Zhang and Kevin Sequeira and Romina Datta and Nishant Agrawal and Charles Sutton and Nithya Attaluri and Mencher Chiang and Wael Farhan and Gregory Thornton and Kate Lin and Travis Choma and Hung Nguyen and Kingshuk Dasgupta and Dirk Robinson and Iulia Comşa and Michael Riley and Arjun Pillai and Basil Mustafa and Ben Golan and Amir Zandieh and Jean-Baptiste Lespiau and Billy Porter and David Ross and Sujeevan Rajayogam and Mohit Agarwal and Subhashini Venugopalan and Bobak Shahriari and Qiqi Yan and Hao Xu and Taylor Tobin and Pavel Dubov and Hongzhi Shi and Adrià Recasens and Anton Kovsharov and Sebastian Borgeaud and Lucio Dery and Shanthal Vasanth and Elena Gribovskaya and Linhai Qiu and Mahdis Mahdieh and Wojtek Skut and Elizabeth Nielsen and CJ Zheng and Adams Yu and Carrie Grimes Bostock and Shaleen Gupta and Aaron Archer and Chris Rawles and Elinor Davies and Alexey Svyatkovskiy and Tomy Tsai and Yoni Halpern and Christian Reisswig and Bartek Wydrowski and Bo Chang and Joan Puigcerver and Mor Hazan Taege and Jian Li and Eva Schnider and Xinjian Li and Dragos Dena and Yunhan Xu and Umesh Telang and Tianze Shi and Heiga Zen and Kyle Kastner and Yeongil Ko and Neesha Subramaniam and Aviral Kumar and Pete Blois and Zhuyun Dai and John Wieting and Yifeng Lu and Yoel Zeldes and Tian Xie and Anja Hauth and Alexandru Ţifrea and Yuqi Li and Sam El-Husseini and Dan Abolafia and Howard Zhou and Wen Ding and Sahra Ghalebikesabi and Carlos Guía and Andrii Maksai and Ágoston Weisz and Sercan Arik and Nick Sukhanov and Aga Świetlik and Xuhui Jia and Luo Yu and Weiyue Wang and Mark Brand and Dawn Bloxwich and Sean Kirmani and Zhe Chen and Alec Go and Pablo Sprechmann and Nithish Kannen and Alen Carin and Paramjit Sandhu and Isabel Edkins and Leslie Nooteboom and Jai Gupta and Loren Maggiore and Javad Azizi and Yael Pritch and Pengcheng Yin and Mansi Gupta and Danny Tarlow and Duncan Smith and Desi Ivanov and Mohammad Babaeizadeh and Ankita Goel and Satish Kambala and Grace Chu and Matej Kastelic and Michelle Liu and Hagen Soltau and Austin Stone and Shivani Agrawal and Min Kim and Kedar Soparkar and Srinivas Tadepalli and Oskar Bunyan and Rachel Soh and Arvind Kannan and DY Kim and Blake JianHang Chen and Afief Halumi and Sudeshna Roy and Yulong Wang and Olcan Sercinoglu and Gena Gibson and Sijal Bhatnagar and Motoki Sano and Daniel von Dincklage and Qingchun Ren and Blagoj Mitrevski and Mirek Olšák and Jennifer She and Carl Doersch and Jilei and Wang and Bingyuan Liu and Qijun Tan and Tamar Yakar and Tris Warkentin and Alex Ramirez and Carl Lebsack and Josh Dillon and Rajiv Mathews and Tom Cobley and Zelin Wu and Zhuoyuan Chen and Jon Simon and Swaroop Nath and Tara Sainath and Alexei Bendebury and Ryan Julian and Bharath Mankalale and Daria Ćurko and Paulo Zacchello and Adam R. Brown and Kiranbir Sodhia and Heidi Howard and Sergi Caelles and Abhinav Gupta and Gareth Evans and Anna Bulanova and Lesley Katzen and Roman Goldenberg and Anton Tsitsulin and Joe Stanton and Benoit Schillings and Vitaly Kovalev and Corey Fry and Rushin Shah and Kuo Lin and Shyam Upadhyay and Cheng Li and Soroush Radpour and Marcello Maggioni and Jing Xiong and Lukas Haas and Jenny Brennan and Aishwarya Kamath and Nikolay Savinov and Arsha Nagrani and Trevor Yacovone and Ryan Kappedal and Kostas Andriopoulos and Li Lao and YaGuang Li and Grigory Rozhdestvenskiy and Kazuma Hashimoto and Andrew Audibert and Sophia Austin and Daniel Rodriguez and Anian Ruoss and Garrett Honke and Deep Karkhanis and Xi Xiong and Qing Wei and James Huang and Zhaoqi Leng and Vittal Premachandran and Stan Bileschi and Georgios Evangelopoulos and Thomas Mensink and Jay Pavagadhi and Denis Teplyashin and Paul Chang and Linting Xue and Garrett Tanzer and Sally Goldman and Kaushal Patel and Shixin Li and Jeremy Wiesner and Ivy Zheng and Ian Stewart-Binks and Jie Han and Zhi Li and Liangchen Luo and Karel Lenc and Mario Lučić and Fuzhao Xue and Ryan Mullins and Alexey Guseynov and Chung-Ching Chang and Isaac Galatzer-Levy and Adam Zhang and Garrett Bingham and Grace Hu and Ale Hartman and Yue Ma and Jordan Griffith and Alex Irpan and Carey Radebaugh and Summer Yue and Lijie Fan and Victor Ungureanu and Christina Sorokin and Hannah Teufel and Peiran Li and Rohan Anil and Dimitris Paparas and Todd Wang and Chu-Cheng Lin and Hui Peng and Megan Shum and Goran Petrovic and Demetra Brady and Richard Nguyen and Klaus Macherey and Zhihao Li and Harman Singh and Madhavi Yenugula and Mariko Iinuma and Xinyi Chen and Kavya Kopparapu and Alexey Stern and Shachi Dave and Chandu Thekkath and Florence Perot and Anurag Kumar and Fangda Li and Yang Xiao and Matthew Bilotti and Mohammad Hossein Bateni and Isaac Noble and Lisa Lee and Amelio Vázquez-Reina and Julian Salazar and Xiaomeng Yang and Boyu Wang and Ela Gruzewska and Anand Rao and Sindhu Raghuram and Zheng Xu and Eyal Ben-David and Jieru Mei and Sid Dalmia and Zhaoyi Zhang and Yuchen Liu and Gagan Bansal and Helena Pankov and Steven Schwarcz and Andrea Burns and Christine Chan and Sumit Sanghai and Ricky Liang and Ethan Liang and Antoine He and Amy Stuart and Arun Narayanan and Yukun Zhu and Christian Frank and Bahar Fatemi and Amit Sabne and Oran Lang and Indro Bhattacharya and Shane Settle and Maria Wang and Brendan McMahan and Andrea Tacchetti and Livio Baldini Soares and Majid Hadian and Serkan Cabi and Timothy Chung and Nikita Putikhin and Gang Li and Jeremy Chen and Austin Tarango and Henryk Michalewski and Mehran Kazemi and Hussain Masoom and Hila Sheftel and Rakesh Shivanna and Archita Vadali and Ramona Comanescu and Doug Reid and Joss Moore and Arvind Neelakantan and Michaël Sander and Jonathan Herzig and Aviv Rosenberg and Mostafa Dehghani and JD Choi and Michael Fink and Reid Hayes and Eric Ge and Shitao Weng and Chia-Hua Ho and John Karro and Kalpesh Krishna and Lam Nguyen Thiet and Amy Skerry-Ryan and Daniel Eppens and Marco Andreetto and Navin Sarma and Silvano Bonacina and Burcu Karagol Ayan and Megha Nawhal and Zhihao Shan and Mike Dusenberry and Shantanu Thakoor and Sagar Gubbi and Duc Dung Nguyen and Reut Tsarfaty and Samuel Albanie and Jovana Mitrović and Meet Gandhi and Bo-Juen Chen and Alessandro Epasto and Georgi Stephanov and Ye Jin and Samuel Gehman and Aida Amini and Jack Weber and Feryal Behbahani and Shawn Xu and Miltos Allamanis and Xi Chen and Myle Ott and Claire Sha and Michal Jastrzebski and Hang Qi and David Greene and Xinyi Wu and Abodunrinwa Toki and Daniel Vlasic and Jane Shapiro and Ragha Kotikalapudi and Zhe Shen and Takaaki Saeki and Sirui Xie and Albin Cassirer and Shikhar Bharadwaj and Tatsuya Kiyono and Srinadh Bhojanapalli and Elan Rosenfeld and Sam Ritter and Jieming Mao and João Gabriel Oliveira and Zoltan Egyed and Bernd Bandemer and Emilio Parisotto and Keisuke Kinoshita and Juliette Pluto and Petros Maniatis and Steve Li and Yaohui Guo and Golnaz Ghiasi and Jean Tarbouriech and Srimon Chatterjee and Julie Jin and Katrina and Xu and Jennimaria Palomaki and Séb Arnold and Madhavi Sewak and Federico Piccinini and Mohit Sharma and Ben Albrecht and Sean Purser-haskell and Ashwin Vaswani and Chongyan Chen and Matheus Wisniewski and Qin Cao and John Aslanides and Nguyet Minh Phu and Maximilian Sieb and Lauren Agubuzu and Anne Zheng and Daniel Sohn and Marco Selvi and Anders Andreassen and Krishan Subudhi and Prem Eruvbetine and Oliver Woodman and Tomas Mery and Sebastian Krause and Xiaoqi Ren and Xiao Ma and Jincheng Luo and Dawn Chen and Wei Fan and Henry Griffiths and Christian Schuler and Alice Li and Shujian Zhang and Jean-Michel Sarr and Shixin Luo and Riccardo Patana and Matthew Watson and Dani Naboulsi and Michael Collins and Sailesh Sidhwani and Emiel Hoogeboom and Sharon Silver and Emily Caveness and Xiaokai Zhao and Mikel Rodriguez and Maxine Deines and Libin Bai and Patrick Griffin and Marco Tagliasacchi and Emily Xue and Spandana Raj Babbula and Bo Pang and Nan Ding and Gloria Shen and Elijah Peake and Remi Crocker and Shubha Srinivas Raghvendra and Danny Swisher and Woohyun Han and Richa Singh and Ling Wu and Vladimir Pchelin and Tsendsuren Munkhdalai and Dana Alon and Geoff Bacon and Efren Robles and Jannis Bulian and Melvin Johnson and George Powell and Felipe Tiengo Ferreira and Yaoyiran Li and Frederik Benzing and Mihajlo Velimirović and Hubert Soyer and William Kong and Tony and Nguyên and Zhen Yang and Jeremiah Liu and Joost van Amersfoort and Daniel Gillick and Baochen Sun and Nathalie Rauschmayr and Katie Zhang and Serena Zhan and Tao Zhou and Alexey Frolov and Chengrun Yang and Denis Vnukov and Louis Rouillard and Hongji Li and Amol Mandhane and Nova Fallen and Rajesh Venkataraman and Clara Huiyi Hu and Jennifer Brennan and Jenny Lee and Jerry Chang and Martin Sundermeyer and Zhufeng Pan and Rosemary Ke and Simon Tong and Alex Fabrikant and William Bono and Jindong Gu and Ryan Foley and Yiran Mao and Manolis Delakis and Dhruva Bhaswar and Roy Frostig and Nick Li and Avital Zipori and Cath Hope and Olga Kozlova and Swaroop Mishra and Josip Djolonga and Craig Schiff and Majd Al Merey and Eleftheria Briakou and Peter Morgan and Andy Wan and Avinatan Hassidim and RJ Skerry-Ryan and Kuntal Sengupta and Mary Jasarevic and Praveen Kallakuri and Paige Kunkle and Hannah Brennan and Tom Lieber and Hassan Mansoor and Julian Walker and Bing Zhang and Annie Xie and Goran Žužić and Adaeze Chukwuka and Alex Druinsky and Donghyun Cho and Rui Yao and Ferjad Naeem and Shiraz Butt and Eunyoung Kim and Zhipeng Jia and Mandy Jordan and Adam Lelkes and Mark Kurzeja and Sophie Wang and James Zhao and Andrew Over and Abhishek Chakladar and Marcel Prasetya and Neha Jha and Sriram Ganapathy and Yale Cong and Prakash Shroff and Carl Saroufim and Sobhan Miryoosefi and Mohamed Hammad and Tajwar Nasir and Weijuan Xi and Yang Gao and Young Maeng and Ben Hora and Chin-Yi Cheng and Parisa Haghani and Yoad Lewenberg and Caden Lu and Martin Matysiak and Naina Raisinghani and Huiyu Wang and Lexi Baugher and Rahul Sukthankar and Minh Giang and John Schultz and Noah Fiedel and Minmin Chen and Cheng-Chun Lee and Tapomay Dey and Hao Zheng and Shachi Paul and Celine Smith and Andy Ly and Yicheng Wang and Rishabh Bansal and Bartek Perz and Susanna Ricco and Stasha Blank and Vaishakh Keshava and Deepak Sharma and Marvin Chow and Kunal Lad and Komal Jalan and Simon Osindero and Craig Swanson and Jacob Scott and Anastasija Ilić and Xiaowei Li and Siddhartha Reddy Jonnalagadda and Afzal Shama Soudagar and Yan Xiong and Bat-Orgil Batsaikhan and Daniel Jarrett and Naveen Kumar and Maulik Shah and Matt Lawlor and Austin Waters and Mark Graham and Rhys May and Sabela Ramos and Sandra Lefdal and Zeynep Cankara and Nacho Cano and Brendan O'Donoghue and Jed Borovik and Frederick Liu and Jordan Grimstad and Mahmoud Alnahlawi and Katerina Tsihlas and Tom Hudson and Nikolai Grigorev and Yiling Jia and Terry Huang and Tobenna Peter Igwe and Sergei Lebedev and Xiaodan Tang and Igor Krivokon and Frankie Garcia and Melissa Tan and Eric Jia and Peter Stys and Shikhar Vashishth and Yu Liang and Balaji Venkatraman and Chenjie Gu and Anastasios Kementsietsidis and Chen Zhu and Junehyuk Jung and Yunfei Bai and Mohammad Javad Hosseini and Faruk Ahmed and Aditya Gupta and Xin Yuan and Shereen Ashraf and Shitij Nigam and Gautam Vasudevan and Pranjal Awasthi and Adi Mayrav Gilady and Zelda Mariet and Ramy Eskander and Haiguang Li and Hexiang Hu and Guillermo Garrido and Philippe Schlattner and George Zhang and Rohun Saxena and Petar Dević and Kritika Muralidharan and Ashwin Murthy and Yiqian Zhou and Min Choi and Arissa Wongpanich and Zhengdong Wang and Premal Shah and Yuntao Xu and Yiling Huang and Stephen Spencer and Alice Chen and James Cohan and Junjie Wang and Jonathan Tompson and Junru Wu and Ruba Haroun and Haiqiong Li and Blanca Huergo and Fan Yang and Tongxin Yin and James Wendt and Michael Bendersky and Rahma Chaabouni and Javier Snaider and Johan Ferret and Abhishek Jindal and Tara Thompson and Andrew Xue and Will Bishop and Shubham Milind Phal and Archit Sharma and Yunhsuan Sung and Prabakar Radhakrishnan and Mo Shomrat and Reeve Ingle and Roopali Vij and Justin Gilmer and Mihai Dorin Istin and Sam Sobell and Yang Lu and Emily Nottage and Dorsa Sadigh and Jeremiah Willcock and Tingnan Zhang and Steve Xu and Sasha Brown and Katherine Lee and Gary Wang and Yun Zhu and Yi Tay and Cheolmin Kim and Audrey Gutierrez and Abhanshu Sharma and Yongqin Xian and Sungyong Seo and Claire Cui and Elena Pochernina and Cip Baetu and Krzysztof Jastrzębski and Mimi Ly and Mohamed Elhawaty and Dan Suh and Eren Sezener and Pidong Wang and Nancy Yuen and George Tucker and Jiahao Cai and Zuguang Yang and Cindy Wang and Alex Muzio and Hai Qian and Jae Yoo and Derek Lockhart and Kevin R. McKee and Mandy Guo and Malika Mehrotra and Artur Mendonça and Sanket Vaibhav Mehta and Sherry Ben and Chetan Tekur and Jiaqi Mu and Muye Zhu and Victoria Krakovna and Hongrae Lee and AJ Maschinot and Sébastien Cevey and HyunJeong Choe and Aijun Bai and Hansa Srinivasan and Derek Gasaway and Nick Young and Patrick Siegler and Dan Holtmann-Rice and Vihari Piratla and Kate Baumli and Roey Yogev and Alex Hofer and Hado van Hasselt and Svetlana Grant and Yuri Chervonyi and David Silver and Andrew Hogue and Ayushi Agarwal and Kathie Wang and Preeti Singh and Four Flynn and Josh Lipschultz and Robert David and Lizzetth Bellot and Yao-Yuan Yang and Long Le and Filippo Graziano and Kate Olszewska and Kevin Hui and Akanksha Maurya and Nikos Parotsidis and Weijie Chen and Tayo Oguntebi and Joe Kelley and Anirudh Baddepudi and Johannes Mauerer and Gregory Shaw and Alex Siegman and Lin Yang and Shravya Shetty and Subhrajit Roy and Yunting Song and Wojciech Stokowiec and Ryan Burnell and Omkar Savant and Robert Busa-Fekete and Jin Miao and Samrat Ghosh and Liam MacDermed and Phillip Lippe and Mikhail Dektiarev and Zach Behrman and Fabian Mentzer and Kelvin Nguyen and Meng Wei and Siddharth Verma and Chris Knutsen and Sudeep Dasari and Zhipeng Yan and Petr Mitrichev and Xingyu Wang and Virat Shejwalkar and Jacob Austin and Srinivas Sunkara and Navneet Potti and Yan Virin and Christian Wright and Gaël Liu and Oriana Riva and Etienne Pot and Greg Kochanski and Quoc Le and Gargi Balasubramaniam and Arka Dhar and Yuguo Liao and Adam Bloniarz and Divyansh Shukla and Elizabeth Cole and Jong Lee and Sheng Zhang and Sushant Kafle and Siddharth Vashishtha and Parsa Mahmoudieh and Grace Chen and Raphael Hoffmann and Pranesh Srinivasan and Agustin Dal Lago and Yoav Ben Shalom and Zi Wang and Michael Elabd and Anuj Sharma and Junhyuk Oh and Suraj Kothawade and Maigo Le and Marianne Monteiro and Shentao Yang and Kaiz Alarakyia and Robert Geirhos and Diana Mincu and Håvard Garnes and Hayato Kobayashi and Soroosh Mariooryad and Kacper Krasowiak and Zhixin and Lai and Shibl Mourad and Mingqiu Wang and Fan Bu and Ophir Aharoni and Guanjie Chen and Abhimanyu Goyal and Vadim Zubov and Ankur Bapna and Elahe Dabir and Nisarg Kothari and Kay Lamerigts and Nicola De Cao and Jeremy Shar and Christopher Yew and Nitish Kulkarni and Dre Mahaarachchi and Mandar Joshi and Zhenhai Zhu and Jared Lichtarge and Yichao Zhou and Hannah Muckenhirn and Vittorio Selo and Oriol Vinyals and Peter Chen and Anthony Brohan and Vaibhav Mehta and Sarah Cogan and Ruth Wang and Ty Geri and Wei-Jen Ko and Wei Chen and Fabio Viola and Keshav Shivam and Lisa Wang and Madeleine Clare Elish and Raluca Ada Popa and Sébastien Pereira and Jianqiao Liu and Raphael Koster and Donnie Kim and Gufeng Zhang and Sayna Ebrahimi and Partha Talukdar and Yanyan Zheng and Petra Poklukar and Ales Mikhalap and Dale Johnson and Anitha Vijayakumar and Mark Omernick and Matt Dibb and Ayush Dubey and Qiong Hu and Apurv Suman and Vaibhav Aggarwal and Ilya Kornakov and Fei Xia and Wing Lowe and Alexey Kolganov and Ted Xiao and Vitaly Nikolaev and Steven Hemingray and Bonnie Li and Joana Iljazi and Mikołaj Rybiński and Ballie Sandhu and Peggy Lu and Thang Luong and Rodolphe Jenatton and Vineetha Govindaraj and Hui and Li and Gabriel Dulac-Arnold and Wonpyo Park and Henry Wang and Abhinit Modi and Jean Pouget-Abadie and Kristina Greller and Rahul Gupta and Robert Berry and Prajit Ramachandran and Jinyu Xie and Liam McCafferty and Jianling Wang and Kilol Gupta and Hyeontaek Lim and Blaž Bratanič and Andy Brock and Ilia Akolzin and Jim Sproch and Dan Karliner and Duhyeon Kim and Adrian Goedeckemeyer and Noam Shazeer and Cordelia Schmid and Daniele Calandriello and Parul Bhatia and Krzysztof Choromanski and Ceslee Montgomery and Dheeru Dua and Ana Ramalho and Helen King and Yue Gao and Lynn Nguyen and David Lindner and Divya Pitta and Oleaser Johnson and Khalid Salama and Diego Ardila and Michael Han and Erin Farnese and Seth Odoom and Ziyue Wang and Xiangzhuo Ding and Norman Rink and Ray Smith and Harshal Tushar Lehri and Eden Cohen and Neera Vats and Tong He and Parthasarathy Gopavarapu and Adam Paszke and Miteyan Patel and Wouter Van Gansbeke and Lucia Loher and Luis Castro and Maria Voitovich and Tamara von Glehn and Nelson George and Simon Niklaus and Zach Eaton-Rosen and Nemanja Rakićević and Erik Jue and Sagi Perel and Carrie Zhang and Yuval Bahat and Angéline Pouget and Zhi Xing and Fantine Huot and Ashish Shenoy and Taylor Bos and Vincent Coriou and Bryan Richter and Natasha Noy and Yaqing Wang and Santiago Ontanon and Siyang Qin and Gleb Makarchuk and Demis Hassabis and Zhuowan Li and Mandar Sharma and Kumaran Venkatesan and Iurii Kemaev and Roxanne Daniel and Shiyu Huang and Saloni Shah and Octavio Ponce and Warren and Chen and Manaal Faruqui and Jialin Wu and Slavica Andačić and Szabolcs Payrits and Daniel McDuff and Tom Hume and Yuan Cao and MH Tessler and Qingze Wang and Yinan Wang and Ivor Rendulic and Eirikur Agustsson and Matthew Johnson and Tanya Lando and Andrew Howard and Sri Gayatri Sundara Padmanabhan and Mayank Daswani and Andrea Banino and Michael Kilgore and Jonathan Heek and Ziwei Ji and Alvaro Caceres and Conglong Li and Nora Kassner and Alexey Vlaskin and Zeyu Liu and Alex Grills and Yanhan Hou and Roykrong Sukkerd and Gowoon Cheon and Nishita Shetty and Larisa Markeeva and Piotr Stanczyk and Tejas Iyer and Yuan Gong and Shawn Gao and Keerthana Gopalakrishnan and Tim Blyth and Malcolm Reynolds and Avishkar Bhoopchand and Misha Bilenko and Dero Gharibian and Vicky Zayats and Aleksandra Faust and Abhinav Singh and Min Ma and Hongyang Jiao and Sudheendra Vijayanarasimhan and Lora Aroyo and Vikas Yadav and Sarah Chakera and Ashwin Kakarla and Vilobh Meshram and Karol Gregor and Gabriela Botea and Evan Senter and Dawei Jia and Geza Kovacs and Neha Sharma and Sebastien Baur and Kai Kang and Yifan He and Lin Zhuo and Marija Kostelac and Itay Laish and Songyou Peng and Louis O'Bryan and Daniel Kasenberg and Girish Ramchandra Rao and Edouard Leurent and Biao Zhang and Sage Stevens and Ana Salazar and Ye Zhang and Ivan Lobov and Jake Walker and Allen Porter and Morgan Redshaw and Han Ke and Abhishek Rao and Alex Lee and Hoi Lam and Michael Moffitt and Jaeyoun Kim and Siyuan Qiao and Terry Koo and Robert Dadashi and Xinying Song and Mukund Sundararajan and Peng Xu and Chizu Kawamoto and Yan Zhong and Clara Barbu and Apoorv Reddy and Mauro Verzetti and Leon Li and George Papamakarios and Hanna Klimczak-Plucińska and Mary Cassin and Koray Kavukcuoglu and Rigel Swavely and Alain Vaucher and Jeffrey Zhao and Ross Hemsley and Michael Tschannen and Heming Ge and Gaurav Menghani and Yang Yu and Natalie Ha and Wei He and Xiao Wu and Maggie Song and Rachel Sterneck and Stefan Zinke and Dan A. Calian and Annie Marsden and Alejandro Cruzado Ruiz and Matteo Hessel and Almog Gueta and Benjamin Lee and Brian Farris and Manish Gupta and Yunjie Li and Mohammad Saleh and Vedant Misra and Kefan Xiao and Piermaria Mendolicchio and Gavin Buttimore and Varvara Krayvanova and Nigamaa Nayakanti and Matthew Wiethoff and Yash Pande and Azalia Mirhoseini and Ni Lao and Jasmine Liu and Yiqing Hua and Angie Chen and Yury Malkov and Dmitry Kalashnikov and Shubham Gupta and Kartik Audhkhasi and Yuexiang Zhai and Sudhindra Kopalle and Prateek Jain and Eran Ofek and Clemens Meyer and Khuslen Baatarsukh and Hana Strejček and Jun Qian and James Freedman and Ricardo Figueira and Michal Sokolik and Olivier Bachem and Raymond Lin and Dia Kharrat and Chris Hidey and Pingmei Xu and Dennis Duan and Yin Li and Muge Ersoy and Richard Everett and Kevin Cen and Rebeca Santamaria-Fernandez and Amir Taubenfeld and Ian Mackinnon and Linda Deng and Polina Zablotskaia and Shashank Viswanadha and Shivanker Goel and Damion Yates and Yunxiao Deng and Peter Choy and Mingqing Chen and Abhishek Sinha and Alex Mossin and Yiming Wang and Arthur Szlam and Susan Hao and Paul Kishan Rubenstein and Metin Toksoz-Exley and Miranda Aperghis and Yin Zhong and Junwhan Ahn and Michael Isard and Olivier Lacombe and Florian Luisier and Chrysovalantis Anastasiou and Yogesh Kalley and Utsav Prabhu and Emma Dunleavy and Shaan Bijwadia and Justin Mao-Jones and Kelly Chen and Rama Pasumarthi and Emily Wood and Adil Dostmohamed and Nate Hurley and Jiri Simsa and Alicia Parrish and Mantas Pajarskas and Matt Harvey and Ondrej Skopek and Yony Kochinski and Javier Rey and Verena Rieser and Denny Zhou and Sun Jae Lee and Trilok Acharya and Guowang Li and Joe Jiang and Xiaofan Zhang and Bryant Gipson and Ethan Mahintorabi and Marco Gelmi and Nima Khajehnouri and Angel Yeh and Kayi Lee and Loic Matthey and Leslie Baker and Trang Pham and Han Fu and Alex Pak and Prakhar Gupta and Cristina Vasconcelos and Adam Sadovsky and Brian Walker and Sissie Hsiao and Patrik Zochbauer and Andreea Marzoca and Noam Velan and Junhao Zeng and Gilles Baechler and Danny Driess and Divya Jain and Yanping Huang and Lizzie Tao and John Maggs and Nir Levine and Jon Schneider and Erika Gemzer and Samuel Petit and Shan Han and Zach Fisher and Dustin Zelle and Courtney Biles and Eugene Ie and Asya Fadeeva and Casper Liu and Juliana Vicente Franco and Adrian Collister and Hao Zhang and Renshen Wang and Ruizhe Zhao and Leandro Kieliger and Kurt Shuster and Rui Zhu and Boqing Gong and Lawrence Chan and Ruoxi Sun and Sujoy Basu and Roland Zimmermann and Jamie Hayes and Abhishek Bapna and Jasper Snoek and Weel Yang and Puranjay Datta and Jad Al Abdallah and Kevin Kilgour and Lu Li and SQ Mah and Yennie Jun and Morgane Rivière and Abhijit Karmarkar and Tammo Spalink and Tao Huang and Lucas Gonzalez and Duc-Hieu Tran and Averi Nowak and John Palowitch and Martin Chadwick and Ellie Talius and Harsh Mehta and Thibault Sellam and Philipp Fränken and Massimo Nicosia and Kyle He and Aditya Kini and David Amos and Sugato Basu and Harrison Jobe and Eleni Shaw and Qiantong Xu and Colin Evans and Daisuke Ikeda and Chaochao Yan and Larry Jin and Lun Wang and Sachin Yadav and Ilia Labzovsky and Ramesh Sampath and Ada Ma and Candice Schumann and Aditya Siddhant and Rohin Shah and John Youssef and Rishabh Agarwal and Natalie Dabney and Alessio Tonioni and Moran Ambar and Jing Li and Isabelle Guyon and Benny Li and David Soergel and Boya Fang and Georgi Karadzhov and Cristian Udrescu and Trieu Trinh and Vikas Raunak and Seb Noury and Dee Guo and Sonal Gupta and Mara Finkelstein and Denis Petek and Lihao Liang and Greg Billock and Pei Sun and David Wood and Yiwen Song and Xiaobin Yu and Tatiana Matejovicova and Regev Cohen and Kalyan Andra and David D'Ambrosio and Zhiwei Deng and Vincent Nallatamby and Ebrahim Songhori and Rumen Dangovski and Andrew Lampinen and Pankil Botadra and Adam Hillier and Jiawei Cao and Nagabhushan Baddi and Adhi Kuncoro and Toshihiro Yoshino and Ankit Bhagatwala and Marcáurelio Ranzato and Rylan Schaeffer and Tianlin Liu and Shuai Ye and Obaid Sarvana and John Nham and Chenkai Kuang and Isabel Gao and Jinoo Baek and Shubham Mittal and Ayzaan Wahid and Anita Gergely and Bin Ni and Josh Feldman and Carrie Muir and Pascal Lamblin and Wolfgang Macherey and Ethan Dyer and Logan Kilpatrick and Víctor Campos and Mukul Bhutani and Stanislav Fort and Yanif Ahmad and Aliaksei Severyn and Kleopatra Chatziprimou and Oleksandr Ferludin and Mason Dimarco and Aditya Kusupati and Joe Heyward and Dan Bahir and Kevin Villela and Katie Millican and Dror Marcus and Sanaz Bahargam and Caglar Unlu and Nicholas Roth and Zichuan Wei and Siddharth Gopal and Deepanway Ghoshal and Edward Lee and Sharon Lin and Jennie Lees and Dayeong Lee and Anahita Hosseini and Connie Fan and Seth Neel and Marcus Wu and Yasemin Altun and Honglong Cai and Enrique Piqueras and Josh Woodward and Alessandro Bissacco and Salem Haykal and Mahyar Bordbar and Prasha Sundaram and Sarah Hodkinson and Daniel Toyama and George Polovets and Austin Myers and Anu Sinha and Tomer Levinboim and Kashyap Krishnakumar and Rachita Chhaparia and Tatiana Sholokhova and Nitesh Bharadwaj Gundavarapu and Ganesh Jawahar and Haroon Qureshi and Jieru Hu and Nikola Momchev and Matthew Rahtz and Renjie Wu and Aishwarya P S and Kedar Dhamdhere and Meiqi Guo and Umang Gupta and Ali Eslami and Mariano Schain and Michiel Blokzijl and David Welling and Dave Orr and Levent Bolelli and Nicolas Perez-Nieves and Mikhail Sirotenko and Aman Prasad and Arjun Kar and Borja De Balle Pigem and Tayfun Terzi and Gellért Weisz and Dipankar Ghosh and Aditi Mavalankar and Dhruv Madeka and Kaspar Daugaard and Hartwig Adam and Viraj Shah and Dana Berman and Maggie Tran and Steven Baker and Ewa Andrejczuk and Grishma Chole and Ganna Raboshchuk and Mahdi Mirzazadeh and Thais Kagohara and Shimu Wu and Christian Schallhart and Bernett Orlando and Chen Wang and Alban Rrustemi and Hao Xiong and Hao Liu and Arpi Vezer and Nolan Ramsden and Shuo-yiin Chang and Sidharth Mudgal and Yan Li and Nino Vieillard and Yedid Hoshen and Farooq Ahmad and Ambrose Slone and Amy Hua and Natan Potikha and Mirko Rossini and Jon Stritar and Sushant Prakash and Zifeng Wang and Xuanyi Dong and Alireza Nazari and Efrat Nehoran and Kaan Tekelioglu and Yinxiao Li and Kartikeya Badola and Tom Funkhouser and Yuanzhen Li and Varun Yerram and Ramya Ganeshan and Daniel Formoso and Karol Langner and Tian Shi and Huijian Li and Yumeya Yamamori and Amayika Panda and Alaa Saade and Angelo Scorza Scarpati and Chris Breaux and CJ Carey and Zongwei Zhou and Cho-Jui Hsieh and Sophie Bridgers and Alena Butryna and Nishesh Gupta and Vaibhav Tulsyan and Sanghyun Woo and Evgenii Eltyshev and Will Grathwohl and Chanel Parks and Seth Benjamin and Rina Panigrahy and Shenil Dodhia and Daniel De Freitas and Chris Sauer and Will Song and Ferran Alet and Jackson Tolins and Cosmin Paduraru and Xingyi Zhou and Brian Albert and Zizhao Zhang and Lei Shu and Mudit Bansal and Sarah Nguyen and Amir Globerson and Owen Xiao and James Manyika and Tom Hennigan and Rong Rong and Josip Matak and Anton Bakalov and Ankur Sharma and Danila Sinopalnikov and Andrew Pierson and Stephen Roller and Geoff Brown and Mingcen Gao and Toshiyuki Fukuzawa and Amin Ghafouri and Kenny Vassigh and Iain Barr and Zhicheng Wang and Anna Korsun and Rajesh Jayaram and Lijie Ren and Tim Zaman and Samira Khan and Yana Lunts and Dan Deutsch and Dave Uthus and Nitzan Katz and Masha Samsikova and Amr Khalifa and Nikhil Sethi and Jiao Sun and Luming Tang and Uri Alon and Xianghong Luo and Dian Yu and Abhishek Nayyar and Bryce Petrini and Will Truong and Vincent Hellendoorn and Nikolai Chinaev and Chris Alberti and Wei Wang and Jingcao Hu and Vahab Mirrokni and Ananth Balashankar and Avia Aharon and Aahil Mehta and Ahmet Iscen and Joseph Kready and Lucas Manning and Anhad Mohananey and Yuankai Chen and Anshuman Tripathi and Allen Wu and Igor Petrovski and Dawsen Hwang and Martin Baeuml and Shreyas Chandrakaladharan and Yuan Liu and Rey Coaguila and Maxwell Chen and Sally Ma and Pouya Tafti and Susheel Tatineni and Terry Spitz and Jiayu Ye and Paul Vicol and Mihaela Rosca and Adrià Puigdomènech and Zohar Yahav and Sanjay Ghemawat and Hanzhao Lin and Phoebe Kirk and Zaid Nabulsi and Sergey Brin and Bernd Bohnet and Ken Caluwaerts and Aditya Srikanth Veerubhotla and Dan Zheng and Zihang Dai and Petre Petrov and Yichong Xu and Ramin Mehran and Zhuo Xu and Luisa Zintgraf and Jiho Choi and Spurthi Amba Hombaiah and Romal Thoppilan and Sashank Reddi and Lukasz Lew and Li Li and Kellie Webster and KP Sawhney and Lampros Lamprou and Siamak Shakeri and Mayank Lunayach and Jianmin Chen and Sumit Bagri and Alex Salcianu and Ying Chen and Yani Donchev and Charlotte Magister and Signe Nørly and Vitor Rodrigues and Tomas Izo and Hila Noga and Joe Zou and Thomas Köppe and Wenxuan Zhou and Kenton Lee and Xiangzhu Long and Danielle Eisenbud and Anthony Chen and Connor Schenck and Chi Ming To and Peilin Zhong and Emanuel Taropa and Minh Truong and Omer Levy and Danilo Martins and Zhiyuan Zhang and Christopher Semturs and Kelvin Zhang and Alex Yakubovich and Pol Moreno and Lara McConnaughey and Di Lu and Sam Redmond and Lotte Weerts and Yonatan Bitton and Tiziana Refice and Nicolas Lacasse and Arthur Conmy and Corentin Tallec and Julian Odell and Hannah Forbes-Pollard and Arkadiusz Socala and Jonathan Hoech and Pushmeet Kohli and Alanna Walton and Rui Wang and Mikita Sazanovich and Kexin Zhu and Andrei Kapishnikov and Rich Galt and Matthew Denton and Ben Murdoch and Caitlin Sikora and Kareem Mohamed and Wei Wei and Uri First and Tim McConnell and Luis C. Cobo and James Qin and Thi Avrahami and Daniel Balle and Yu Watanabe and Annie Louis and Adam Kraft and Setareh Ariafar and Yiming Gu and Eugénie Rives and Charles Yoon and Andrei Rusu and James Cobon-Kerr and Chris Hahn and Jiaming Luo and Yuvein and Zhu and Niharika Ahuja and Rodrigo Benenson and Raphaël Lopez Kaufman and Honglin Yu and Lloyd Hightower and Junlin Zhang and Darren Ni and Lisa Anne Hendricks and Gabby Wang and Gal Yona and Lalit Jain and Pablo Barrio and Surya Bhupatiraju and Siva Velusamy and Allan Dafoe and Sebastian Riedel and Tara Thomas and Zhe Yuan and Mathias Bellaiche and Sheena Panthaplackel and Klemen Kloboves and Sarthak Jauhari and Canfer Akbulut and Todor Davchev and Evgeny Gladchenko and David Madras and Aleksandr Chuklin and Tyrone Hill and Quan Yuan and Mukundan Madhavan and Luke Leonhard and Dylan Scandinaro and Qihang Chen and Ning Niu and Arthur Douillard and Bogdan Damoc and Yasumasa Onoe and Fabian Pedregosa and Fred Bertsch and Chas Leichner and Joseph Pagadora and Jonathan Malmaud and Sameera Ponda and Andy Twigg and Oleksii Duzhyi and Jingwei Shen and Miaosen Wang and Roopal Garg and Jing Chen and Utku Evci and Jonathan Lee and Leon Liu and Koji Kojima and Masa Yamaguchi and Arunkumar Rajendran and AJ Piergiovanni and Vinodh Kumar Rajendran and Marco Fornoni and Gabriel Ibagon and Harry Ragan and Sadh MNM Khan and John Blitzer and Andrew Bunner and Guan Sun and Takahiro Kosakai and Scott Lundberg and Ndidi Elue and Kelvin Guu and SK Park and Jane Park and Arunachalam Narayanaswamy and Chengda Wu and Jayaram Mudigonda and Trevor Cohn and Hairong Mu and Ravi Kumar and Laura Graesser and Yichi Zhang and Richard Killam and Vincent Zhuang and Mai Giménez and Wael Al Jishi and Ruy Ley-Wild and Alex Zhai and Kazuki Osawa and Diego Cedillo and Jialu Liu and Mayank Upadhyay and Marcin Sieniek and Roshan Sharma and Tom Paine and Anelia Angelova and Sravanti Addepalli and Carolina Parada and Kingshuk Majumder and Avery Lamp and Sanjiv Kumar and Xiang Deng and Artiom Myaskovsky and Tea Sabolić and Jeffrey Dudek and Sarah York and Félix de Chaumont Quitry and Jiazhong Nie and Dee Cattle and Alok Gunjan and Bilal Piot and Waleed Khawaja and Seojin Bang and Simon Wang and Siavash Khodadadeh and Raghavender R and Praynaa Rawlani and Richard Powell and Kevin Lee and Johannes Griesser and GS Oh and Cesar Magalhaes and Yujia Li and Simon Tokumine and Hadas Natalie Vogel and Dennis Hsu and Arturo BC and Disha Jindal and Matan Cohen and Zi Yang and Junwei Yuan and Dario de Cesare and Tony Bruguier and Jun Xu and Monica Roy and Alon Jacovi and Dan Belov and Rahul Arya and Phoenix Meadowlark and Shlomi Cohen-Ganor and Wenting Ye and Patrick Morris-Suzuki and Praseem Banzal and Gan Song and Pranavaraj Ponnuramu and Fred Zhang and George Scrivener and Salah Zaiem and Alif Raditya Rochman and Kehang Han and Badih Ghazi and Kate Lee and Shahar Drath and Daniel Suo and Antonious Girgis and Pradeep Shenoy and Duy Nguyen and Douglas Eck and Somit Gupta and Le Yan and Joao Carreira and Anmol Gulati and Ruoxin Sang and Daniil Mirylenka and Emma Cooney and Edward Chou and Mingyang Ling and Cindy Fan and Ben Coleman and Guilherme Tubone and Ravin Kumar and Jason Baldridge and Felix Hernandez-Campos and Angeliki Lazaridou and James Besley and Itay Yona and Neslihan Bulut and Quentin Wellens and AJ Pierigiovanni and Jasmine George and Richard Green and Pu Han and Connie Tao and Geoff Clark and Chong You and Abbas Abdolmaleki and Justin Fu and Tongzhou Chen and Ashwin Chaugule and Angad Chandorkar and Altaf Rahman and Will Thompson and Penporn Koanantakool and Mike Bernico and Jie Ren and Andrey Vlasov and Sergei Vassilvitskii and Maciej Kula and Yizhong Liang and Dahun Kim and Yangsibo Huang and Chengxi Ye and Dmitry Lepikhin and Wesley Helmholz},
      year={2025},
      eprint={2507.06261},
      archivePrefix={arXiv},
      primaryClass={cs.CL},
      url={https://arxiv.org/abs/2507.06261}, 
}

@misc{liu2025voxtral,
      title={Voxtral}, 
      author={Alexander H. Liu and Andy Ehrenberg and Andy Lo and Clément Denoix and Corentin Barreau and Guillaume Lample and Jean-Malo Delignon and Khyathi Raghavi Chandu and Patrick von Platen and Pavankumar Reddy Muddireddy and Sanchit Gandhi and Soham Ghosh and Srijan Mishra and Thomas Foubert and Abhinav Rastogi and Adam Yang and Albert Q. Jiang and Alexandre Sablayrolles and Amélie Héliou and Amélie Martin and Anmol Agarwal and Antoine Roux and Arthur Darcet and Arthur Mensch and Baptiste Bout and Baptiste Rozière and Baudouin De Monicault and Chris Bamford and Christian Wallenwein and Christophe Renaudin and Clémence Lanfranchi and Darius Dabert and Devendra Singh Chaplot and Devon Mizelle and Diego de las Casas and Elliot Chane-Sane and Emilien Fugier and Emma Bou Hanna and Gabrielle Berrada and Gauthier Delerce and Gauthier Guinet and Georgii Novikov and Guillaume Martin and Himanshu Jaju and Jan Ludziejewski and Jason Rute and Jean-Hadrien Chabran and Jessica Chudnovsky and Joachim Studnia and Joep Barmentlo and Jonas Amar and Josselin Somerville Roberts and Julien Denize and Karan Saxena and Karmesh Yadav and Kartik Khandelwal and Kush Jain and Lélio Renard Lavaud and Léonard Blier and Lingxiao Zhao and Louis Martin and Lucile Saulnier and Luyu Gao and Marie Pellat and Mathilde Guillaumin and Mathis Felardos and Matthieu Dinot and Maxime Darrin and Maximilian Augustin and Mickaël Seznec and Neha Gupta and Nikhil Raghuraman and Olivier Duchenne and Patricia Wang and Patryk Saffer and Paul Jacob and Paul Wambergue and Paula Kurylowicz and Philomène Chagniot and Pierre Stock and Pravesh Agrawal and Rémi Delacourt and Romain Sauvestre and Roman Soletskyi and Sagar Vaze and Sandeep Subramanian and Saurabh Garg and Shashwat Dalal and Siddharth Gandhi and Sumukh Aithal and Szymon Antoniak and Teven Le Scao and Thibault Schueller and Thibaut Lavril and Thomas Robert and Thomas Wang and Timothée Lacroix and Tom Bewley and Valeriia Nemychnikova and Victor Paltz and Virgile Richard and Wen-Ding Li and William Marshall and Xuanyu Zhang and Yihan Wan and Yunhao Tang},
      year={2025},
      eprint={2507.13264},
      archivePrefix={arXiv},
      primaryClass={cs.SD},
      url={https://arxiv.org/abs/2507.13264}, 
}

@misc{microsoft2025phi4minitechnicalreportcompact,
      title={Phi-4-Mini Technical Report: Compact yet Powerful Multimodal Language Models via Mixture-of-LoRAs}, 
      author={Microsoft and : and Abdelrahman Abouelenin and Atabak Ashfaq and Adam Atkinson and Hany Awadalla and Nguyen Bach and Jianmin Bao and Alon Benhaim and Martin Cai and Vishrav Chaudhary and Congcong Chen and Dong Chen and Dongdong Chen and Junkun Chen and Weizhu Chen and Yen-Chun Chen and Yi-ling Chen and Qi Dai and Xiyang Dai and Ruchao Fan and Mei Gao and Min Gao and Amit Garg and Abhishek Goswami and Junheng Hao and Amr Hendy and Yuxuan Hu and Xin Jin and Mahmoud Khademi and Dongwoo Kim and Young Jin Kim and Gina Lee and Jinyu Li and Yunsheng Li and Chen Liang and Xihui Lin and Zeqi Lin and Mengchen Liu and Yang Liu and Gilsinia Lopez and Chong Luo and Piyush Madan and Vadim Mazalov and Arindam Mitra and Ali Mousavi and Anh Nguyen and Jing Pan and Daniel Perez-Becker and Jacob Platin and Thomas Portet and Kai Qiu and Bo Ren and Liliang Ren and Sambuddha Roy and Ning Shang and Yelong Shen and Saksham Singhal and Subhojit Som and Xia Song and Tetyana Sych and Praneetha Vaddamanu and Shuohang Wang and Yiming Wang and Zhenghao Wang and Haibin Wu and Haoran Xu and Weijian Xu and Yifan Yang and Ziyi Yang and Donghan Yu and Ishmam Zabir and Jianwen Zhang and Li Lyna Zhang and Yunan Zhang and Xiren Zhou},
      year={2025},
      eprint={2503.01743},
      archivePrefix={arXiv},
      primaryClass={cs.CL},
      url={https://arxiv.org/abs/2503.01743}, 
}

@inproceedings{wei-etal-2024-unveiling,
    title = "Unveiling Selection Biases: Exploring Order and Token Sensitivity in Large Language Models",
    author = "Wei, Sheng-Lun  and
      Wu, Cheng-Kuang  and
      Huang, Hen-Hsen  and
      Chen, Hsin-Hsi",
    editor = "Ku, Lun-Wei  and
      Martins, Andre  and
      Srikumar, Vivek",
    booktitle = "Findings of the Association for Computational Linguistics: ACL 2024",
    month = aug,
    year = "2024",
    address = "Bangkok, Thailand",
    publisher = "Association for Computational Linguistics",
    url = "https://aclanthology.org/2024.findings-acl.333/",
    doi = "10.18653/v1/2024.findings-acl.333",
    pages = "5598--5621",
}

@inproceedings{tatman-2017-gender,
    title = "Gender and Dialect Bias in {Y}ou{T}ube{'}s Automatic Captions",
    author = "Tatman, Rachael",
    editor = "Hovy, Dirk  and
      Spruit, Shannon  and
      Mitchell, Margaret  and
      Bender, Emily M.  and
      Strube, Michael  and
      Wallach, Hanna",
    booktitle = "Proceedings of the First {ACL} Workshop on Ethics in Natural Language Processing",
    month = apr,
    year = "2017",
    address = "Valencia, Spain",
    publisher = "Association for Computational Linguistics",
    url = "https://aclanthology.org/W17-1606/",
    doi = "10.18653/v1/W17-1606",
    pages = "53--59",
}

@misc{elghazaly2025exploringgenderdisparitiesautomatic,
      title={Exploring Gender Disparities in Automatic Speech Recognition Technology}, 
      author={Hend ElGhazaly and Bahman Mirheidari and Nafise Sadat Moosavi and Heidi Christensen},
      year={2025},
      eprint={2502.18434},
      archivePrefix={arXiv},
      primaryClass={cs.CL},
      url={https://arxiv.org/abs/2502.18434}, 
}

@article{10.1121/10.0024876,
    author = {Graham, Calbert and Roll, Nathan},
    title = {Evaluating OpenAI's Whisper ASR: Performance analysis across diverse accents and speaker traits},
    journal = {JASA Express Letters},
    volume = {4},
    number = {2},
    pages = {025206},
    year = {2024},
    month = {02},
    issn = {2691-1191},
    doi = {10.1121/10.0024876},
    url = {https://doi.org/10.1121/10.0024876},
    eprint = {https://pubs.aip.org/asa/jel/article-pdf/doi/10.1121/10.0024876/19692982/025206_1_10.0024876.pdf},
}

@inproceedings{ijcai2023p578,
  title     = {SQuAD-SRC: A Dataset for Multi-Accent Spoken Reading Comprehension},
  author    = {Tang, Yixuan and Tung, Anthony K.H:},
  booktitle = {Proceedings of the Thirty-Second International Joint Conference on
               Artificial Intelligence, {IJCAI-23}},
  publisher = {International Joint Conferences on Artificial Intelligence Organization},
  editor    = {Edith Elkind},
  pages     = {5206--5214},
  year      = {2023},
  month     = {8},
  note      = {Main Track},
  doi       = {10.24963/ijcai.2023/578},
  url       = {https://doi.org/10.24963/ijcai.2023/578},
}

@inproceedings{attanasio-etal-2024-twists,
    title = "Twists, Humps, and Pebbles: Multilingual Speech Recognition Models Exhibit Gender Performance Gaps",
    author = "Attanasio, Giuseppe  and
      Savoldi, Beatrice  and
      Fucci, Dennis  and
      Hovy, Dirk",
    editor = "Al-Onaizan, Yaser  and
      Bansal, Mohit  and
      Chen, Yun-Nung",
    booktitle = "Proceedings of the 2024 Conference on Empirical Methods in Natural Language Processing",
    month = nov,
    year = "2024",
    address = "Miami, Florida, USA",
    publisher = "Association for Computational Linguistics",
    url = "https://aclanthology.org/2024.emnlp-main.1188/",
    doi = "10.18653/v1/2024.emnlp-main.1188",
    pages = "21318--21340",
}

@inproceedings{babu22_interspeech,
  title     = {XLS-R: Self-supervised Cross-lingual Speech Representation Learning at Scale},
  author    = {Arun Babu and Changhan Wang and Andros Tjandra and Kushal Lakhotia and Qiantong Xu and Naman Goyal and Kritika Singh and Patrick {von Platen} and Yatharth Saraf and Juan Pino and Alexei Baevski and Alexis Conneau and Michael Auli},
  year      = {{2022}},
  booktitle = {{Interspeech 2022}},
  pages     = {{2278--2282}},
  doi       = {{10.21437/Interspeech.2022-143}},
  issn      = {{2958-1796}},
}

@inproceedings{harris-etal-2024-modeling,
    title = "Modeling Gender and Dialect Bias in Automatic Speech Recognition",
    author = "Harris, Camille  and
      Mgbahurike, Chijioke  and
      Kumar, Neha  and
      Yang, Diyi",
    editor = "Al-Onaizan, Yaser  and
      Bansal, Mohit  and
      Chen, Yun-Nung",
    booktitle = "Findings of the Association for Computational Linguistics: EMNLP 2024",
    month = nov,
    year = "2024",
    address = "Miami, Florida, USA",
    publisher = "Association for Computational Linguistics",
    url = "https://aclanthology.org/2024.findings-emnlp.890/",
    doi = "10.18653/v1/2024.findings-emnlp.890",
    pages = "15166--15184",
}

@inproceedings{Kulkarni_2024, series={interspeech 2024},
   title={Unveiling Biases while Embracing Sustainability: Assessing the Dual Challenges of Automatic Speech Recognition Systems},
   url={http://dx.doi.org/10.21437/Interspeech.2024-2494},
   DOI={10.21437/interspeech.2024-2494},
   booktitle={Interspeech 2024},
   publisher={ISCA},
   author={Kulkarni, Ajinkya and Kulkarni, Atharva and Couceiro, Miguel and Trancoso, Isabel},
   year={2024},
   month=sep, pages={4628--4632},
   collection={interspeech\_2024} }

@inproceedings{tadimeti-etal-2022-evaluation,
    title = "Evaluation of Off-the-shelf Speech Recognizers on Different Accents in a Dialogue Domain",
    author = "Tadimeti, Divya  and
      Georgila, Kallirroi  and
      Traum, David",
    editor = "Calzolari, Nicoletta  and
      B{\'e}chet, Fr{\'e}d{\'e}ric  and
      Blache, Philippe  and
      Choukri, Khalid  and
      Cieri, Christopher  and
      Declerck, Thierry  and
      Goggi, Sara  and
      Isahara, Hitoshi  and
      Maegaard, Bente  and
      Mariani, Joseph  and
      Mazo, H{\'e}l{\`e}ne  and
      Odijk, Jan  and
      Piperidis, Stelios",
    booktitle = "Proceedings of the Thirteenth Language Resources and Evaluation Conference",
    month = jun,
    year = "2022",
    address = "Marseille, France",
    publisher = "European Language Resources Association",
    url = "https://aclanthology.org/2022.lrec-1.645/",
    pages = "6001--6008",
}

@inproceedings{chan22b_interspeech,
  title     = {{Training and typological bias in ASR performance for world Englishes}},
  author    = {May Pik Yu Chan and June Choe and Aini Li and Yiran Chen and Xin Gao and Nicole Holliday},
  year      = {{2022}},
  booktitle = {{Interspeech 2022}},
  pages     = {{1273--1277}},
  doi       = {{10.21437/Interspeech.2022-10869}},
  issn      = {{2958-1796}},
}

@article{
doi10.1073/pnas.1915768117,
author = {Allison Koenecke  and Andrew Nam  and Emily Lake  and Joe Nudell  and Minnie Quartey  and Zion Mengesha  and Connor Toups  and John R. Rickford  and Dan Jurafsky  and Sharad Goel },
title = {Racial disparities in automated speech recognition},
journal = {Proceedings of the National Academy of Sciences},
volume = {117},
number = {14},
pages = {7684--7689},
year = {2020},
doi = {10.1073/pnas.1915768117},
URL = {https://www.pnas.org/doi/abs/10.1073/pnas.1915768117},
eprint = {https://www.pnas.org/doi/pdf/10.1073/pnas.1915768117},}

@inproceedings{singh-etal-2025-global,
    title = "Global {MMLU}: Understanding and Addressing Cultural and Linguistic Biases in Multilingual Evaluation",
    author = "Singh, Shivalika  and
      Romanou, Angelika  and
      Fourrier, Cl{\'e}mentine  and
      Adelani, David Ifeoluwa  and
      Ngui, Jian Gang  and
      Vila-Suero, Daniel  and
      Limkonchotiwat, Peerat  and
      Marchisio, Kelly  and
      Leong, Wei Qi  and
      Susanto, Yosephine  and
      Ng, Raymond  and
      Longpre, Shayne  and
      Ruder, Sebastian  and
      Ko, Wei-Yin  and
      Bosselut, Antoine  and
      Oh, Alice  and
      Martins, Andre  and
      Choshen, Leshem  and
      Ippolito, Daphne  and
      Ferrante, Enzo  and
      Fadaee, Marzieh  and
      Ermis, Beyza  and
      Hooker, Sara",
    editor = "Che, Wanxiang  and
      Nabende, Joyce  and
      Shutova, Ekaterina  and
      Pilehvar, Mohammad Taher",
    booktitle = "Proceedings of the 63rd Annual Meeting of the Association for Computational Linguistics (Volume 1: Long Papers)",
    month = jul,
    year = "2025",
    address = "Vienna, Austria",
    publisher = "Association for Computational Linguistics",
    url = "https://aclanthology.org/2025.acl-long.919/",
    doi = "10.18653/v1/2025.acl-long.919",
    pages = "18761--18799",
    ISBN = "979-8-89176-251-0",
}

@inproceedings{
nachmani2024spoken,
title={Spoken Question Answering and Speech Continuation Using Spectrogram-Powered {LLM}},
author={Eliya Nachmani and Alon Levkovitch and Roy Hirsch and Julian Salazar and Chulayuth Asawaroengchai and Soroosh Mariooryad and Ehud Rivlin and RJ Skerry-Ryan and Michelle Tadmor Ramanovich},
booktitle={The Twelfth International Conference on Learning Representations},
year={2024},
url={https://openreview.net/forum?id=izrOLJov5y}
}

@inproceedings{shih24b_interspeech,
  title     = {{GSQA: An End-to-End Model for Generative Spoken Question Answering}},
  author    = {Min-Han Shih and Ho-Lam Chung and Yu-Chi Pai and Ming-Hao Hsu and Guan-Ting Lin and Shang-Wen Li and Hung-yi Lee},
  year      = {2024},
  booktitle = {{Interspeech 2024}},
  pages     = {2970--2974},
  doi       = {10.21437/Interspeech.2024-1514},
  issn      = {2958-1796},
}

@inproceedings{
tang2024salmonn,
title={{SALMONN}: Towards Generic Hearing Abilities for Large Language Models},
author={Changli Tang and Wenyi Yu and Guangzhi Sun and Xianzhao Chen and Tian Tan and Wei Li and Lu Lu and Zejun MA and Chao Zhang},
booktitle={The Twelfth International Conference on Learning Representations},
year={2024},
url={https://openreview.net/forum?id=14rn7HpKVk}
}

@inproceedings{zhang-etal-2023-speechgpt,
    title = "{S}peech{GPT}: Empowering Large Language Models with Intrinsic Cross-Modal Conversational Abilities",
    author = "Zhang, Dong  and
      Li, Shimin  and
      Zhang, Xin  and
      Zhan, Jun  and
      Wang, Pengyu  and
      Zhou, Yaqian  and
      Qiu, Xipeng",
    editor = "Bouamor, Houda  and
      Pino, Juan  and
      Bali, Kalika",
    booktitle = "Findings of the Association for Computational Linguistics: EMNLP 2023",
    month = dec,
    year = "2023",
    address = "Singapore",
    publisher = "Association for Computational Linguistics",
    url = "https://aclanthology.org/2023.findings-emnlp.1055/",
    doi = "10.18653/v1/2023.findings-emnlp.1055",
    pages = "15757--15773",
}

@misc{rubenstein2023audiopalmlargelanguagemodel,
      title={AudioPaLM: A Large Language Model That Can Speak and Listen}, 
      author={Paul K. Rubenstein and Chulayuth Asawaroengchai and Duc Dung Nguyen and Ankur Bapna and Zalán Borsos and Félix de Chaumont Quitry and Peter Chen and Dalia El Badawy and Wei Han and Eugene Kharitonov and Hannah Muckenhirn and Dirk Padfield and James Qin and Danny Rozenberg and Tara Sainath and Johan Schalkwyk and Matt Sharifi and Michelle Tadmor Ramanovich and Marco Tagliasacchi and Alexandru Tudor and Mihajlo Velimirović and Damien Vincent and Jiahui Yu and Yongqiang Wang and Vicky Zayats and Neil Zeghidour and Yu Zhang and Zhishuai Zhang and Lukas Zilka and Christian Frank},
      year={2023},
      eprint={2306.12925},
      archivePrefix={arXiv},
      primaryClass={cs.CL},
      url={https://arxiv.org/abs/2306.12925}, 
}

@inproceedings{10.1145/3706599.3720162,
author = {Bendarkawi, Jad and Ponce, Ashley and Mata, Sean Chidozie and Aliu, Aminah and Liu, Yuhan and Zhang, Lei and Liaqat, Amna and Rao, Varun Nagaraj and Monroy-Hern\'{a}ndez, Andr\'{e}s},
title = {ConversAR: Exploring Embodied LLM-Powered Group Conversations in Augmented Reality for Second Language Learners},
year = {2025},
isbn = {9798400713958},
publisher = {Association for Computing Machinery},
address = {New York, NY, USA},
url = {https://doi.org/10.1145/3706599.3720162},
doi = {10.1145/3706599.3720162},
booktitle = {Proceedings of the Extended Abstracts of the CHI Conference on Human Factors in Computing Systems},
articleno = {153},
numpages = {11},
location = {
},
series = {CHI EA '25}
}

@misc{ma2025assessmentl2oralproficiency,
      title={Assessment of L2 Oral Proficiency using Speech Large Language Models}, 
      author={Rao Ma and Mengjie Qian and Siyuan Tang and Stefano Bannò and Kate M. Knill and Mark J. F. Gales},
      year={2025},
      eprint={2505.21148},
      archivePrefix={arXiv},
      primaryClass={cs.CL},
      url={https://arxiv.org/abs/2505.21148}, 
}

@inproceedings{ICLR2024_37771cc0,
 author = {Bel\'{e}m, Catarina and Seshadri, Preethi and Razeghi, Yasaman and Singh, Sameer},
 booktitle = {International Conference on Representation Learning},
 editor = {B. Kim and Y. Yue and S. Chaudhuri and K. Fragkiadaki and M. Khan and Y. Sun},
 pages = {12876--12915},
 title = {Are Models Biased on Text without Gender-related Language?},
 url = {https://proceedings.iclr.cc/paper_files/paper/2024/file/37771cc0be272368102a37f202bb88d8-Paper-Conference.pdf},
 volume = {2024},
 year = {2024}
}

@inproceedings{
vo2025bscore,
title={B-score: Detecting biases in large language models using response history},
author={An Vo and Mohammad Reza Taesiri and Daeyoung Kim and Anh Totti Nguyen},
booktitle={Forty-second International Conference on Machine Learning},
year={2025},
url={https://openreview.net/forum?id=kl7SbPfBsB}
}

@misc{hofmann2024dialectprejudicepredictsai,
      title={Dialect prejudice predicts AI decisions about people's character, employability, and criminality}, 
      author={Valentin Hofmann and Pratyusha Ria Kalluri and Dan Jurafsky and Sharese King},
      year={2024},
      eprint={2403.00742},
      archivePrefix={arXiv},
      primaryClass={cs.CL},
      url={https://arxiv.org/abs/2403.00742}, 
}

\clearpage
\appendix

\section{Cost Analysis}
All experiments involving TTS generation and model inference via the Gemini API incurred a total cost of under \$550~USD. 
For the other models, the APIs provided by Mistral and NVIDIA were temporarily free during the experiment period.

\section{Dataset Construction Details}
\label{sec:appendix-dataset}

\subsection{Question Rewriting}
\label{sec:rewriting-details}
To perform the rewriting of questions and options, we employ the \texttt{GPT OSS 120B} model via the NVIDIA API, which ensures both stability and scalability. The model is prompted with task-specific instructions shown in Figure~\ref{fig:spoken_rendering_prompt}, which enforce eight conversion rules. These rules cover aspects such as reading mathematical expressions (e.g., \texttt{"$x^2 + y^2$"}), disambiguating domain-specific terms using subject context (e.g., \texttt{"Na"} $\rightarrow$ \texttt{"sodium"} in chemistry), handling numbers and units (e.g., \texttt{"3kg"} $\rightarrow$ \texttt{"three kilograms"}), interpreting parentheses, and rendering placeholders like \texttt{"BLANK"} appropriately across different languages. This step ensures the generation of high-quality spoken-readable text prior to audio synthesis.
\subsection{TTS Generation Prompt}
\label{sec:tts_generation_prompt}
Figure~\ref{fig:tts-prompt} illustrates the prompt template used for TTS synthesis, showing how textual instructions, speaker characteristics, and prosodic cues are combined to guide the model toward more natural, expressive, and context-aware speech.
\subsection{Quality Assessment of Rewriting}
\label{sec:rewriting_qc}
As we mentioned in Section ~\ref{sec:quality-check-main}, to evaluate the reliability of the rewritten questions, we manually inspected a subset of automatically flagged cases.
Table~\ref{tab:rewrite-qc} summarizes the proportion of flagged instances and verified true errors across English, Chinese, and Korean.

\subsection{Quality Assessment of Voice Generation with Stratified Sampling}
\label{sec:quality_assessment}
Sampling for quality assessment was stratified for representativeness and diversity. Within each WER interval, we pooled all accents and allocated per-accent quotas proportional to their sample counts using the largest-remainder method. For each question–answer pair, we selected one item (preferring the question-description row), then sampled in a subject-wise round-robin to balance subject coverage. When an interval lacked enough unique subjects or questions to meet the quota, we filled the remainder from the available pool.

\begin{figure}[t]
\centering
\begin{tcolorbox}
\small
\texttt{Read this text in \{language\} with a \{accent\} accent: \{text\}}
\end{tcolorbox}
\caption{Prompt template used for TTS synthesis. The placeholder \texttt{\{text\}} is substituted with the rewritten question or answer option.}
\label{fig:tts-prompt}
\end{figure}
\begin{table}[t]
  \centering
  \small
  \setlength{\tabcolsep}{4pt}
  \begin{tabular}{lcc}
    \toprule
    \textbf{Language} & \textbf{Flagged Cases(\%)} & \textbf{Verified Errors (\%)} \\
    \midrule
    English & 477 (23.85\%) & 20 (1.00\%) \\
    Chinese & 494 (24.70\%) & 93 (4.65\%) \\
    Korean  & 577 (28.85\%) & 79 (3.95\%) \\
    \bottomrule
  \end{tabular}
  \caption{Quality control statistics for rewritten questions. 
  The table reports the number and percentage of automatically flagged cases and human-verified true errors.}
  \label{tab:rewrite-qc}
\end{table}

\section{Experimental Setup Details}
\subsection{Models}
\label{sec:appendix-model}
As described in Section~\ref{sec:model-main}, we evaluate nine large multimodal models across both commercial and open-source settings. Table~\ref{tab:evaluated_models} summarizes these models, including their API endpoints provided by Google\footnote{\url{http://aistudio.google.com}}, NVIDIA\footnote{\url{https://build.nvidia.com}}, and Mistral\footnote{\url{https://console.mistral.ai}}, along with their corresponding providers.

\subsection{Audio Concatenation}
\label{appendix-audio-concat}
For each question, we constructed an audio query by concatenating the question and four option descriptions with instruction tokens (\textit{"question"}, \textit{"A" -- "D"}) pre-generateed by the TTS model. Instruction tokens were rendered in American, British, and Indian accents for English, and in American accent for Chinese and Korean.

The concatenation process followed the designated order: each instruction token placed before its corresponding content; four orderings (\textit{original}, \textit{reversed}, \textit{order backward}, \textit{token backward}) were implemented, with fixed pauses inserted between segments for clarity. To accommodate input length constraints (30s) in models such as \texttt{Gemma} and \texttt{Phi 4}, each final audio was further segmented into fixed-length chunks and exported as waveform files. This ensured consistent and compatible inputs across all experimental variables. This ensured consistent inputs across all experimental variables.
\begin{figure*}[htbp]
\centering
\begin{tcolorbox}[width=\textwidth]
\small
\textbf{[System]} \\
Convert the inputs into how a native \{language\} speaker would read them. 
Output only the results. Do not explain reasoning or revise the plain text content.

Inputs \\
Subject: \texttt{<subject>} \\
Question: \texttt{<question>} \\
Options: A) \texttt{<option\_a>}, B) \texttt{<option\_b>}, C) \texttt{<option\_c>}, D) \texttt{<option\_d>} \\

The conversion should follow the 8 rules:
\begin{enumerate}
  \item Use the "Subject" to disambiguate domain terms or abbreviations (e.g., "ms" read as "millisecond" in Physics, "HPO42-" read as "hydrogen phosphate" in Chemistry, 'P' read as "phosphorus" in Chemistry).
  \item Read math expressions naturally. Examples:  
        "f(x)" $\to$ "f of x";  
        $Z_n$ $\to$ "integers modulo n";  
        $(12)(123)$ $\to$ "the permutation consisting of the cycle one-two, and the cycle one-two-three".  
        If letter case matters, read it explicitly: "O(n) and o(n)" $\to$ "big-O of n and little-o of n".
  \item Read numbers and units cleanly with correct plurals.
  \item Read parentheses or brackets only if they affect meaning.
  \item Read Roman-numeral labels (I, II, i, ii ...) as labels: "Roman numeral one", "Roman numeral two", etc.
  \item  Read "BLANK" in place of blank markers \verb|"_____"|; 
number multiple blanks ("BLANK 1", "BLANK 2"); keep surrounding grammar; never guess the answer.
  \item All conversions must be written directly in the \{language\} specified by the instructions. 
When the prompt contains "BLANK", render it in the language indicated 
(e.g., Chinese: \begin{CJK}{UTF8}{gbsn}空格\end{CJK}; 
Japanese: \begin{CJK}{UTF8}{min}空白\end{CJK}; 
Korean: \begin{CJK}{UTF8}{mj}공백\end{CJK}). 
All other text must also be in the same language.
\end{enumerate}

Output format (XML):
\begin{verbatim}
<output>
  <question>&lt;spoken rendering of the question&gt;</question>
  <option_a>&lt;spoken rendering of option A&gt;</option_a>
  <option_b>&lt;spoken rendering of option B&gt;</option_b>
  <option_c>&lt;spoken rendering of option C&gt;</option_c>
  <option_d>&lt;spoken rendering of option D&gt;</option_d>
</output>
\end{verbatim}
\end{tcolorbox}

\caption{Prompt for spoken-style rendering of MMLU-style items}
\label{fig:spoken_rendering_prompt}
\end{figure*}

\begin{table*}[htbp]
\centering
\small
\setlength{\tabcolsep}{8pt}
\begin{tabularx}{\linewidth}{X l X} 
\toprule
\textbf{Model} & \textbf{API Endpoint} & \textbf{Provider} \\
\midrule
\multicolumn{3}{c}{\textit{Commercial APIs}} \\
\midrule
Gemini 2.5 Flash       & \texttt{gemini-2.5-flash}        & Google \\
Gemini 2.5 Flash Lite  & \texttt{gemini-2.5-flash-lite}   & Google \\
Gemini 2.0 Flash       & \texttt{gemini-2.0-flash}        & Google \\
Gemini 2.0 Flash Lite  & \texttt{gemini-2.0-flash-lite}   & Google \\
\midrule
\multicolumn{3}{c}{\textit{Open-source Models}} \\
\midrule
Gemma 3n E4B          & \texttt{google/gemma-3n-e4b-it}  & Nvidia NIM \\
Gemma 3n E2B         & \texttt{google/gemma-3n-e2b-it}  & Nvidia NIM \\
Voxtral Small          & \texttt{voxtral-small-2507}      & Mistral \\
Voxtral Mini           & \texttt{voxtral-mini-2507}       & Mistral \\
Phi 4 Multimodal      & \texttt{microsoft/phi-4-multimodal-instruct} & Nvidia NIM \\
\bottomrule
\end{tabularx}
\caption{Evaluated models.}
\label{tab:evaluated_models}
\end{table*}

\clearpage
\subsection{Model Inference and Post-processing}
\label{sec:model-inference-and-post-processing}
All models were queried through their official APIs with deterministic greedy decoding (temperature set to zero and candidate count set to one) to eliminate randomness. A unified text prompt was used to standardize outputs as shown in Figure~\ref{fig:standard-prompt}
For chain-of-thought (CoT) inference, we used the prompt as shown in Figure~\ref{fig:cot-prompt} to standardize outputs. The maximum output length was capped at 4k tokens, sufficient to cover multiple-choice responses.

Model responses were post-processed to extract the final answer letter via pattern matching (e.g., \texttt{Answer:[[A]]}, \texttt{Answer: A}). The letter was mapped to an index per \textit{option order}, and invalid outputs were marked as parsing failures.

\subsection{Derivation of Fleiss' \texorpdfstring{$\kappa$}{kappa}}
\label{app:fleiss_kappa_derivation}

\paragraph{Setup.}
Fix a variable $v$, consider a set of items constructed by the other variables combination, indexed by $i=1,\dots,I$, where $I$ denotes the number of all variable combination except for $v$. 
Each item is assigned to one of $j$ categorical (A, B, C, D) by $n_i$ ratings.
In our application, the $n_i$ ratings arise from the same model answering item $i$ under different levels of $v$. 
Let $n_{ij}$ denote the number of ratings that chose category $j$ for item $i$, so that $\sum_{j=1}^{J} n_{ij} = n_i$.

\paragraph{Observed agreement per item.}
For item $i$, the proportion of agreeing rater-pairs is
\begin{equation}
P_i = \frac{1}{n_i(n_i-1)} \sum_{j=1}^4 n_{ij}(n_{ij}-1),
\label{eq:Pi}
\end{equation}
since each category-$j$ contributes $\binom{n_{ij}}{2}$ agreeing pairs and there are $\binom{n_i}{2}$ total unordered pairs.
\paragraph{Expected agreement under chance.}
Under the usual "random assignment with fixed marginals model", two independently drawn ratings match with probability
\begin{equation}
P_e \;=\; \sum_{j=1}^{J} p_j^2, \text{ where } p_j \;=\; 
\frac{\sum_{i=1}^{I} n_{ij}}{\sum_{i=1}^{I} n_i}.
\label{eq:Pe}
\end{equation}

\begin{figure}[t]
\begin{center}
\begin{tcolorbox}
\small
\texttt{Answer the following question. Output only "Answer:[[LETTER]]", where LETTER is one of A, B, C, D.}
\end{tcolorbox}
\captionof{figure}{Standard Prompt template}
\label{fig:standard-prompt}
\end{center}
\end{figure}

\begin{figure}[t]
\begin{center}
\begin{tcolorbox}
\small
\texttt{Answer the following question. Let's think step by step, and write concise reasoning. The last answer line should be "Answer:[[LETTER]]", where LETTER is one of A, B, C, D.}
\end{tcolorbox}
\captionof{figure}{CoT Prompt template}
\label{fig:cot-prompt}
\end{center}
\end{figure}

\paragraph{Averaging observed agreement.}
Aggregate the per-item agreements in \eqref{eq:Pi} by the number of ratings:
\begin{equation}
\bar P \;=\; \frac{\sum_{i=1}^{I} n_i\, P_i}{\sum_{i=1}^{I} n_i}.
\label{eq:Pbar}
\end{equation}
This weighting treats each rating pair equally across items when $n_i$ varies.

\paragraph{Fleiss' Kappa.}
Fleiss' $\kappa$ standardizes the excess agreement over chance:
\begin{equation}
\kappa \;=\; \frac{\bar P - P_e}{1 - P_e},
\label{eq:kappa}
\end{equation}
with $\kappa=1$ if $\bar P=1$ (perfect agreement), $\kappa=0$ if $\bar P=P_e$ (no better than chance), and $\kappa<0$ when observed agreement falls below chance.

\section{Detailed Experiment Results}
\label{sec:appendix-detailed results}

\begin{figure*}[t]
    \centering
    \includegraphics[width=\linewidth]{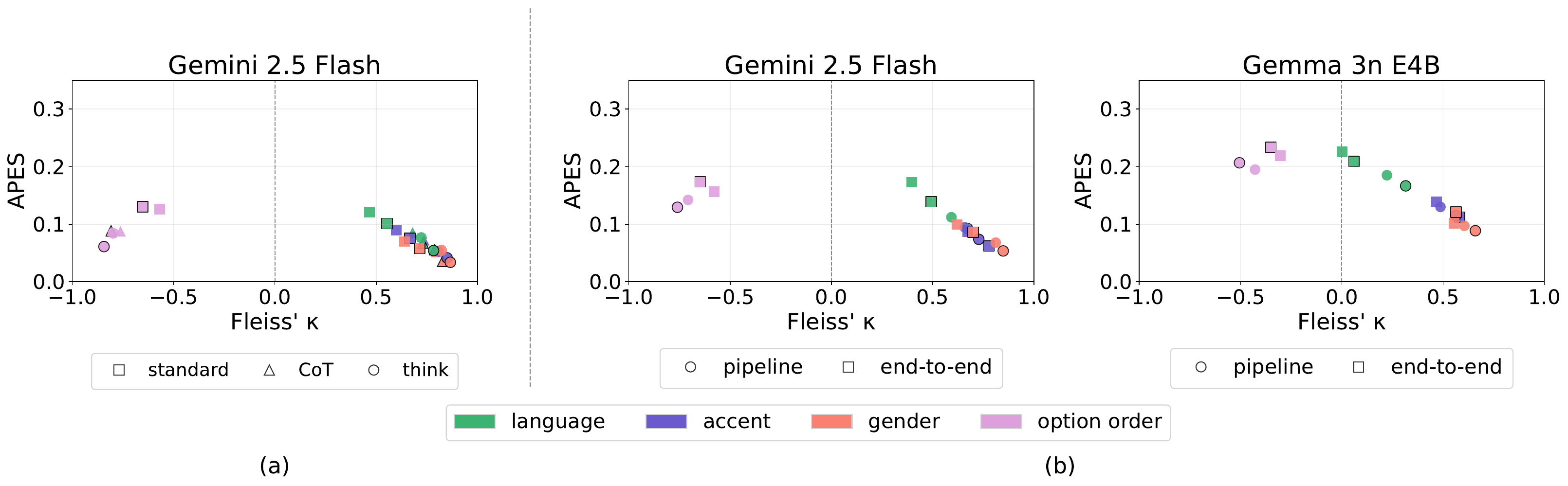}
    \caption{CS/CA stratified APES--$\kappa$ analysis under cross-variable perturbations (language, accent, gender, and option order). Unboxed markers denote CS items, while boxed markers denote CA items.}
    \label{fig:new_figs_figure_three_panel}
\end{figure*}
\begin{figure*}[t]
    \centering
    \includegraphics[width=\linewidth]{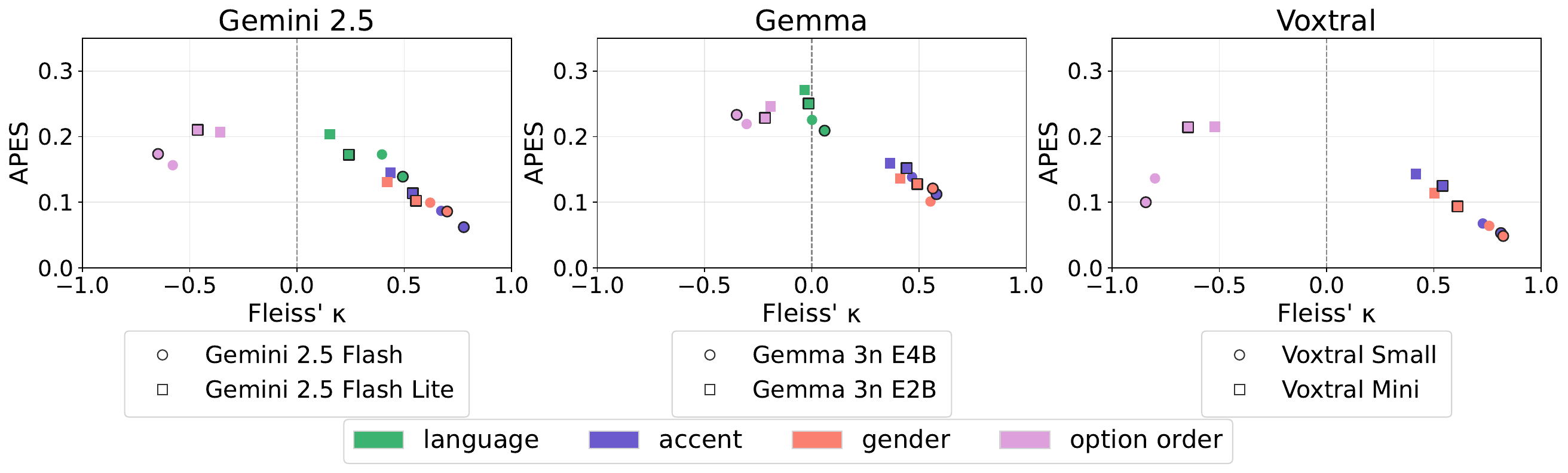}
    \caption{Grouped APES--$\kappa$ robustness plots across model families under cross-variable perturbations. Unboxed markers denote CS items, while boxed markers denote CA items.}
    \label{fig:new_figs_kappa_vs_apes_quadrants_groups}
\end{figure*}
\begin{table*}[t]
\footnotesize
\centering
\setlength{\tabcolsep}{4pt}
\renewcommand{\arraystretch}{1.15}
\begin{tabular}{cccc cccc cccc}
\toprule
\multicolumn{4}{c}{Chinese} & \multicolumn{4}{c}{English} & \multicolumn{4}{c}{Korean} \\
\cmidrule(lr){1-4} \cmidrule(lr){5-8} \cmidrule(lr){9-12}
input & original & reversed & $\Delta$
& input & original & reversed & $\Delta$
& input & original & reversed & $\Delta$ \\
\midrule
Beijing  \textcolor{magenta}{\female} & \textbf{46.75} & 45.00 & \colorbox{green!30}{1.75} 
& American  \textcolor{magenta}{\female} & \textbf{62.25} & 61.00 & \colorbox{green!30}{1.25} 
& Jeolla  \textcolor{magenta}{\female} & \textbf{43.75} & 42.50 & \colorbox{green!30}{1.25} \\
Beijing  \textcolor{blue}{\male} & \textbf{44.75} & 41.75 & \colorbox{green!30}{3.00} 
& American  \textcolor{blue}{\male} & \textbf{61.50} & 61.75 & \colorbox{blue!30}{-0.25} 
& Jeolla  \textcolor{blue}{\male} & \textbf{45.75} & 44.00 & \colorbox{green!30}{1.75} \\
Northeastern  \textcolor{magenta}{\female} & \textbf{43.50} & 40.00 & \colorbox{green!30}{3.50} 
& British  \textcolor{magenta}{\female} & \textbf{61.75} & 61.00 & \colorbox{green!30}{0.75} 
& Seoul  \textcolor{magenta}{\female} & \textbf{44.50} & 43.50 & \colorbox{green!30}{1.00} \\
Northeastern  \textcolor{blue}{\male} & \textbf{41.60} & 40.35 & \colorbox{green!30}{1.25} 
& British  \textcolor{blue}{\male} & \textbf{60.75} & 58.50 & \colorbox{green!30}{2.25} 
& Seoul  \textcolor{blue}{\male} & \textbf{46.00} & 44.75 & \colorbox{green!30}{1.25} \\
 &  &  &  
& Indian  \textcolor{magenta}{\female} & \textbf{64.00} & 60.50 & \colorbox{green!30}{3.50} 
&  &  &  &  \\
 &  &  &  
& Indian  \textcolor{blue}{\male} & \textbf{59.25} & 58.00 & \colorbox{green!30}{1.25} 
&  &  &  &  \\
\bottomrule
\end{tabular}
\caption{Accuracy comparison across option-order settings for \textit{Gemini 2.5 Flash Lite}, grouped by language, accent, and gender.}
\label{tab:accuracy-gemini-2.5-flash-lite}
\end{table*}
\begin{table*}[t]
\footnotesize
\centering
\setlength{\tabcolsep}{4pt}
\renewcommand{\arraystretch}{1.15}
\begin{tabular}{cccc cccc cccc}
\toprule
\multicolumn{4}{c}{Chinese} & \multicolumn{4}{c}{English} & \multicolumn{4}{c}{Korean} \\
\cmidrule(lr){1-4} \cmidrule(lr){5-8} \cmidrule(lr){9-12}
input & original & reversed & $\Delta$
& input & original & reversed & $\Delta$
& input & original & reversed & $\Delta$ \\
\midrule
Beijing  \textcolor{magenta}{\female} & \textbf{70.75} & 67.75 & \colorbox{green!30}{3.00} 
& American  \textcolor{magenta}{\female} & \textbf{83.75} & 80.75 & \colorbox{green!30}{3.00} 
& Jeolla  \textcolor{magenta}{\female} & \textbf{68.00} & 65.25 & \colorbox{green!30}{2.75} \\
Beijing  \textcolor{blue}{\male} & 69.00 & \textbf{69.50} & \colorbox{blue!30}{-0.50} 
& American  \textcolor{blue}{\male} & \textbf{83.25} & 80.00 & \colorbox{green!30}{3.25} 
& Jeolla  \textcolor{blue}{\male} & 66.00 & \textbf{67.00} & \colorbox{blue!30}{-1.00} \\
Northeastern  \textcolor{magenta}{\female} & \textbf{68.25} & 66.00 & \colorbox{green!30}{2.25} 
& British  \textcolor{magenta}{\female} & \textbf{81.75} & 80.50 & \colorbox{green!30}{1.25} 
& Seoul  \textcolor{magenta}{\female} & \textbf{70.25} & 68.25 & \colorbox{green!30}{2.00} \\
Northeastern  \textcolor{blue}{\male} & 65.50 & \textbf{69.25} & \colorbox{blue!30}{-3.75} 
& British  \textcolor{blue}{\male} & \textbf{81.75} & 78.75 & \colorbox{green!30}{3.00} 
& Seoul  \textcolor{blue}{\male} & \textbf{67.25} & 65.00 & \colorbox{green!30}{2.25} \\
 &  &  &  
& Indian  \textcolor{magenta}{\female} & \textbf{81.50} & 80.00 & \colorbox{green!30}{1.50} 
&  &  &  &  \\
 &  &  &  
& Indian  \textcolor{blue}{\male} & \textbf{83.00} & 80.00 & \colorbox{green!30}{3.00} 
&  &  &  &  \\
\bottomrule
\end{tabular}
\caption{Accuracy comparison across option-order settings for \textit{Gemini 2.0 Flash}, grouped by language, accent, and gender.}
\label{tab:accuracy-gemini-2.0-flash}
\end{table*}
\begin{table*}[t]
\footnotesize
\centering
\setlength{\tabcolsep}{4pt}
\renewcommand{\arraystretch}{1.15}
\begin{tabular}{cccc cccc cccc}
\toprule
\multicolumn{4}{c}{Chinese} & \multicolumn{4}{c}{English} & \multicolumn{4}{c}{Korean} \\
\cmidrule(lr){1-4} \cmidrule(lr){5-8} \cmidrule(lr){9-12}
input & original & reversed & $\Delta$
& input & original & reversed & $\Delta$
& input & original & reversed & $\Delta$ \\
\midrule
Beijing  \textcolor{magenta}{\female} & \textbf{65.50} & 63.50 & \colorbox{green!30}{2.00} & American  \textcolor{magenta}{\female} & \textbf{79.75} & 77.00 & \colorbox{green!30}{2.75} & Jeolla  \textcolor{magenta}{\female} & 66.50 & \textbf{66.75} & \colorbox{blue!30}{-0.25} \\
Beijing  \textcolor{blue}{\male} & \textbf{66.25} & 64.75 & \colorbox{green!30}{1.50} & American  \textcolor{blue}{\male} & 77.75 & \textbf{78.25} & \colorbox{blue!30}{-0.50} & Jeolla  \textcolor{blue}{\male} & \textbf{62.75} & 61.25 & \colorbox{green!30}{1.50} \\
Northeastern  \textcolor{magenta}{\female} & \textbf{66.75}& 61.75 & \colorbox{green!30}{5.00} & British  \textcolor{magenta}{\female} & \textbf{78.00} & 74.25 & \colorbox{green!30}{3.75} & Seoul  \textcolor{magenta}{\female} & \textbf{64.75} & 64.00 & \colorbox{green!30}{0.75} \\
Northeastern  \textcolor{blue}{\male} & 65.00 & \textbf{65.50} & \colorbox{blue!30}{-0.50} & British  \textcolor{blue}{\male} & \textbf{78.50} & 74.75 & \colorbox{green!30}{3.75} & Seoul  \textcolor{blue}{\male} & \textbf{66.25} & 62.50 & \colorbox{green!30}{3.75} \\
 &  &  &  & Indian  \textcolor{magenta}{\female} & \textbf{79.00} & 77.25 & \colorbox{green!30}{1.75} &  &  &  &  \\
 &  &  &  & Indian  \textcolor{blue}{\male} & \textbf{77.75} & 77.25 & \colorbox{green!30}{0.50} &  &  &  &  \\
\bottomrule
\end{tabular}
\caption{Accuracy comparison across option-order settings for \textit{Gemini 2.0 Flash Lite}, grouped by language, accent, and gender.}
\label{tab:accuracy-gemini-2.0-flash-lite}
\end{table*}
\begin{table*}[t]
\footnotesize
\centering
\setlength{\tabcolsep}{4pt}
\renewcommand{\arraystretch}{1.15}
\begin{tabular}{cccc cccc cccc}
\toprule
\multicolumn{4}{c}{Chinese} & \multicolumn{4}{c}{English} & \multicolumn{4}{c}{Korean} \\
\cmidrule(lr){1-4} \cmidrule(lr){5-8} \cmidrule(lr){9-12}
input & original & reversed & $\Delta$
& input & original & reversed & $\Delta$
& input & original & reversed & $\Delta$ \\
\midrule
Beijing  \textcolor{magenta}{\female} & \textbf{37.00} & 34.75 & \colorbox{green!30}{2.25} 
& American  \textcolor{magenta}{\female} & \textbf{59.75} & 53.25 & \colorbox{green!30}{6.50} 
& Jeolla  \textcolor{magenta}{\female} & \textbf{36.25} & 32.25 & \colorbox{green!30}{4.00} \\
Beijing  \textcolor{blue}{\male} & 34.25 & \textbf{35.00} & \colorbox{blue!30}{-0.75} 
& American  \textcolor{blue}{\male} & \textbf{59.25} & 56.75 & \colorbox{green!30}{2.50} 
& Jeolla  \textcolor{blue}{\male} & \textbf{35.75} & 33.50 & \colorbox{green!30}{2.25} \\
Northeastern  \textcolor{magenta}{\female} & 33.50 & \textbf{34.75} & \colorbox{blue!30}{-1.25} 
& British  \textcolor{magenta}{\female} & \textbf{59.75} & 52.00 & \colorbox{green!30}{7.75} 
& Seoul  \textcolor{magenta}{\female} & 33.25 & \textbf{34.50} & \colorbox{blue!30}{-1.25} \\
Northeastern  \textcolor{blue}{\male} & 37.00 & \textbf{37.50} & \colorbox{blue!30}{-0.50} 
& British  \textcolor{blue}{\male} & \textbf{61.25} & 51.00 & \colorbox{green!30}{10.25} 
& Seoul  \textcolor{blue}{\male} & \textbf{37.25} & 36.00 & \colorbox{green!30}{1.25} \\
 &  &  &  
& Indian  \textcolor{magenta}{\female} & \textbf{56.75} & 53.50 & \colorbox{green!30}{3.25} 
&  &  &  &  \\
 &  &  &  
& Indian  \textcolor{blue}{\male} & \textbf{60.00} & 51.50 & \colorbox{green!30}{8.50} 
&  &  &  &  \\
\bottomrule
\end{tabular}
\caption{Accuracy comparison across option-order settings for \textit{Gemma 3n E4B}, grouped by language, accent, and gender.}
\label{tab:accuracy-googlegemma-3n-e4b-it}
\end{table*}
\begin{table*}[t]
\footnotesize
\centering
\setlength{\tabcolsep}{4pt}
\renewcommand{\arraystretch}{1.15}
\begin{tabular}{cccc cccc cccc}
\toprule
\multicolumn{4}{c}{Chinese} & \multicolumn{4}{c}{English} & \multicolumn{4}{c}{Korean} \\
\cmidrule(lr){1-4} \cmidrule(lr){5-8} \cmidrule(lr){9-12}
input & original & reversed & $\Delta$
& input & original & reversed & $\Delta$
& input & original & reversed & $\Delta$ \\
\midrule
Beijing  \textcolor{magenta}{\female} & \textbf{33.75} & 30.50 & \colorbox{green!30}{3.25} 
& American  \textcolor{magenta}{\female} & \textbf{50.50} & 46.25 & \colorbox{green!30}{4.25} 
& Jeolla  \textcolor{magenta}{\female} & \textbf{32.75} & 29.00 & \colorbox{green!30}{3.75} \\
Beijing  \textcolor{blue}{\male} & \textbf{37.25} & 28.00 & \colorbox{green!30}{9.25} 
& American  \textcolor{blue}{\male} & \textbf{52.00} & 48.50 & \colorbox{green!30}{3.50} 
& Jeolla  \textcolor{blue}{\male} & \textbf{31.50} & 30.25 & \colorbox{green!30}{1.25} \\
Northeastern  \textcolor{magenta}{\female} & \textbf{32.50} & 29.50 & \colorbox{green!30}{3.00} 
& British  \textcolor{magenta}{\female} & \textbf{51.00} & 47.75 & \colorbox{green!30}{3.25} 
& Seoul  \textcolor{magenta}{\female} & \textbf{34.25} & 33.50 & \colorbox{green!30}{0.75} \\
Northeastern  \textcolor{blue}{\male} & \textbf{35.00} & 29.00 & \colorbox{green!30}{6.00} 
& British  \textcolor{blue}{\male} & \textbf{51.25} & 50.50 & \colorbox{green!30}{0.75} 
& Seoul  \textcolor{blue}{\male} & 28.75 & \textbf{33.50} & \colorbox{blue!30}{-4.75} \\
 &  &  &  
& Indian  \textcolor{magenta}{\female} & 48.25 & \textbf{48.75} & \colorbox{blue!30}{-0.50} 
&  &  &  &  \\
 &  &  &  
& Indian  \textcolor{blue}{\male} & \textbf{49.00} & 47.50 & \colorbox{green!30}{1.50} 
&  &  &  &  \\
\bottomrule
\end{tabular}
\caption{Accuracy comparison across option-order settings for \textit{Gemma 3n E2B}, grouped by language, accent, and gender.}
\label{tab:accuracy-googlegemma-3n-e2b-it}
\end{table*}
\begin{table*}[t]
\footnotesize
\centering
\setlength{\tabcolsep}{4pt}
\renewcommand{\arraystretch}{1.15}
\begin{tabular}{cccc cccc}
\toprule
\multicolumn{4}{c}{Chinese} & \multicolumn{4}{c}{English}\\
\cmidrule(lr){1-4} \cmidrule(lr){5-8}
input & original & reversed & $\Delta$
& input & original & reversed & $\Delta$\\
\midrule
Beijing  \textcolor{magenta}{\female} & \textbf{39.10} & 33.83 & \colorbox{green!30}{5.27} 
& American  \textcolor{magenta}{\female} & \textbf{48.99} & 43.47 & \colorbox{green!30}{5.52} \\
Beijing  \textcolor{blue}{\male} & \textbf{33.67} & 31.16 & \colorbox{green!30}{2.51} 
& American  \textcolor{blue}{\male} & \textbf{46.12} & 42.86 & \colorbox{green!30}{3.26}\\
Northeastern  \textcolor{magenta}{\female} & \textbf{34.84} & 30.08 & \colorbox{green!30}{4.76} 
& British  \textcolor{magenta}{\female} & \textbf{46.25} & 45.25 & \colorbox{green!30}{1.00} \\
Northeastern  \textcolor{blue}{\male} & \textbf{33.75} & 33.50 & \colorbox{green!30}{0.25} 
& British  \textcolor{blue}{\male} & \textbf{45.75} & 43.75 & \colorbox{green!30}{2.00}  \\
 &  &  &  
& Indian  \textcolor{magenta}{\female} & \textbf{47.75} & 44.75 & \colorbox{green!30}{3.00} \\
 &  &  &  
& Indian  \textcolor{blue}{\male} & 42.46 & \textbf{43.97} & \colorbox{blue!30}{-1.51}\\
\bottomrule
\end{tabular}
\caption{Accuracy comparison across option-order settings for \textit{Phi 4 Multimodal}, grouped by language, accent, and gender.}
\label{tab:accuracy-microsoftphi-4-multimodal-instruct}
\end{table*}
\begin{table*}[t]
\footnotesize
\centering
\setlength{\tabcolsep}{4pt}
\renewcommand{\arraystretch}{1.15}
\begin{subtable}[t]{0.48\textwidth}
    \centering
    \begin{tabular}{cccc}
    \toprule
    \multicolumn{4}{c}{English} \\
    \cmidrule(lr){1-4}
    input & original & reversed & $\Delta$ \\
    \midrule
    American  \textcolor{magenta}{\female} & \textbf{80.50} & 75.75 & \colorbox{green!30}{4.75} \\
    American  \textcolor{blue}{\male} & \textbf{79.50} & 77.75 & \colorbox{green!30}{1.75} \\
    British  \textcolor{magenta}{\female} & \textbf{80.00} & 74.75 & \colorbox{green!30}{5.25} \\
    British  \textcolor{blue}{\male} & \textbf{79.50} & 78.25 & \colorbox{green!30}{1.25} \\
    Indian  \textcolor{magenta}{\female} & \textbf{79.25} & 78.25 & \colorbox{green!30}{1.00} \\
    Indian  \textcolor{blue}{\male} & \textbf{78.75} & 77.25 & \colorbox{green!30}{1.50} \\
    \bottomrule
    \end{tabular}
    \caption{Accuracy comparison for \textit{Voxtral-Small-2507}.}
    \label{tab:accuracy-voxtral-small-2507}
\end{subtable}
\hfill
\begin{subtable}[t]{0.48\textwidth}
    \centering
    \begin{tabular}{cccc}
    \toprule
    \multicolumn{4}{c}{English} \\
    \cmidrule(lr){1-4}
    input & original & reversed & $\Delta$ \\
    \midrule
    American  \textcolor{magenta}{\female} & \textbf{48.50} & 46.50 & \colorbox{green!30}{2.00} \\
    American  \textcolor{blue}{\male} & \textbf{52.50} & 46.75 & \colorbox{green!30}{5.75} \\
    British  \textcolor{magenta}{\female} & \textbf{48.50} & 47.00 & \colorbox{green!30}{1.50} \\
    British  \textcolor{blue}{\male} & \textbf{49.50} & 47.75 & \colorbox{green!30}{1.75} \\
    Indian  \textcolor{magenta}{\female} & \textbf{50.75} & 45.25 & \colorbox{green!30}{5.50} \\
    Indian  \textcolor{blue}{\male} & \textbf{49.25} & 44.75 & \colorbox{green!30}{4.50} \\
    \bottomrule
    \end{tabular}
    \caption{Accuracy comparison for \textit{Voxtral-Mini-2507}.}
    \label{tab:accuracy-voxtral-mini-2507}
\end{subtable}
\caption{Accuracy comparison of Voxtral models under different option-order settings, grouped by language, accent, and gender. Note that the language setting is limited to English, as Voxtral currently does not support Korean or Chinese.}
\label{tab:accuracy-voxtral}
\end{table*}
\begin{table*}[t]
\centering
\footnotesize 
\setlength{\tabcolsep}{2pt} 
\renewcommand{\arraystretch}{1.1}
\begin{tabular}{@{}cccc cccc cccc@{}} 
\toprule
\multicolumn{4}{c}{Chinese} & \multicolumn{4}{c}{English} & \multicolumn{4}{c}{Korean} \\
\cmidrule(lr){1-4} \cmidrule(lr){5-8} \cmidrule(lr){9-12}
Setting & \textcolor{magenta}{\female} & \textcolor{blue}{\male} & $\Delta$
& Setting & \textcolor{magenta}{\female} & \textcolor{blue}{\male} & $\Delta$
& Setting & \textcolor{magenta}{\female} & \textcolor{blue}{\male} & $\Delta$ \\
\midrule
Beijing original      & \textbf{66.75} & 62.75 & \colorbox{green!30}{4.00}
& American original   & 80.00 & \textbf{82.00} & \colorbox{blue!30}{-2.00}
& Jeolla original     & \textbf{63.50} & 61.50 & \colorbox{green!30}{2.00} \\
Beijing reversed      & \textbf{64.00} & 61.50 & \colorbox{green!30}{2.50}
& American reversed   & 78.50 & \textbf{79.75} & \colorbox{blue!30}{-1.25}
& Jeolla reversed     & 58.50 & \textbf{61.00} & \colorbox{blue!30}{-2.50} \\
Northeastern original & 64.50 & \textbf{64.75} & \colorbox{blue!30}{-0.25}
& British original    & \textbf{81.25} & 81.00 & \colorbox{green!30}{0.25}
& Seoul original      & \textbf{63.75} & 63.00 & \colorbox{green!30}{0.75} \\
Northeastern reversed & 57.75 & \textbf{61.25} & \colorbox{blue!30}{-3.50}
& British reversed    & 76.75 & \textbf{78.50} & \colorbox{blue!30}{-1.75}
& Seoul reversed      & \textbf{62.00} & 62.00 & 0.00 \\
\multicolumn{4}{c}{}
& Indian original     & \textbf{80.50} & 80.00 & \colorbox{green!30}{0.50}
& \multicolumn{4}{c}{} \\
\multicolumn{4}{c}{}
& Indian reversed     & \textbf{78.25} & 79.25 & \colorbox{blue!30}{-1.00}
& \multicolumn{4}{c}{} \\
\bottomrule
\end{tabular}
\caption{Accuracy comparison across \textit{gender} conditions for \textit{Gemini 2.5 Flash}. 
Each cell reports the mean accuracy (\%) for \textit{Female} and \textit{Male}, 
with $\Delta$ denoting the difference (Female - Male). 
Results are grouped by \textit{language} (Chinese, English, Korean), \textit{accent}, and \textit{option order}.} 
\label{tab:accuracy-gemini-2.5-flash-gender}
\end{table*}

\begin{table*}[t] 
\footnotesize 
\centering \setlength{\tabcolsep}{4pt} \renewcommand{\arraystretch}{1.15} \begin{tabular}{ccc cccc ccc} \toprule \multicolumn{3}{c}{Chinese} & \multicolumn{4}{c}{English} & \multicolumn{3}{c}{Korean} \\ \cmidrule(lr){1-3} \cmidrule(lr){4-7} \cmidrule(lr){8-10} Setting & Beijing & Northeastern & Setting & American & British & Indian & Setting & Seoul & Jeolla \\ \midrule original \textcolor{magenta}{\female} & \textbf{66.75} & 64.50 & original \textcolor{magenta}{\female} & 80.00 & \textbf{81.25} & 80.50 & original \textcolor{magenta}{\female} & 63.75 & \textbf{63.50} \\ original \textcolor{blue}{\male} & 62.75 & \textbf{64.75} & original \textcolor{blue}{\male} & \textbf{82.00} & 81.00 & 80.00 & original \textcolor{blue}{\male} & \textbf{63.00} & 61.50 \\ reversed \textcolor{magenta}{\female} & \textbf{64.00} & 57.75 & reversed \textcolor{magenta}{\female} & \textbf{78.50} & 76.75 & 78.25 & reversed \textcolor{magenta}{\female} & \textbf{62.00} & 58.50 \\ reversed \textcolor{blue}{\male} & \textbf{61.50} & 61.25 & reversed \textcolor{blue}{\male} & \textbf{79.75} & 78.50 & 79.25 & reversed \textcolor{blue}{\male} & \textbf{62.00} & 61.00 \\ \bottomrule \end{tabular} \caption{Accuracy comparison across \textit{accents} conditions for \textit{Gemini 2.5 Flash}. Each cell reports the mean accuracy (\%) for each accents, grouped by \textit{language} (Chinese, English, Korean), \textit{option order}, and \textit{gender}.} 
\label{tab:accuracy-gemini-2.5-flash-accent} 
\end{table*} 
\begin{table*}[t]
\footnotesize 
\centering \setlength{\tabcolsep}{6pt} \renewcommand{\arraystretch}{1.15} \begin{tabular}{c ccc} \toprule Setting & Chinese & English & Korean \\ \midrule original \textcolor{magenta}{\female} & 65.62 & \textbf{80.58} & 63.62 \\ original \textcolor{blue}{\male} & 63.75 & \textbf{81.00} & 62.25 \\ reversed \textcolor{magenta}{\female} & 60.88 & \textbf{77.83} & 60.25 \\ reversed \textcolor{blue}{\male} & 61.38 & \textbf{79.17} & 61.50 \\ \bottomrule \end{tabular} \caption{Accuracy comparison across \textit{language} conditions for \textit{Gemini 2.5 Flash}. Each cell reports the mean accuracy (\%) for each language, grouped by \textit{option order} and \textit{gender}.} 
\label{tab:accuracy-gemini-2.5-flash-lang} 
\end{table*}

\subsection{CS vs. CA under Variable Perturbations}
\label{cscavariableperturbations}
We further stratify questions into Culturally Sensitive (CS) and Culturally Agnostic (CA) subsets to assess cross-variable robustness under \textit{language}, \textit{accent}, \textit{gender}, and \textit{option order} perturbations. Figures~\ref{fig:new_figs_figure_three_panel} and~\ref{fig:new_figs_kappa_vs_apes_quadrants_groups} show that across models and variables, CA items consistently exhibit lower uncertainty (APES) than CS items, indicating stronger robustness under cross-variable shifts
\subsection{Accuracy Analysis}
\label{sec:appendix-acc}
Tables~\ref{tab:accuracy-gemini-2.5-flash-lite} to~\ref{tab:accuracy-voxtral} provide accuracy comparison across \textit{option order} grouped by \textit{language}, \textit{accent}, and \textit{gender} for the other eight models: Gemini 2.5 Flash Lite, Gemini 2.0 Flash, Gemini 2.0 Flash Lite, Gemma 3n E4B, Gemma 3n E2B, Voxtral Small, Voxtral Mini, Phi 4 Multimodal.
Tables~\ref{tab:accuracy-gemini-2.5-flash-gender} to~\ref{tab:accuracy-gemini-2.5-flash-lang} provide a factor-wise breakdown for Gemini 2.5 Flash, reporting accuracy comparisons grouped by \textit{gender}, \textit{accent}, and \textit{language}, respectively. 
These stratified analyses clarify not only whether option reordering affects accuracy, but also how its impact interacts with speech-specific factors, yielding a more interpretable picture of model behavior under controlled variations.

\clearpage
\begin{table*}[t]
\centering
\footnotesize
\begin{tabular}{lrrrrrr rrr}
\toprule
\multirow{2}{*}{\textbf{Model}} 
& \multicolumn{3}{c}{\textbf{Language}} 
& \multicolumn{3}{c}{\textbf{Accent}} 
& \multicolumn{3}{c}{\textbf{Gender}} \\
\cmidrule(lr){2-4}
\cmidrule(lr){5-7}
\cmidrule(lr){8-10}
& Ori & Rev & $\Delta$ 
& Ori & Rev & $\Delta$ 
& Ori & Rev & $\Delta$ \\
\midrule
Gemini 2.5 Flash
& 0.182 & 0.190 & \colorbox{green!30}{0.008}
& 0.156 & 0.168 & \colorbox{green!30}{0.012}
& 0.103 & 0.111 & \colorbox{green!30}{0.008} \\
Gemini 2.5 Flash Lite
& 0.238 & 0.252 & \colorbox{green!30}{0.014}
& 0.218 & 0.226 & \colorbox{green!30}{0.008}
& 0.123 & 0.129 & \colorbox{green!30}{0.006} \\
Gemma 3n E2B 
& 0.288 & 0.289 & \colorbox{green!30}{0.001}
& 0.213 & 0.211 & \colorbox{blue!30}{-0.002}
& 0.166 & 0.163 & \colorbox{blue!30}{-0.003} \\
Gemma 3n E4B 
& 0.271 & 0.265 & \colorbox{blue!30}{-0.006}
& 0.194 & 0.193 & \colorbox{blue!30}{-0.001}
& 0.158 & 0.142 & \colorbox{blue!30}{-0.016} \\
Phi~4 Multimodal
& 0.290 & 0.285 & \colorbox{blue!30}{-0.005}
& 0.172 & 0.188 & \colorbox{green!30}{0.016}
& 0.156 & 0.167 & \colorbox{green!30}{0.011} \\
\bottomrule
\end{tabular}
\caption{APES under original and reversed option order for language, accent, and gender.}
\label{tab:7}
\end{table*}
\begin{table*}[ht]
\centering
\footnotesize
\begin{tabular}{lcccc}
\toprule
\textbf{Model} & \textbf{Original} & \textbf{Order Backward} & \textbf{Token Backward} & \textbf{Reversed} \\
\midrule
Gemini 2.5 Flash& \textbf{0.267} & 0.328 & 0.309 & 0.297 \\
Gemini 2.5 Flash Lite & \textbf{0.440} & 0.503 & 0.499 & 0.445 \\
Gemma 3n E2B  & 0.517 & 0.492 & \textbf{0.489} & 0.524 \\
Gemma 3n E4B  & 0.472 & 0.494 & \textbf{0.465} & 0.493 \\
Phi~4 Multimodal & 0.388 & \textbf{0.353} & 0.366 & 0.415 \\
\bottomrule
\end{tabular}
\caption{Mean entropy under different option-order perturbations. Lower values indicate more stable (less order-sensitive) behavior.}
\label{tab:8}
\end{table*}
\clearpage
\subsection{Supplementary Results for Impact of Model Scale}
\label{appendix:model_scaling}
\begin{figure}[htbp]
    \centering
    \includegraphics[width=\linewidth]{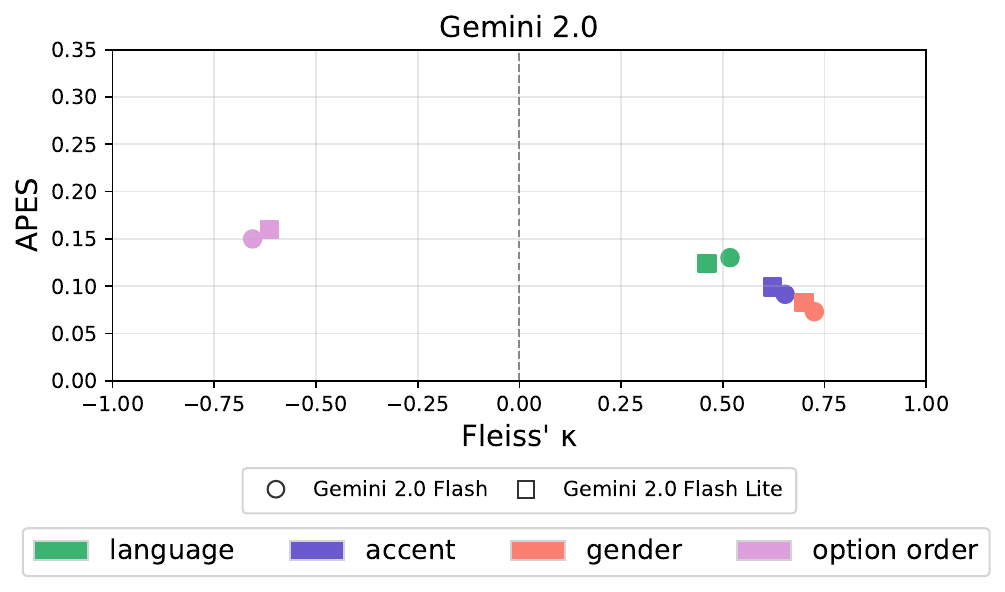}
    \caption{Fleiss' $\kappa$ versus APES by Gemini 2.0 family.}
    \label{fig:kappa_vs_apes_gemini_2.0}
\end{figure}

Figure~\ref{fig:kappa_vs_apes_gemini_2.0} presents the scaling results for the Gemini 2.0 series, showing the same trend as Figure~\ref{fig:kappashift}: larger variants yield higher agreement (Fleiss’ $\kappa$) 
and lower uncertainty (APES) across variables.

\subsection{Supplementary Results for Impact of Option Reordering}
\label{appendix:Option_Reordering}
Table~\ref{tab:7} reports APES under the original and fully reversed option orders, summarized by language, accent, and gender. For all models, the APES differences between the original and reversed settings are small across the three factors. In both settings, language yields larger APES values than accent and gender, and the ordering of factor magnitudes remains the same (\textit{language} > \textit{accent} > \textit{gender}).

Table~\ref{tab:8} reports mean entropy under four option-order configurations (original, order-backward, token-backward, and fully reversed). Across models, entropy values vary across configurations within a limited range. For each model, the table identifies which configuration attains the lowest mean entropy, with the lowest-entropy setting differing across models. Overall, these tables provide additional measurements of robustness under alternative option-order perturbations.
\end{document}